\documentclass[10pt,twocolumn,letterpaper]{article}

\usepackage[pagenumbers]{cvpr} %

\usepackage{xcolor}
\definecolor{linkcolor}{RGB}{0, 51, 153}
\usepackage{hyperref}
\hypersetup{
  colorlinks=true,
  linkcolor=linkcolor,
  citecolor=linkcolor,
  urlcolor=linkcolor,
  pdfauthor={},
  pdftitle={}
}
\usepackage{url}

\usepackage{amsmath,amssymb,amsfonts}
\usepackage{amsthm}

\theoremstyle{definition}

\usepackage{algorithm}
\usepackage{algpseudocode}

\usepackage{graphicx}
\usepackage{calc}
\usepackage{float}
\usepackage{subcaption}
\usepackage{booktabs}
\usepackage{multirow}
\usepackage{array}
\usepackage{enumitem}
\usepackage{changepage}
\usepackage{fancyvrb}
\usepackage{verbatim}
\usepackage{nicefrac}
\usepackage{pgfplots}
\pgfplotsset{compat=1.17}

\usepackage[table]{xcolor}

\usepackage[font=footnotesize,labelfont=bf,justification=raggedright,singlelinecheck=false]{caption}
\usepackage{fvextra}
\DefineVerbatimEnvironment{Verbatim}{Verbatim}
  {fontsize=\small, breaklines=true, breaksymbolleft={}, breakanywhere=true}

\usepackage{tikz}
\usetikzlibrary{arrows.meta,calc,positioning}

\usepackage[numbers,sort&compress]{natbib}
\newcommand{\best}[1]{\textbf{#1}}

\newlength{\imwid}
\newlength{\tagwid}

\newcolumntype{M}{>{\centering\arraybackslash}m{\ThumbSize}}

\definecolor{cvprblue}{rgb}{0.21,0.49,0.74}

\def\paperID{*****} %
\def\confName{CVPR}
\def\confYear{2026}

\title{Delta Rectified Flow Sampling for Text-to-Image Editing}

\author{
Gaspard Beaudouin\textsuperscript{1,2}\thanks{This work was done during the author's internship at the Harvard AI and Robotics Lab, Harvard University.}\quad
Minghan Li\textsuperscript{1}\quad
Jaeyeon Kim\textsuperscript{3}\\
Sung-Hoon Yoon\textsuperscript{1,4}\thanks{Co-corresponding author. \href{mailto:shyoon@dgist.ac.kr}{\texttt{shyoon@dgist.ac.kr}}}\quad
Mengyu Wang\textsuperscript{1,5}\thanks{Co-corresponding author.  \href{mailto:mengyu_wang@meei.harvard.edu}{\texttt{mengyu\_wang@meei.harvard.edu}}}\\[6pt]
\textsuperscript{1}Harvard AI and Robotics Lab, Harvard University\\
\textsuperscript{2}École Nationale des Ponts et Chaussées, Institut Polytechnique de Paris\\
\textsuperscript{3}Computer Science Department, 
Harvard University\\
\textsuperscript{4}Multimodal Intelligence and Perception Lab,
DGIST\\
\textsuperscript{5}Kempner Institute for the Study of Natural and Artificial Intelligence, Harvard University
}

\begin{document}
\maketitle

\vspace{-1mm}
\newcommand{\gradcap}[1]{#1}

\begin{abstract}

\noindent We propose Delta Rectified Flow Sampling (DRFS), a novel inversion-free, path-aware editing framework within rectified flow models for text-to-image editing. DRFS is a distillation-based method that explicitly models the discrepancy between the source and target velocity fields in order to mitigate over-smoothing artifacts rampant in prior distillation sampling approaches. We further introduce a time-dependent \textit{shift} term to push noisy latents closer to the target trajectory, enhancing the alignment with the target distribution. We theoretically demonstrate that disabling this \textit{shift} recovers Delta Denoising Score (DDS), bridging score-based diffusion optimization and velocity-based rectified-flow optimization. Moreover, under rectified-flow dynamics, a linear \textit{shift} schedule recovers the inversion-free method FlowEdit as a strict special case, yielding a unifying view of optimization and ODE editing. %
We conduct an analysis to guide the design of our shift term, and experimental results on the widely used PIE Benchmark indicate that DRFS achieves superior editing quality, fidelity, and controllability while requiring no architectural modifications. Code is available at \url{https://github.com/Harvard-AI-and-Robotics-Lab/DeltaRectifiedFlowSampling}.

\end{abstract}

\section{Introduction}

Diffusion-based and flow-based generative models \cite{Rombach_2022_CVPR_sd1_5,esser2024scalingrectifiedflowtransformers,sauer2024sd3.5,flux2024} have recently achieved remarkable success in high-fidelity image synthesis and editing, particularly in text-to-image (T2I) applications~\cite{wang2025taming,xu2025unveilinversioninvarianceflow,poole2023dreamfusion, Hertz_2023_ICCV}. A common approach to text-guided image editing involves optimizing an input image to align with a new target prompt, while preserving regions that should remain unchanged. 

T2I editing has evolved along two primary lines: non-energy-based methods %
and energy-based optimization methods. 
Non-energy-based  %
methods, such as RF-inversion~\cite{rout2025semantic}, typically perform editing through two conditional velocity fields: one for inversion and one for generation.
These methods often combine heuristic strategies such as attention injection or latent averaging to improve fidelity and controllability~\cite{deng2025fireflow,wang2025taming,xu2025unveilinversioninvarianceflow}. For example, FTEdit~\cite{xu2025unveilinversioninvarianceflow} reduces artifacts by averaging outputs across multiple inversion steps, effectively trading off speed for stability. FlowEdit~\cite{kulikov2024floweditinversionfreetextbasedediting} %
eliminates the explicit inversion phase. Instead, it directly estimates the target latent by calculating the offset between source and target velocities, enabling faster inference while maintaining editability. %

\begin{figure}[t]
    \centering
    \begin{subfigure}[t]{0.23\linewidth}
        \centering
        \includegraphics[width=\linewidth]{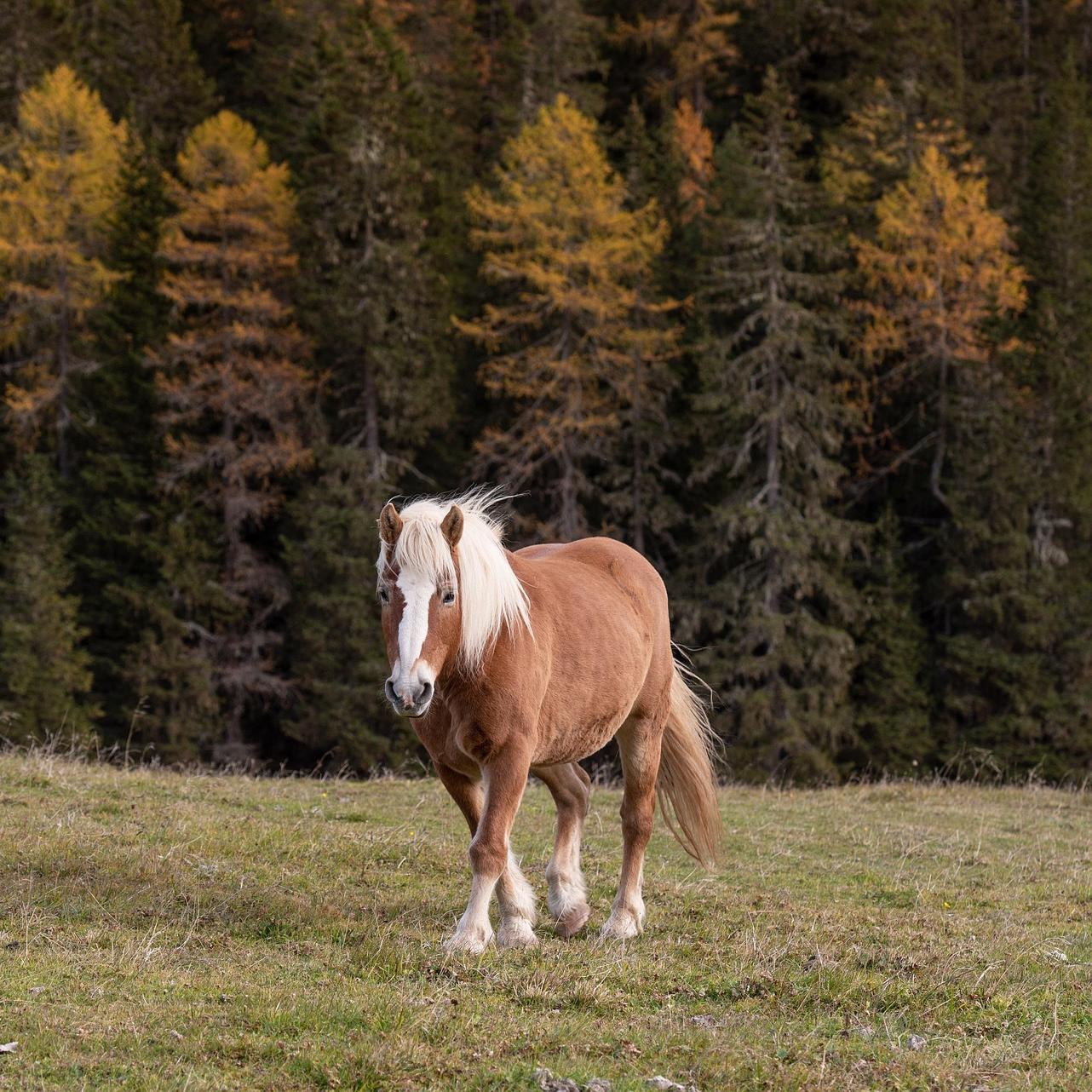}
        \caption{Source \\image}
        \label{fig:img1}
    \end{subfigure}
    \hfill
    \begin{subfigure}[t]{0.23\linewidth}
        \centering
        \includegraphics[width=\linewidth]{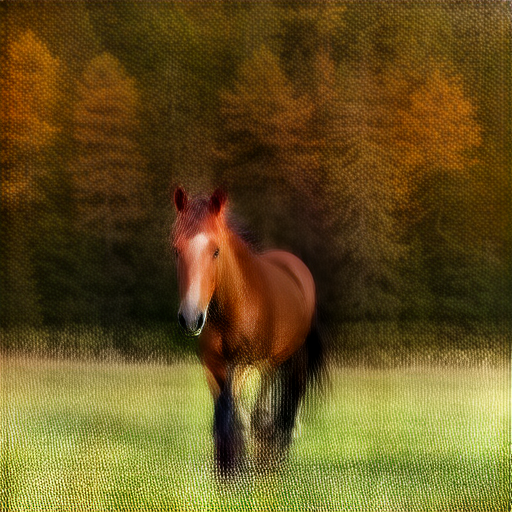}
        \caption{RFDS \\ w/ src prompt}
        \label{fig:img1.5}
    \end{subfigure}
    \hfill
    \begin{subfigure}[t]{0.23\linewidth}
        \centering
        \includegraphics[width=\linewidth]{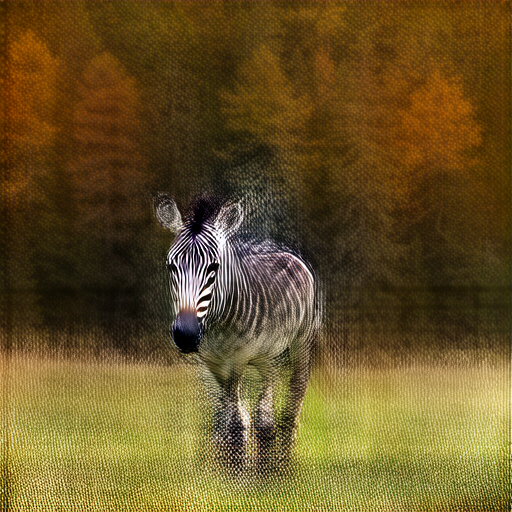}
        \caption{RFDS \\w/ tgt prompt}
        \label{fig:img2}
    \end{subfigure}
    \hfill
    \begin{subfigure}[t]{0.23\linewidth}
        \centering
        \includegraphics[width=\linewidth]{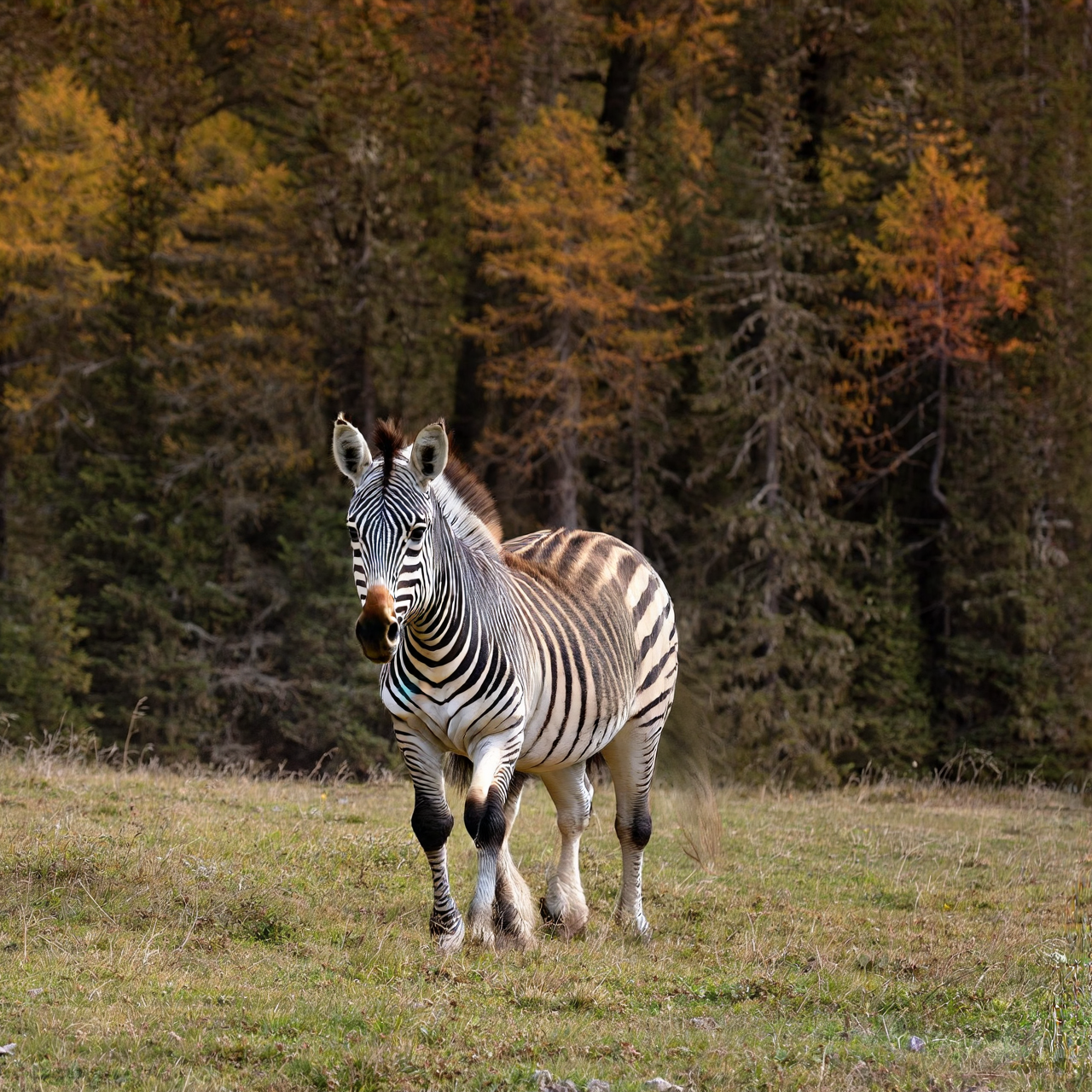}
        \caption{DRFS \\w/ tgt prompt}
        \label{fig:img3}
    \end{subfigure}
    \caption{
\textbf{Comparison between RFDS and DRFS (ours)}. Source prompt: \textit{Brown horse} walking in a grassy meadow with an autumn forest backdrop and target prompt: \textit{Zebra} walking in a grassy meadow with an autumn forest backdrop. As shown in (b) and (c), RFDS results in over-smoothing and detail loss. In contrast, DRFS (d) preserves textures.
    }\label{fig:oversmooth}
    \vspace{-2mm}
\end{figure}

In contrast, energy-based approaches~\cite{poole2023dreamfusion,Hertz_2023_ICCV,yang2025texttoimage} formulate image editing as an explicit optimization problem over the noise or velocity space. Among them, Score Distillation Sampling (SDS)~\cite{poole2023dreamfusion} and Delta Denoising Score (DDS)~\cite{Hertz_2023_ICCV} define loss functions based on predicted noise residuals, allowing for efficient optimization guided by frozen diffusion priors. Building on this idea, Rectified Flow Distillation Sampling (RFDS)~\cite{yang2025texttoimage} extends energy-based methods into the velocity field of rectified flow models. By directly optimizing the sample trajectory using gradients from pre-trained T2I Rectified Flow priors, RFDS enables effective plug-and-play editing. However, a key limitation of RFDS for editing is its \emph{over-smoothing} artifacts, where background and high-frequency details are unintentionally altered, compromising visual fidelity.

In this paper, we introduce \textbf{Delta Rectified Flow Sampling (DRFS)}, a novel text-to-image (T2I) editing method within the rectified flow framework. DRFS addresses the over-smoothing issue inherent in RFDS by proposing a novel energy function building on the intuition of DDS, %
and also boosts editing performance by introducing a time-scaled \textit{shift} term. This shift promotes much better alignment with the target distribution, enhances semantic consistency while preserving fine-grained visual details, making DRFS a path-aware formulation that explicitly leverages the editing trajectory.

Our empirical evaluations demonstrate that DRFS outperforms existing state-of-the-art methods, such as FlowEdit~\cite{kulikov2024floweditinversionfreetextbasedediting}, FTEdit~\cite{xu2025unveilinversioninvarianceflow} and DNAEdit~\cite{xie2025dnaedit}. On the PIE benchmark, DRFS consistently reduces the editing distance and improves background preservation metrics compared to both diffusion-based and rectified-flow baselines, while achieving the best edited-region CLIP similarity among SD3- and SD3.5-based methods (Table~\ref{tab:pie_results}). 
Moreover, we demonstrate that this shift term also provides a cohesive theoretical framework, unifying existing distillation-based methods and ODE-based FlowEdit under a generalized viewpoint. Our contributions are threefold:

(i) \textbf{A rectified-flow specific objective for T2I editing: residual subtraction.}
Unlike a naive ``delta'' that subtracts conditional predictions, DRFS subtracts \emph{full residuals} between model velocity and data dynamics, which cancels components shared by source and target dynamics and yields an RF-specific drift term in the gradient.

(ii) \textbf{A shifted evaluation state to correct trajectory mismatch.}
We explicitly formulate the model-data mismatch induced by evaluating target velocities off the forward posterior of the ideal edited latent, and introduce a control variable $c_t$ that reduces this mismatch and stabilizes optimization.

(iii) \textbf{A unifying view of optimization and ODE sampling.}
Under rectified-flow parameterization, setting $c_t=t$ recovers the FlowEdit trajectory as a strict special case; we also connect DRFS to DDS and derive stability-driven design criteria for $c_t$.

\section{Background and Over-smoothing in RFDS}

\paragraph{Flow Matching and Rectified Flow for Generation and Editing.}
Flow matching models \cite{albergo2023stochasticinterpolantsunifyingframework,lipman2023flow,DBLPLiuG023}, for text-to-image generation learn a velocity field $v_{\theta}$ that transports samples from a distribution $p_1$ that is tractable (typically a standard Gaussian $\mathcal{N}(0,I))$, to a distribution $p_0$ that we want to model (e.g. the distribution over images). A trajectory from $p_1$ to $p_0$ is defined by the velocity field via the ordinary differential equation:
\begin{equation}\label{eq: flow_mathcing}
\mathrm{d}x_t = v_\theta(x_t, t)\mathrm{d}t, \quad {t: 1 \rightarrow 0}, \ x_1 \sim p_1. %
\end{equation}
To train this velocity field, pairs from the source and target distributions are interpolated using time-dependent noise scheduling parameters $(a_t,b_t)$, yielding intermediate states $x_t = a_t x_0 + b_t x_1, x_0 \sim p_0 , x_1 \sim p_1$. The training objective, called the conditional flow matching loss, is defined as
\begin{align*}
   \mathcal{L}(\theta) = \mathbb{E}_{t,x_t}\left[\left\| v_\theta(a_{t}x_{0}+b_{t}x_{1},\,t)-(\dot{a}_{t}x_{0}+\dot{b}_{t}x_{1})\right\|^2 \right].
\end{align*}

Rectified Flow (RF) \cite{DBLPLiuG023} further simplifies this process by assuming a straight-line trajectory in the latent space, with $a_t = 1-t, \ b_t = t$. Typically, models are trained conditioned on a text prompt $\varphi$, resulting in a velocity vector field $v_\theta(x_t,t,\varphi)$ that transports toward a conditional target distribution (i.e. images corresponding to the given text prompt $\varphi$). In practice, sampling from the ordinary differential equation involves Classifier Free Guidance (CFG) \citep{ho2021classifierfree} with a guidance scale $w>1$: $\tilde v_\theta(x_t^{}, t, \varphi)=w(v_\theta(x_t^{}, t, \varphi^{})- v_\theta(x_t^{}, t, \varnothing))+ v_\theta(x_t^{}, t, \varnothing)$, where $\varnothing$ is the null prompt.

Text-to-image (T2I) editing~\cite{deng2025fireflow,kulikov2024floweditinversionfreetextbasedediting} leverages the alignment priors of generative models to modify an input image \( x_0^{\text{src}} \) described by a source prompt \( \varphi^{\text{src}} \), and produce an edited image \( x_0^{\text{tgt}} \) that semantically aligns with a target prompt \( \varphi^{\text{tgt}} \).

\paragraph{Diffusion Model Distillation Sampling.}
Score Distillation Sampling (SDS)~\cite{poole2023dreamfusion}, Delta Denoising Score (DDS)~\cite{Hertz_2023_ICCV}, and their variants \cite{wang2023prolificdreamer,katzir2024noisefree, mcallister2024rethinking,Koo:2024PDS} formulate image generation or editing as an optimization problem over an energy function derived from diffusion models. These techniques leverage pre-trained T2I diffusion priors to guide image synthesis that aligns with a given prompt.
Let $\Theta$ represent the parameters of a differentiable generator $g$, where $g(\Theta)$ is the output image to be optimized. The optimization objective for SDS and DDS can be expressed, respectively:

{\small
\begin{align}
    \mathcal{E}_{\text{SDS}}\bigl(x_0^{\mathrm{}} = g(\Theta), \varphi^{}\bigr)
    &= \mathbb{E}_{t, \varepsilon}\Bigl[
    \,
    \bigl\lVert
        \varepsilon_{\theta}(x_{t}^{\mathrm{}},t,\varphi^{\mathrm{}})
        - \varepsilon)
    \bigr\rVert^2 
    \Bigr], \label{eq:bg_sds} 
\end{align}
\begin{multline}
\mathcal{E}_{\text{DDS}}\bigl(
  x_{0}^{\text{tgt}}=g(\Theta),\,
  x_{0}^{\text{src}},\,
  \varphi^{\text{tgt}},\,
  \varphi^{\text{src}}
\bigr)
= \mathbb{E}_{t,\varepsilon}\!
  \Bigl[\,
    \bigl\|
      \varepsilon_\theta\bigl(x_{t}^{\text{tgt}},t,\varphi^{\text{tgt}}\bigr)
\\[-1pt] %
    -\,\varepsilon_\theta\bigl(x_{t}^{\text{src}},t,\varphi^{\text{src}}\bigr)
    \bigr\|^{2}
  \Bigr],
\label{eq:bg_dds}
\end{multline}

}
where $\varepsilon_{\theta}$ is the predicted noise from diffusion models. SDS aligns images with text prompts by minimizing the gap between predicted and true noise. DDS, tailored for T2I editing, matches source and target denoising trajectories to better preserve backgrounds.

\paragraph{Rectified Flow Distillation Sampling.} RFDS~\cite{yang2025texttoimage} extends the SDS from diffusion models to flow matching by defining the following energy function with the velocity field $v_\theta$ in Eq. \eqref{eq: flow_mathcing}:
\begin{align}
    \mathcal{E}_{\mathrm{\text{RFDS}}}\bigl(x_0^{\mathrm{}} = g(\Theta), \varphi^{}\bigr)
= \mathbb{E}_{t, \varepsilon}\Bigl[
    \bigl\lVert
        ( v_\theta(x_t^{}, t, \varphi^{}) - \dot x_t
    \bigr\rVert^2
\Bigr], 
\label{eq: rfds_loss}
\end{align}

where $\dot x_t = \dot a_t x_0^{} + \dot b_t \varepsilon$, $\dot{a}_t$ and $\dot{b}_t$ denote the time derivatives of the noise schedulers $a_t$ and $b_t$, respectively. With $w_{\text{RFDS}}$ a weighting function (often set to 1), the gradients w.r.t. generator parameters $\Theta$ and the Gaussian noise $\varepsilon$ are respectively approximated as:
{\small
\begin{align*}
    &\nabla_\Theta \mathcal{E}_{\text{RFDS}}(x_0^{\text{}}=g(\Theta), \varphi^{}) \simeq \\
    & \quad \mathbb{E}_{t, \varepsilon} \left[w_{\text{RFDS}}(t)\left( v_\theta(x_t^{}, t, \varphi^{}) - \dot x_t \right)]\frac{\partial x_0^{}}{\partial \Theta} \right].
\end{align*}}

\paragraph{Over-smoothing in RFDS.} RFDS can be applied to T2I editing by using \( \varphi = \varphi^{\text{tgt}} \) in Eq.~\eqref{eq: rfds_loss}. However, as shown in Fig. \ref{fig:oversmooth} (b) and (c), it suffers from over-smoothing and loss of source image details during the editing process. To address this,  \cite{yang2025texttoimage} additionally proposed iRFDS to invert the image (by optimizing a noise $\varepsilon$ in order to minimize \eqref{eq: rfds_loss}) before editing, obtaining a favorable noise aligned with the image structure. However, this requires additional computational cost. We observe that, similar to SDS, the root cause of over-smoothing in RFDS lies in the gradient term $\nabla_{\Theta} \mathcal{E}_{\mathrm{\text{RFDS}}}$  in Eq.~\eqref{eq: rfds_loss},  which fails to distinguish between regions of the image that need editing and those that should be preserved. As a result, non-zero gradients appear even in regions that are supposed to remain unchanged, leading to the destruction of high-frequency details in those areas.

\section{Delta Rectified Flow Sampling (DRFS)} \label{sec::DRFS}
We introduce Delta Rectified Flow Sampling (DRFS), a DDS-inspired method designed for text-to-image editing that explicitly minimizes the distillation sampling discrepancy between source and target prompts, mitigating the over-smoothing issue observed in RFDS.

\paragraph{Mitigating over-smoothing in RFDS.}  %
We begin by revisiting a key design principle from DDS (Eq.~\eqref{eq:bg_dds}), which minimizes the difference between the velocities toward the source prompt $v_\theta(x_t^{\text{src}}) := v_\theta(x_t^{\text{src}}, t,\varphi^{\text{src}})$ and target prompt $v_\theta(x_t^{\text{tgt}}) := v_\theta(x_t^{\text{tgt}}, t,\varphi^{\text{tgt}})$. Building on this insight, one can define the energy function as:

\begin{align}
    \mathcal{E}
    = \mathbb{E}_{t, \varepsilon}\Bigl[
    \bigl\lVert
        v_{\theta}(x_{t}^{\text{tgt}}) - v_{\theta}(x_{t}^{\text{src}}) - (\dot{x}_t^{\text{tgt}} - \dot x_t^{\text{src}}))
    \bigr\rVert^2 \Bigr],
\label{eq:base_E_function}
\end{align}

where $x_t^{\text{tgt}} = a_t x_0^{\text{tgt}} + b_t\varepsilon$ and $x_t^{\text{src}} = a_t x_0^{\text{src}} + b_t\varepsilon$ and ($a_t$, $b_t$) are rectified flow noise schedulers. 
By introducing a residual $r=v(x_t,t,\varphi)-\dot x_t$ for each respective $\varphi$, one can notice that RFDS energy function becomes $\mathbb{E}_{t, \varepsilon}(\|r^{\text{tgt}}\|^2)$. \eqref{eq:base_E_function}, in contrast, is the difference between residuals with respect to the target and source prompts, i.e., $\mathcal{E} = \mathbb{E}_{t,\varepsilon}\left[\left\| r^{\text{tgt}}-r^{\text{src}}\right\|^2 \right]$ (see Fig. ~\eqref{fig:oversmooth} and ~\eqref{fig:dds-gradients}).

Consequently, the optimization only penalizes the differences between the source and target residuals, leaving the information common to both images essentially untouched.

\paragraph{Sanity Check.}
Consider the case where $\varphi^{\text{tgt}} = \varphi^{\text{src}}$. In that situation, the “common information" corresponds to the entire image, and optimizing the energy function should produce no change in the optimized image. At the beginning of optimization, $x_0^{\text{tgt}} = x_0^{\text{src}}$, hence $x_t^{\text{tgt}} = x_t^{\text{src}}$, which leads to $\mathcal{E} = 0$, exactly as expected. 

\begin{figure*}[t]
  \centering
  \begin{subfigure}[t]{0.32\linewidth}
    \vspace{0pt}
    \centering
    \includegraphics[width=\linewidth,height=4cm,keepaspectratio]{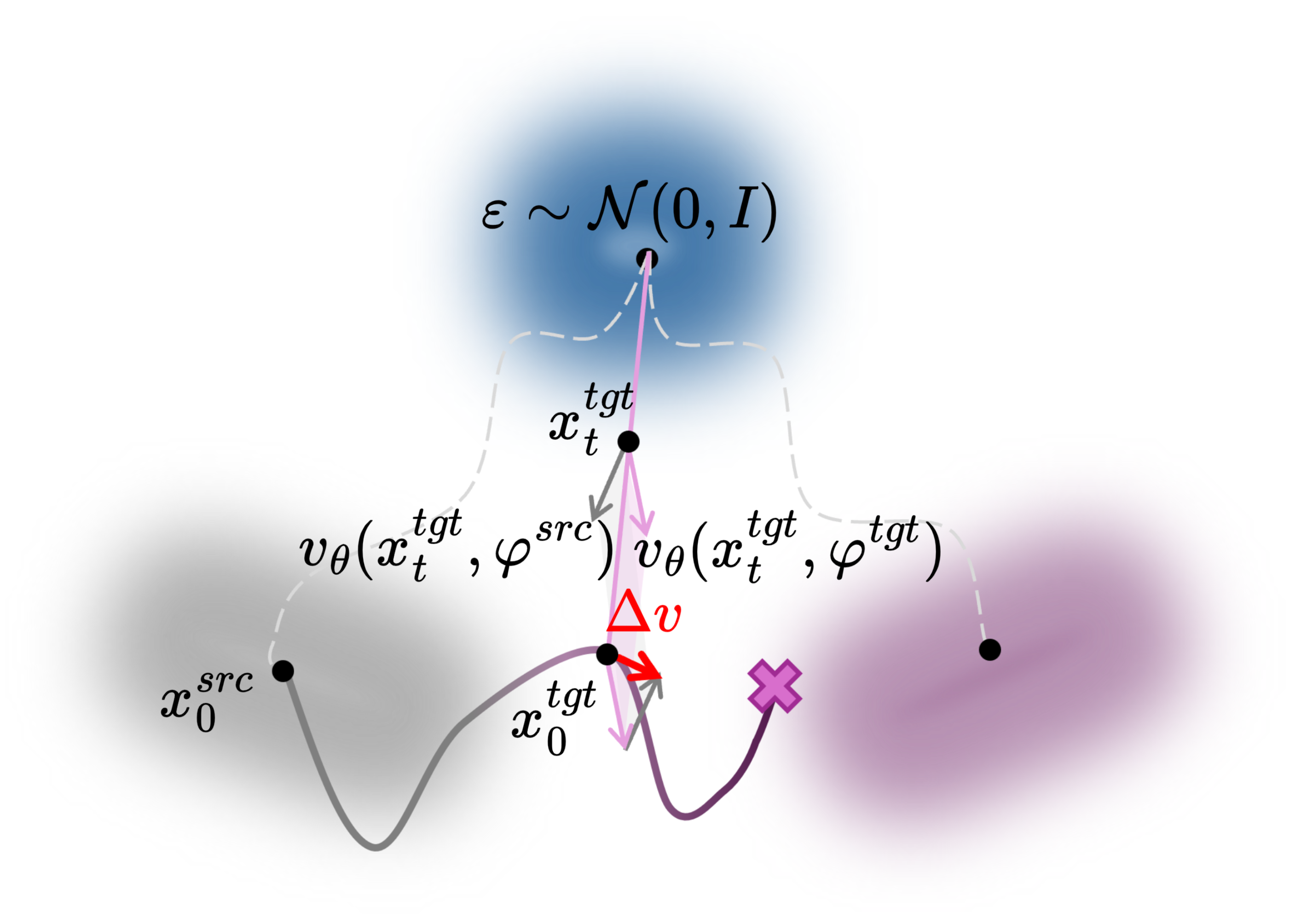}
    \caption{RFDS}
  \end{subfigure}\hfill
  \begin{subfigure}[t]{0.32\linewidth}
    \vspace{0pt}
    \centering
    \includegraphics[width=\linewidth,height=4cm,keepaspectratio]{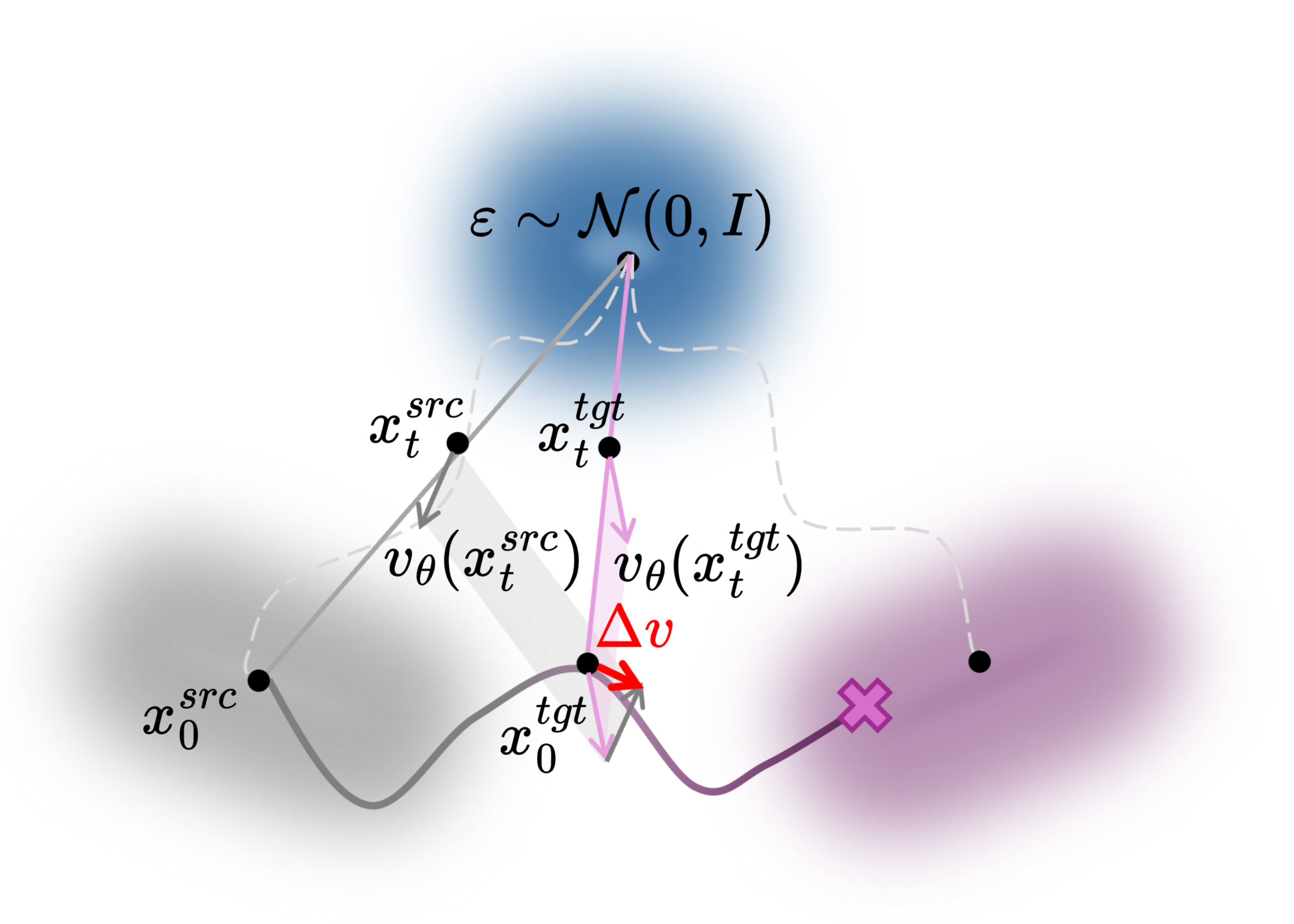}
    \caption{DRFS with $c_t=0$:\\  $ \hat x_t^{\text{tgt}}=x_t^{\text{tgt}}$}
  \end{subfigure}\hfill
  \begin{subfigure}[t]{0.32\linewidth}
    \vspace{0pt}
    \centering
    \includegraphics[width=\linewidth,height=4cm,keepaspectratio]{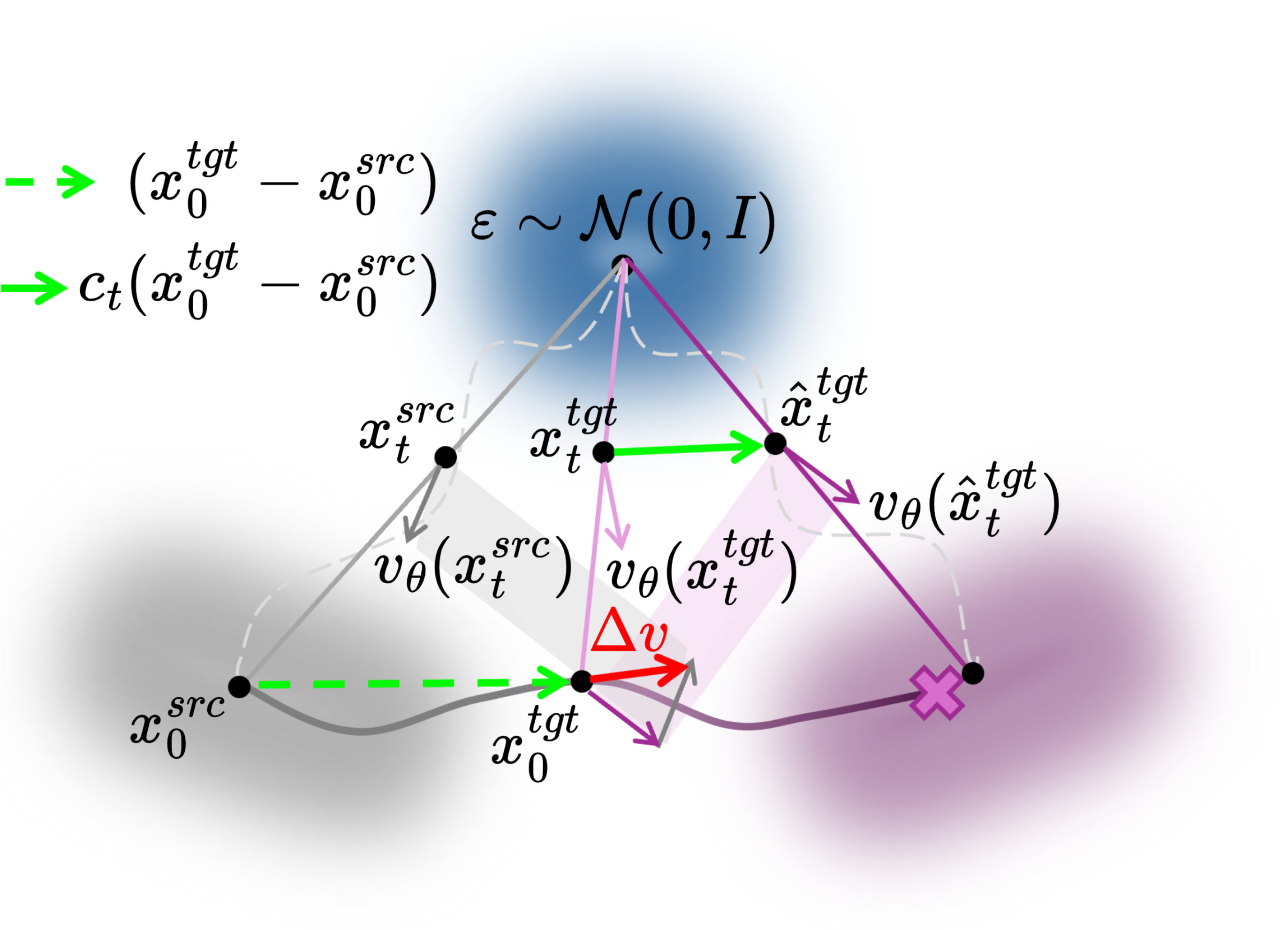}
    \caption{DRFS with $c_t>0$:\\$\hat x_t^{\text{tgt}}=x_t^{\text{tgt}}+c_t\bigl(x_0^{\text{tgt}}-x_0^{\text{src}}\bigr)$}
  \end{subfigure}
  \caption{ \textbf{Visual comparison of the sampling strategies for editing}. %
  When $c_t >0$, a shift term $c_t(x_0^{\text{tgt}} - x_0^{\text{src}})$ is added to $x_t^{\text{tgt}}$, pushing $x_t^{\text{tgt}}$closer to the target trajectory, enhancing precision in the evaluation of $v_{\theta}(\hat x_t^{\text{tgt}})$ to guide the optimization process. %
  The pink cross represents the final edited image.
    The green arrows indicate the offset vector from $x_0^{\text{src}}$ to $x_0^{\text{tgt}}$, scaled by $c_t$. Our new energy function optimization allows to reduce the RFDS oversmoothing, while introducing the offset (when $c_t \geq 0$) encourages a straighter and more stable trajectory from source to target distribution, aiding the desired target.} %
  \label{fig:dessins}
\end{figure*}

\paragraph{DRFS energy function.}
Using $x_t^{\text{tgt}}$ directly in the interpolation, however, 
may cause $x_t^{\text{tgt}}$ to deviate from the forward posterior of the target distribution. Since $x_0^{\text{tgt}}$ lies midway along the editing path from source to target, the interpolated $x_t^{\text{tgt}}$ might stray from the intended semantic trajectory. This misalignment weakens the editing effect and hinders convergence toward the desired target distribution (see Fig.~\ref{fig:dessins}(b)).

To address this, we introduce a simple linear compensation to correct $x_t^{\text{tgt}}$. Specifically, a modified target latent $ \hat{x}^{\text{tgt}}_t$ is defined as:
\begin{equation}\label{eq:hat_x_tgt}
    \hat{x}^{\text{tgt}}_t = a_t x^{\text{tgt}}_0 + b_t \varepsilon  + c_t (x_0^{\text{tgt}} - x_0^{\text{src}}).
\end{equation}
The offset term $c_t\,(x_0^{\text{tgt}}-x_0^{\text{src}})$, with $c_t\ge 0$, incrementally \textbf{aligns the sampling trajectory toward the target} distribution over time. The correction leads to a \textbf{more accurate target velocity} $v_\theta(\hat{x}_t^{\text{tgt}})$, mitigating the distortion caused by the misalignment between source and target paths (see Fig.~\ref{fig:dessins}(c)). To summarize, it reduces model–data mismatch early in optimization.

The DRFS energy function is defined as follows:
\begin{multline}
\mathcal{E}_{\text{DRFS}}\!\bigl(
  x_{0}^{\text{tgt}}\!=\!g(\Theta),\,
  x_{0}^{\text{src}},\,
  \varphi^{\text{tgt}},\,
  \varphi^{\text{src}}
\bigr) =\\
\mathbb{E}_{t,\varepsilon}\!\Bigl[
  \bigl\lVert
    v_\theta(\hat{x}_{t}^{\text{tgt}})
    - v_\theta(x_{t}^{\text{src}})
    - \bigl(\dot{\hat{x}}_{t}^{\text{tgt}}
           - \dot{x}_{t}^{\text{src}}\bigr)
  \bigr\rVert^{2}
\Bigr].
\label{eq:E_DRFS}
\end{multline}

By explicitly differentiating the source and target velocities, DRFS suppresses gradients in regions that should remain unchanged-such as backgrounds-effectively preserving them. We visualize how these gradients vanish in irrelevant areas in the Appendix. Furthermore, the introduction of the offset term enables a more accurate estimation of the target velocity. As a result, DRFS reduces over-smoothing %
\emph{and} improves editing performance.

\paragraph{Approximated gradient.} The gradient of DRFS with respect to the parameters $\Theta$ is
{\footnotesize
\begin{align*}
    \nabla_\Theta \mathcal{E}_{\text{DRFS}}
=2\,\mathbb{E}_{t, \varepsilon}\Big[
    \big(v_\theta(\hat x_t^{\text{tgt}})
    - v_\theta(x_t^{\text{src}}) - \underbrace{(\dot{\hat{x}}_t^{\text{tgt}} - \dot x_t^{\text{src}} )}_{(\dot a_t+\dot c_t)(x_0^{\text{tgt}}-x_0^{\text{src}})}\big)\\
    \, \bigl(\underbrace{\frac{\partial v_\theta(\hat x_t^{\text{tgt}})}{\partial \hat x_t^{\text{tgt}}}}_{\text{Network Jacobian}} \underbrace{\frac{\partial \hat x_t^{\text{tgt}}}{\partial x_0^{\text{tgt}}}}_{a_t+c_t}- \underbrace{\frac{\partial \dot{\hat{x}}_t^{\text{tgt}}}{\partial x_0^{\text{tgt}}}}_{\dot a_t + \dot c_t}\bigr)
    \underbrace{\frac{\partial x_0^{\text{tgt}}}{\partial\Theta}}_{\text{Generator Jacobian}}
\Big].
\end{align*}}

Following standard practice ~\citep{poole2023dreamfusion,yang2025texttoimage, patel2024steeringrectifiedflowmodels}, we approximate the network Jacobian term with the identity matrix to avoid the high computational cost of computing it explicitly. Additionally, we directly optimize the target latent, i.e., $\Theta = x_0^{\text{tgt}}$. These simplifications yield the following DRFS gradient:
\begin{align}
\nabla_\Theta \mathcal{E}_{\text{DRFS}}
  &= \mathbb{E}_{t,\varepsilon}\!\Bigl[
        w_{\text{DRFS}}(t)\,
        \bigl(
          v_\theta(\hat{x}_{t}^{\text{tgt}})
          - v_\theta(x_{t}^{\text{src}}) \bigr.\notag\\
  &\phantom{=\;\mathbb{E}_{t,\varepsilon}\!\Bigl[\,w_{\text{DRFS}}(t)\,}%
        \bigl.
          - (\dot a_t + \dot c_t)\bigl(x_{0}^{\text{tgt}} - x_{0}^{\text{src}}\bigr)
        \bigr)
      \Bigr],
\label{eq:DRFS_grad}
\end{align}
where $w_{\text{DRFS}}(t) = 2(a_t + c_t -\dot a_t -\dot c_t)$ is a time-dependent weighting function. To clarify, $\nabla_\Theta \mathcal{E}_{\text{DRFS}}$ is a function of $x_0^{\text{tgt}}$ and $x_0^{\text{src}}$.  As described in Algorithm ~\ref{alg:DRFSp}, the optimization process of DRFS is initialized with $x_0^{\text{tgt}} = x_0^{\text{src}}$ and $x_0^{\text{tgt}}$ is optimized via the approximated gradient in Eq. \eqref{eq:DRFS_grad}. At each step of optimization, $(t,\varepsilon)$ pair(s) are sampled to calculate $({\hat{x}}_t^{\text{tgt}},x_t^{\text{src}})$ to estimate the expectation.

\paragraph{Timestep schedulers and design choice of $c_t$.} We compare two strategies for sampling $(t, \varepsilon)$ pairs during optimization.
\begin{itemize}[leftmargin=1.5em, itemsep=0pt, parsep=0pt]
    \item \textit{Descending scheduler} begins with large $t$ 
    (high-noise latents) for coarse update and gradually shifts to small $t$ (low-noise latents) for refinement.
    \item \textit{Random scheduler} \cite{Hertz_2023_ICCV} samples $t$ uniformly at each step.
\end{itemize}
We adopt a descending timestep scheduler, which consistently produces better results in practice. In other words, we begin with large $t$ values and progressively decrease them as the optimization proceeds.

This choice of time-step scheduling motivates a first condition for the shift coefficient, %
which is having \textbf{$c_t \propto t$}, so that the shift vanishes as $t \rightarrow 0$. Indeed, with this formulation, the shift term $c_t (x_0^{\text{tgt}} - x_0^{\text{src}})$ naturally decays as $t$ decreases, aligning with the fact that $x_0^{\text{tgt}}$ moves closer to the target distribution in later stages. \emph{When $t \simeq 0$, $x_0^{\text{tgt}}$ should already lie in the target distribution}, so no further shift should be applied. We detail another required initial condition  ($c_t \rightarrow0 $ as $t \rightarrow 1$)  in Appendix ~\ref{app:shift}.

We now investigate the impact of the shift term on the editing path  in the following section.

\begin{algorithm}[h]
\caption{\textbf{DRFS} with descending timestep schedule}
\label{alg:DRFSp}
\begin{algorithmic}[1]
\Require Source image $x_0^{\text{src}}$; prompts $(\varphi^{\text{src}},\varphi^{\text{tgt}})$; batch size $B$; iterations $N$; weight $w_{\text{DRFS}}(\cdot)$; optimiser $\mathcal{O}$; schedule $\{\tau_j\}_{j=1}^{N}$ with $1=\tau_N>\dots>\tau_1>0$; shift coefficient $c_t$
\State $x_0^{\text{tgt}}\gets x_0^{\text{src}}$
\For{$k = 0$ \textbf{to} $N-1$}
  \State $t\gets\tau_{N-k}$;\; $g\gets\mathbf{0}$ \Comment{Descending schedule}
  \For{$i = 1$ \textbf{to} $B$} \Comment{Monte-Carlo sample}
    \State $\varepsilon_i \sim \mathcal{N}(0,I)$
    \State $x_t^{\text{src}}\gets(1-t)\,x_0^{\text{src}}+t\,\varepsilon_i$
    \State $\hat{x}_t^{\text{tgt}}\gets(1-t)\,x_0^{\text{tgt}} + t\,\varepsilon_i + c_t\!\left(x_0^{\text{tgt}}-x_0^{\text{src}}\right)$
    \State $g \mathrel{+}= \dfrac{w_{\text{DRFS}}(t)}{B}\!\Bigl[v_\theta(\hat{x}_t^{\text{tgt}})-v_\theta(x_t^{\text{src}})-(\dot a_t+\dot c_t)\!\left(x_0^{\text{tgt}}-x_0^{\text{src}}\right)\Bigr]$
  \EndFor
  \State $x_0^{\text{tgt}}\gets\mathcal{O}\!\left(x_0^{\text{tgt}},g\right)$
\EndFor
\State \Return $x_0^{\text{tgt}}$
\end{algorithmic}
\end{algorithm}

\section{Theoretical analysis of DRFS}
\label{sec:sec5}
In this section, we show that the DRFS framework  provides a unified theoretical perspective that encompasses both DDS~\cite{Hertz_2023_ICCV} and the inversion-free editing method FlowEdit~\cite{kulikov2024floweditinversionfreetextbasedediting}. We further analyze DRFS editing trajectories and show that the shift coefficient $c_t$ jointly controls path straightness, update magnitude, and error propagation along the trajectory (see Appendix~\ref{app:shift} for details).

\subsection{Connections to DDS and FlowEdit}
\label{sec:sec41}
\paragraph{DDS and DRFS.}
We prove that $\mathcal{E}_{\text{DRFS}}$ (Eq.~\ref{eq:E_DRFS}) reduces to $\mathcal{E}_{\text{DDS}}$ (Eq.~\ref{eq:bg_dds})  when $c_t = 0$. A flow matching model that predicts a velocity field $v_\theta(x,t,\varphi)$ is equivalent to a diffusion model that predicts a noise with the relation of $\varepsilon_{\theta}(x, t, \varphi) = \frac{a_t}{\dot b_t a_t -  \dot a_t b_t} ( v_{\theta}(x,t, \varphi)- \frac{\dot a_t}{a_t}x)
$ \cite{zheng2023guidedflowsgenerativemodeling}.Therefore, the noise difference appearing in $\mathcal{E}_{\text{DDS}}$ (Eq.~\ref{eq:bg_dds}) can be rewritten as
\begin{equation*}
\begin{aligned}
   &\varepsilon_{\theta}(x_{t}^{\text{tgt}},t,\varphi^{\text{tgt}})
-\varepsilon_{\theta}(x_{t}^{\text{src}},t,\varphi^{\text{src}})
=\\
&\quad \frac{a_{t}}{\dot{b}_{t}a_{t}-\dot{a}_{t}b_{t}}
\Bigl(
  v_\theta(x_{t}^{\text{tgt}})
  -\,v_\theta(x_{t}^{\text{src}})
  -\frac{\dot{a}_{t}}{a_{t}}
    \bigl(x_{t}^{\text{tgt}}-x_{t}^{\text{src}}\bigr)
\Bigr).   
\end{aligned}
\end{equation*}

Taking expectation on both sides with respect to $(t,\varepsilon)$, the left side corresponds to $\mathcal{E}_{\text{DDS}}$ Eq.~\eqref{eq:bg_dds}, while the right side becomes equivalent to Eq.~\eqref{eq:E_DRFS} in which $c_t=0$.

\paragraph{FlowEdit and DRFS.} FlowEdit~\cite{kulikov2024floweditinversionfreetextbasedediting} is the first inversion-free editing method
that bypasses the costly inversion process of generative models. We prove that the editing trajectory of DRFS reduces to FlowEdit~\cite{kulikov2024floweditinversionfreetextbasedediting} under the Rectified Flow parameterization $(a_t,b_t) =(1-t,t)$ and shift coefficient $c_t=t$.
To avoid notational confusion, we denote the editing trajectory as $x_0^{tgt}(t), t \colon 1 \to 0$. Hence, $x_0^{\text{tgt}}(1) = x_0^{\text{src}}$ and $x_0^{\text{tgt}}(0)$ correspond to the initial source image and the final edited result, respectively. FlowEdit evolves a given source image $x_0^{\text{tgt}}=x_0^{\text{src}}$ over time $t \colon 1 \to 0$ with the following dynamics. 
{\small
\begin{align} \label{eq:flowedit}
 dx_{0}^{\text{tgt}}(t)=
\Bigl[
  v_\theta\!\bigl(
    x_{0}^{\text{tgt}}(t)
    + x_{t}^{\text{src}}
    - x_{0}^{\text{src}},\,
    t,\,
    \varphi^{\text{tgt}}
  \bigr) -\,v_\theta\!\bigl(
      x_{t}^{\text{src}},\,t,\,
      \varphi^{\text{src}}
    \bigr)
\Bigr]\,dt.   
\end{align}}

At a high level, the design choice in FlowEdit ensures that the term $(x_{0}^{tgt}(t) + x_t^{src} - x_0^{src})$ lies within the forward posterior of the target distribution. We now show that FlowEdit can be viewed as a \emph{specific instance} of DRFS. Given the expression of the DRFS energy function (Eq.~\ref{eq:E_DRFS}), the term $(x_{0}^{tgt}(t) + x_t^{src} - x_0^{src})$ can be interpreted as a $\hat{x}_t^{tgt}$, at which the velocity with respect to $\varphi^{tgt}$ is evaluated. Interpreting this term as the DRFS latent $\hat{x}_t^{tgt}$ yields

\begin{align*}
x_{0}^{\text{tgt}}(t) + x_{t}^{\text{src}} - x_{0}^{\text{src}}
&= (1-t)\,x_{0}^{\text{tgt}}(t)
  + t\varepsilon
  + t\bigl(x_{0}^{\text{tgt}}(t) - x_{0}^{\text{src}}\bigr)
\\
&= \hat{x}_{t}^{\text{tgt}}
\;\;\Longrightarrow\;\;
\text{DRFS with } c_{t}=t.
\end{align*}

Therefore, FlowEdit is a special case of our DRFS framework under
$ (a_t,b_t,c_t) = (1-t,t,t)$, along with a descending timestep scheduler and a specific learning rate. Therefore, from this viewpoint, an ordinary differential equation \eqref{eq:flowedit} is interpreted as the flow that minimizes a specific energy function. Our formulation provides richer optimization possibilities and greater flexibility: the novel time-dependent shift coefficient we introduce yields a gradual semantic shift that prevents error amplification in early, high-noise steps and follows a principled, well-defined derivation.

\subsection{Trajectory analysis of DRFS}
\label{subsec:straight}
\begin{figure}[h]
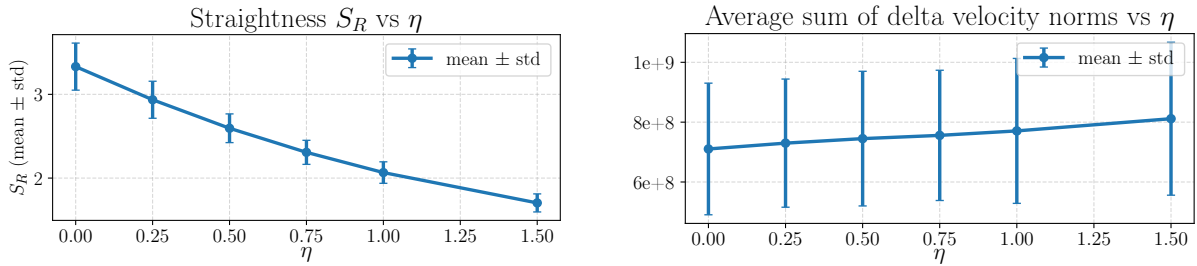

  \centering
  \begin{subfigure}[b]{0.48\linewidth}
    \centering
    \resizebox{\linewidth}{!}{\input{figures/S_R_vs_eta3.pgf}}
    \vspace{-7mm}
    \caption{$S_R$ (lower $\!\!\to$ straighter)}

    \label{fig:subfig:S_R}
  \end{subfigure}
  \hfill
  \begin{subfigure}[b]{0.48\linewidth}
    \centering
    \resizebox{\linewidth}{!}{\input{figures/vnorm_sum_vs_etacar_14.pgf}}
    \vspace{-7mm}
    \caption{ $\sum_t\|v_\theta(\hat{x}_t^{{tgt}})-v_\theta(x_t^{{src}})\|^2$ } %
    \label{fig:subfig:grad_norms_eta}
  \end{subfigure}

  \vspace{-0.3em}
  \caption{
  Effect of the offset coefficient $c_t$. Subfigure (a) shows that larger $\eta$ yields straighter trajectories; (b) shows it also increases update magnitude via amplified $\|v_\theta(\hat{x}_t^{{tgt}})-v_\theta(x_t^{{src}})\|^2$.
  }
  \label{fig:DRFS_straightness_and_updates}
\end{figure}

In this section, we empirically demonstrate that $c_t >0$ results in straighter editing paths and larger updates, taking the example of $c_t=\eta t$. Hence, a large $\eta$ corresponds to a large $c_t$. To quantify the straightness of a given editing path $\{x_{0,k}^{tgt}\}_{k=0}^{N}$, we define the path–to–chord ratio $S_R = \sum_{k=0}^{N-1} \|x_{0,k+1}^{tgt}-x_{0,k}^{tgt}\| / \| x_{0,N}^{tgt}-x_{0,0}^{tgt} \|$. $S_R=1$ stands for a perfectly straight path and increases as a path becomes less straight.
As shown in Fig. \ref{fig:DRFS_straightness_and_updates}(a), the DRFS editing path becomes straighter as $\eta$ increases. %
In addition, we reveal that DRFS produces larger update $\|v_{\theta}(\hat{x}_{t}^{tgt}) - v_{\theta}(x_{t}^{src})\|$ in Fig. \ref{fig:DRFS_straightness_and_updates}(b),  meaning that $v_\theta(\hat x_t^{tgt})$ direction differentiates more from $v_\theta( x_t^{src})$ when $c_t$ is larger. This suggests that larger values of $c_t$ steer the optimization more aggressively toward the target prompt, potentially accelerating the editing process.
\textit{Taking an intermediate value for $c_t$ %
allows us to control editing strength and straightness of the editing path}, in order to achieve both good alignment with the target prompt and a high level of fidelity. %

We further detail our choice of the $c_t$ coefficient in the Implementation Details section and in Appendix~\ref{app:shift}. We show empirically in Table ~\ref{tab:abl_ct_pie_results}, and theoretically  that both DDS and FlowEdit (implicit) shift choices are suboptimal.

\section{Experiments}

\subsection{Baselines and Implementation Details}
\paragraph{Baselines.} We compare our method against a range of representative baselines. Diffusion-based methods include PnP-Inv~\cite{ju2023directinversionboostingdiffusionbased}, P2P~\cite{hertz2022prompttopromptimageeditingcross}, and Null-text Inv~\cite{mokady2022nulltextinversioneditingreal}, which rely on pre-trained diffusion models with different editing strategies. Rectified flow-based methods include RF-Inv~\cite{rout2025semantic}, RF-Solver~\cite{wang2025taming}, FireFlow~\cite{deng2025fireflow}, FlowEdit~\cite{kulikov2024floweditinversionfreetextbasedediting}, and FTEdit~\cite{xu2025unveilinversioninvarianceflow}, DNAEdit ~\cite{xie2025dnaedit}, which perform editing via velocity field integration or inversion heuristics.
Among these, iRFDS~\cite{yang2025texttoimage} is the only RF distillation-based %
baseline.

\paragraph{Implementation details.} 

We employ widely adopted Rectified Flow models, namely the Stable Diffusion series (SD3 and SD3.5)~\cite{esser2024scalingrectifiedflowtransformers}. An SGD optimizer and a descending time-step schedule are used. Following our analysis (in Appendix~\ref{app:shift}), we use an intermediate shift coefficient that increases from $0$ to $1$ throughout optimization,  resulting in $c_t =\frac{k}{T}t \simeq(1-t)t$ where $k$ and $T$ are the current step and the number of total steps, respectively. \emph{We observed that this gradual shift helps avoid error amplification in early, noisy steps}, while maintaining background details. %
Gradients are computed using a single sampled time step ($\text{batch size}=1$), with source and target CFG values respectively set to 6 and 16.5, and with unit weighting as in previous distillation-based methods. %
Please refer to the Appendix for more details.

\newcommand{\second}[1]{\underline{#1}}   %

\begin{table*}[t]
    \caption{\textbf{Quantitative comparison on the PIE benchmark.} The best and second-best results are shown in bold and underlined, respectively.}
    \label{tab:pie_results}
    \vspace{-2mm}
    \centering
    \resizebox{\textwidth}{!}{%
    \setlength{\tabcolsep}{4.6pt}
    \begin{tabular}{l|c|c|c|cccc|cc}
        \toprule
        \multirow{2}{*}{\textbf{Method}} & \multirow{2}{*}{\textbf{Model}} & \multicolumn{2}{c|}{\textbf{Structure}} & \multicolumn{4}{c|}{\textbf{Background Preservation}} & \multicolumn{2}{c}{\textbf{CLIP Similarity}} \\
        \cmidrule(lr){3-10}
        & & Editing & Distance $_{\times 10^3} \downarrow$ & PSNR $\uparrow$ & LPIPS $_{\times 10^3} \downarrow$ & MSE $_{\times 10^4} \downarrow$ & SSIM{$_{\times 10^2} \uparrow$} & Whole $\uparrow$ & Edited $\uparrow$  \\
        \midrule
        \textbf{DDIM~\cite{song2021denoising}}& Diffusion & P2P & 69.4 & 17.87 & 208.80 & 219.88 & 71.14 & 25.01 & 22.44 \\
        \textbf{DDIM~\cite{song2021denoising}} & Diffusion & PnP & 28.22 & 22.28 & 113.46 & 83.64 & 79.05 & \best{25.41} & \second{22.55} \\
        \textbf{Null-Text~\cite{mokady2022nulltextinversioneditingreal}} & Diffusion & P2P & \second{13.44} & \second{27.03} & \second{60.67} & \second{35.86} & 84.11 & 24.75 & 21.86 \\
        \textbf{PnP-Inv~\cite{ju2023directinversionboostingdiffusionbased}} & Diffusion & P2P & \best{11.65} & \best{27.22} & \best{54.55} & \best{32.86} & 84.76 & \second{25.02} & 22.10 \\
        \textbf{PnP-Inv~\cite{ju2023directinversionboostingdiffusionbased}} & Diffusion & PnP & 24.29 & 22.46 & 106.06 & 80.45 & 79.68 & \best{25.41} & \best{22.62} \\
        \midrule
        \textbf{RF-Inv~\cite{rout2025semantic}} & Flux & -  & 40.6  & 20.82  & 184.8  & 129.1   & 71.92  & 25.20  & 22.11  \\
        \textbf{RF-Solver~\cite{wang2025taming}}   & Flux & RF-Solver & 31.1 & 22.90 & 135.81 & 80.11 & 81.90 & 26.00 & 22.88 \\
        \textbf{FlowChef~\cite{patel2024steeringrectifiedflowmodels}} & Flux & - & 34.50 & 22.75 & 211.12 & 71.72 & 72.81 & 23.97 & 21.40 \\
        \textbf{FireFlow~\cite{deng2025fireflow}}    & Flux & RF-Solver & 28.3 & 23.28 & 120.82 & 70.39 & 82.82 & 25.98 & 22.94 \\
        \textbf{FlowEdit~\cite{kulikov2024floweditinversionfreetextbasedediting}}   & Flux & - & 27.7 & 21.91 & 111.70 & 94.0 & 83.39 & 25.61 & 22.70 \\
        \textbf{DNAEdit} & Flux & - & 18.87 & 24.99 & 95.06 & 50.45 & 85.71 & 25.79 & 22.87 \\
        \textbf{FlowEdit~\cite{kulikov2024floweditinversionfreetextbasedediting}} & SD3 & - & 27.24 & 22.13 & 105.46 & 87.34 & 83.48 & \second{26.83} & \second{23.67} \\
        \textbf{iRFDS~\cite{yang2025texttoimage}} & SD3 & - & 62.72 & 19.61 & 186.39 & 179.76 & 74.59 & 24.54 & 21.67 \\
        \textbf{FlowEdit~\cite{kulikov2024floweditinversionfreetextbasedediting}} & SD3.5 & - & \second{12.73} & \second{26.59} & \second{56.17} & \second{33.84} & 89.34 & 26.31 & 23.00 \\
        \textbf{FTEdit~\cite{xu2025unveilinversioninvarianceflow}} & SD3.5 & AdaLN & 18.17 & 26.62 & 80.55 & 40.24 & \best{91.50} & 25.74 & 22.27 \\
        \textbf{DNAEdit~\cite{xie2025dnaedit}} & SD3.5 & - & 14.19 & 26.66 & 74.57 & 32.76 & 88.63 & 25.63 & 22.71 \\
        \rowcolor{gray!10}
        \textbf{DRFS (Ours)} & SD3 & - & {23.05} & {23.38} & {93.81} & {67.49} & {84.85} & \best{26.90} & \textbf{23.83} \\
        \rowcolor{gray!10}
        \textbf{DRFS (Ours)} & SD3.5 & - & \best{12.00} & \best{26.97} & \best{55.83} & \best{30.76} & \second{89.41} & 26.45 & 23.17 \\
        \bottomrule
    \end{tabular}
    }
\end{table*}

\paragraph{Evaluation datasets and metrics.}
We evaluate DRFS on the PIE benchmark~\cite{ju2023directinversionboostingdiffusionbased}, which comprises 700 diverse images spanning various editing tasks. For assessing reconstruction quality and background preservation, we report image-level metrics: LPIPS~\cite{lpips}, SSIM~\cite{ssim}, MSE, PSNR, and structure distance~\cite{ju2023directinversionboostingdiffusionbased}. To measure semantic alignment with the target prompt, we use CLIP similarity~\cite{pmlr-v139-radford21a}. Additional results are provided in the Appendix, where we evaluate on an additional dataset composed of 300 photos from~\cite{8014884} and a collection of stock images from~\cite{pixabay2025}. 
Captions and editing prompts are generated with Qwen VL-7B %
~\cite{wang2024qwen2vlenhancingvisionlanguagemodels}, and we evaluate DRFS under varying CFG scales to demonstrate robustness. This additional dataset is also used in Figure~\ref{fig:quali}.

\setlength{\tabcolsep}{3.5pt}                  %
\newlength{\ThumbSize}
\setlength{\ThumbSize}{%
  \dimexpr(\textwidth - 14\tabcolsep)/8\relax}

\newcolumntype{M}{>{\centering\arraybackslash}m{\ThumbSize}}

\newcommand{\Thumb}[1]{\includegraphics[width=\ThumbSize]{#1}}

\begin{figure*}[t]
  \centering
  \renewcommand{\arraystretch}{0.68}

  \newcommand{\aboveLegend}{\\[-0.18em]}
  \newcommand{\belowLegend}{\\[0.25em]}

  \begin{tabular}{@{}*{8}{M}@{}}          %
    \scriptsize Source &
    \scriptsize DRFS &
    \scriptsize FlowEdit (SD3) &
    \scriptsize iRFDS &
    \scriptsize FireFlow &
    \scriptsize RF‑Solver &
    \scriptsize RF‑Inv &
    \scriptsize Null‑Text \\[0.05em]

    \Thumb{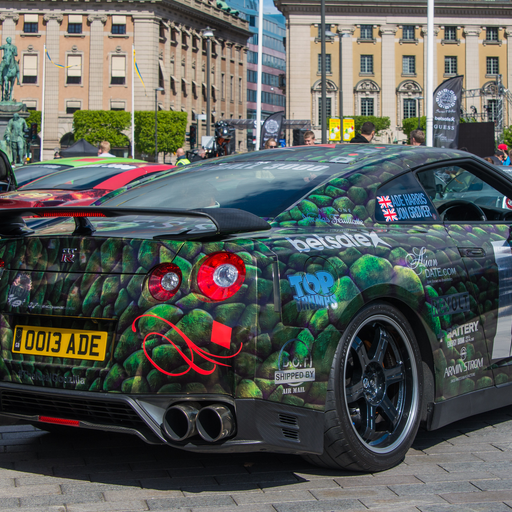} &
    \Thumb{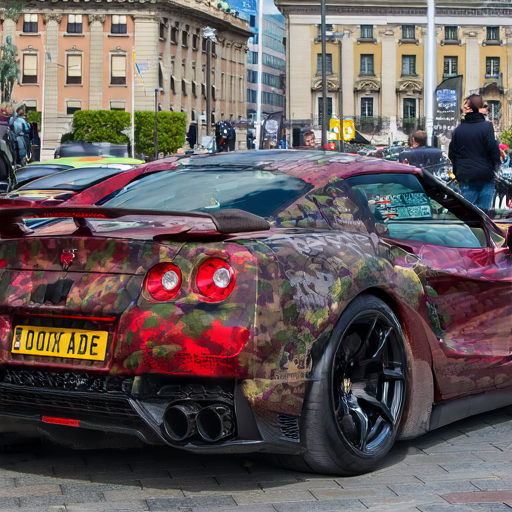} &
    \Thumb{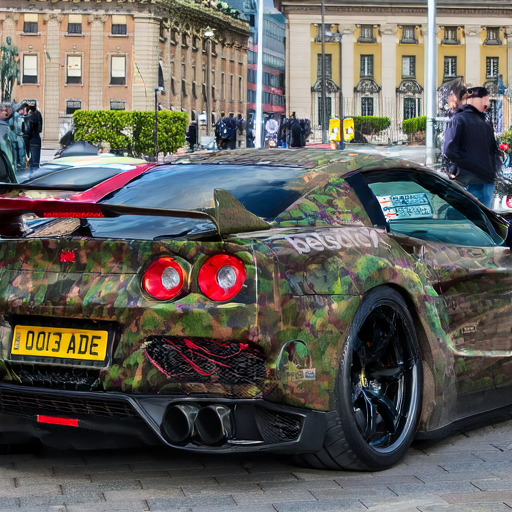} &
    \Thumb{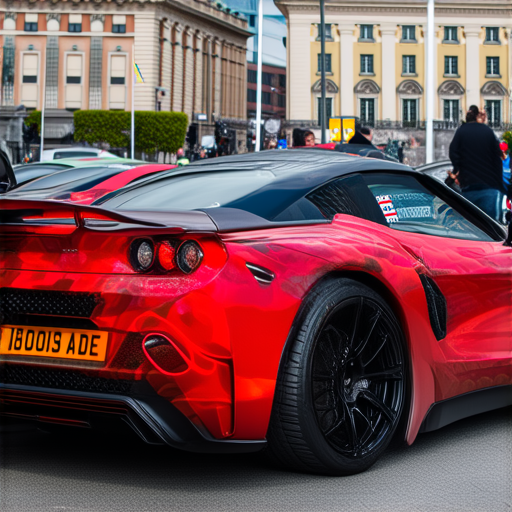} &
    \Thumb{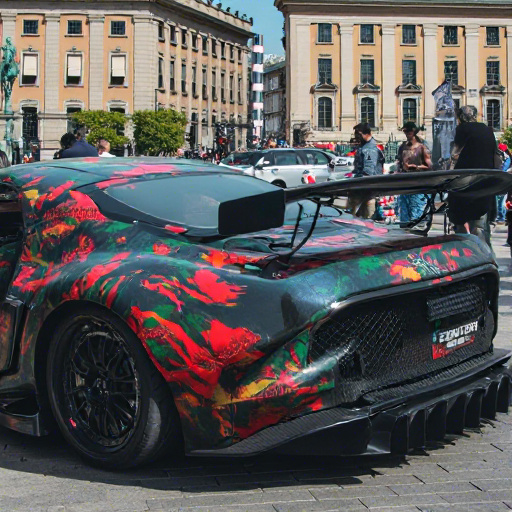} &
    \Thumb{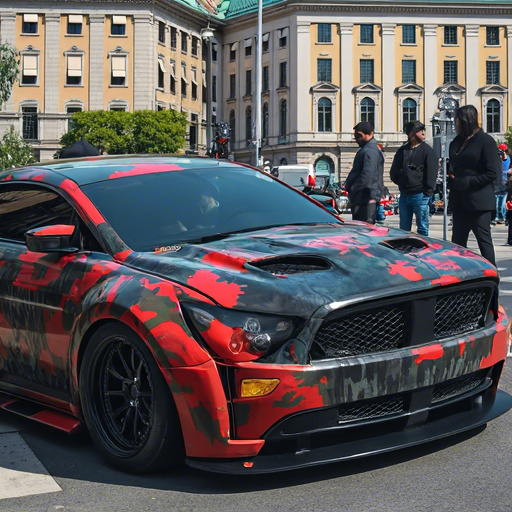} &
    \Thumb{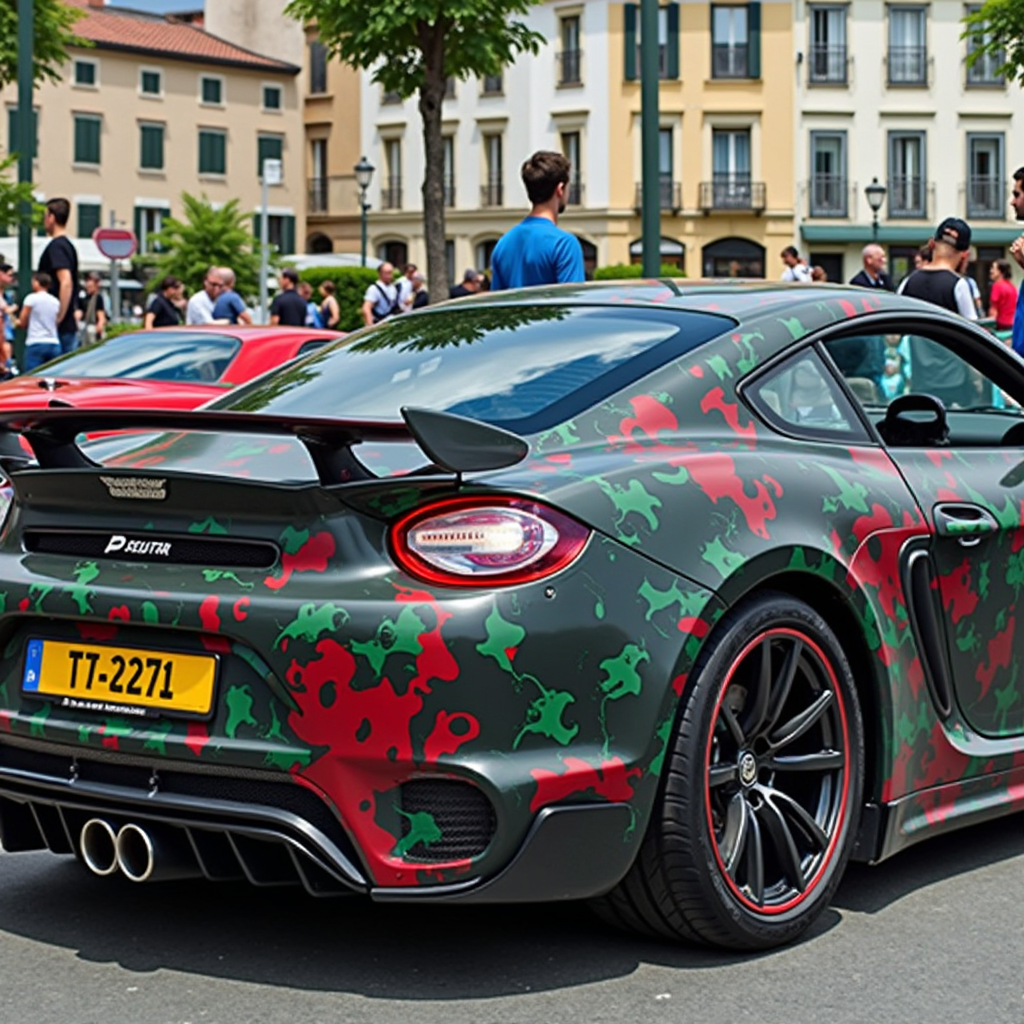} &
    \Thumb{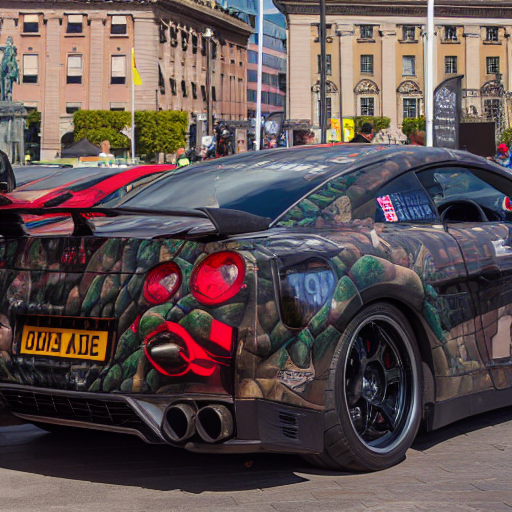}
    \aboveLegend
    \multicolumn{8}{c}{\scriptsize green camouflage $\rightarrow$ red camouflage}
    \belowLegend

    \Thumb{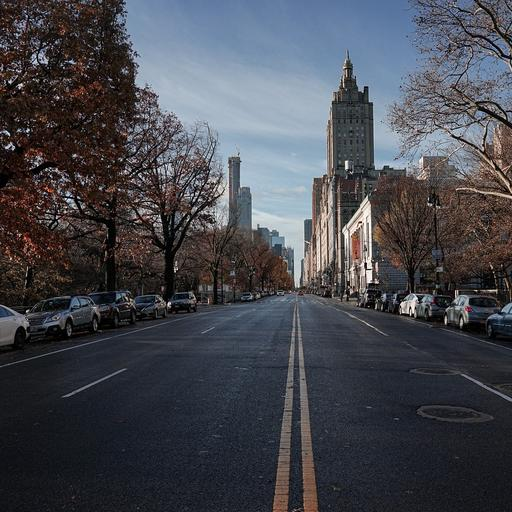} &
    \Thumb{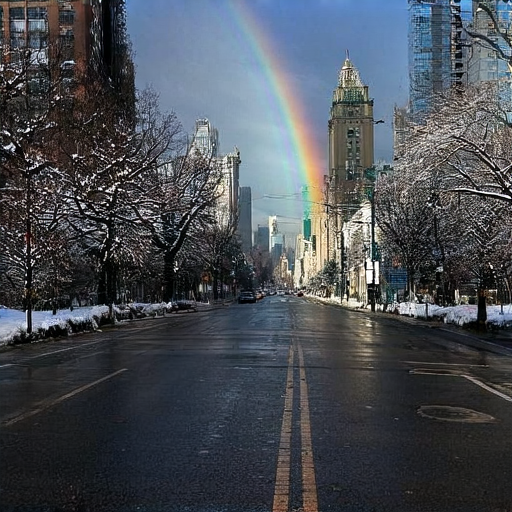} &
    \Thumb{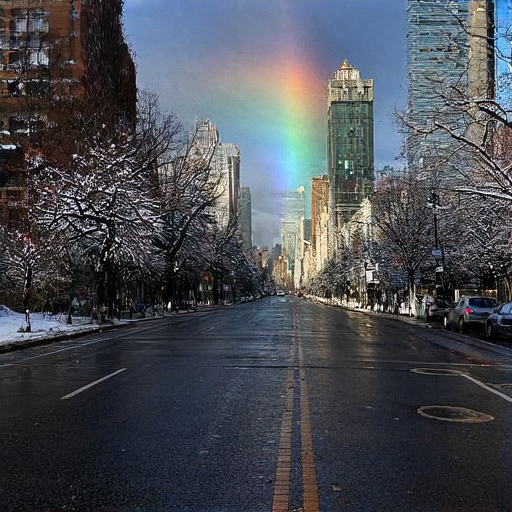} &
    \Thumb{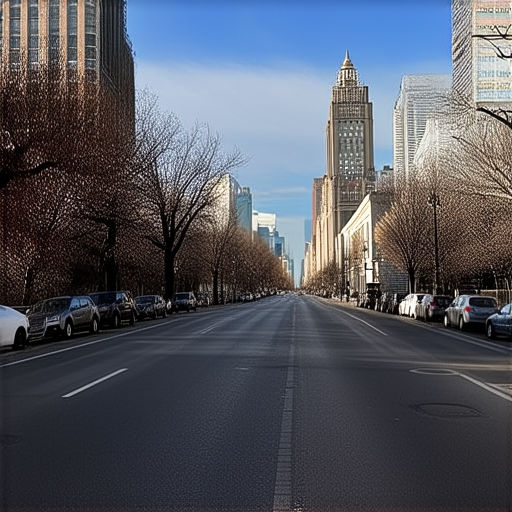} &
    \Thumb{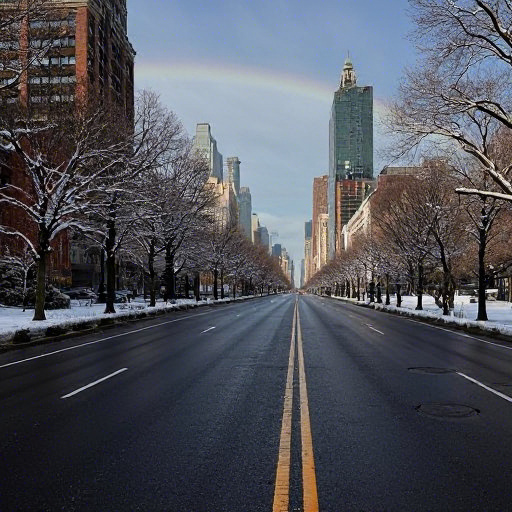} &
    \Thumb{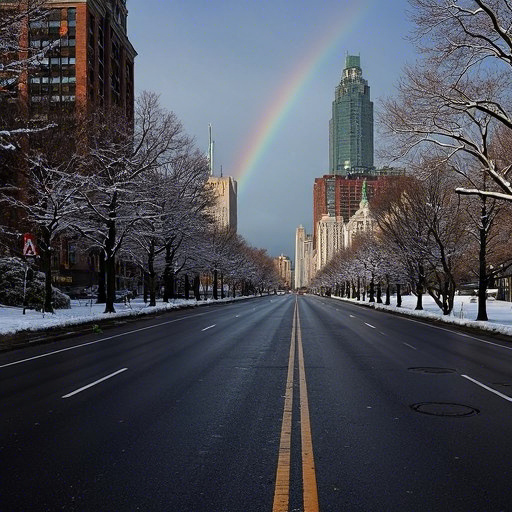} &
    \Thumb{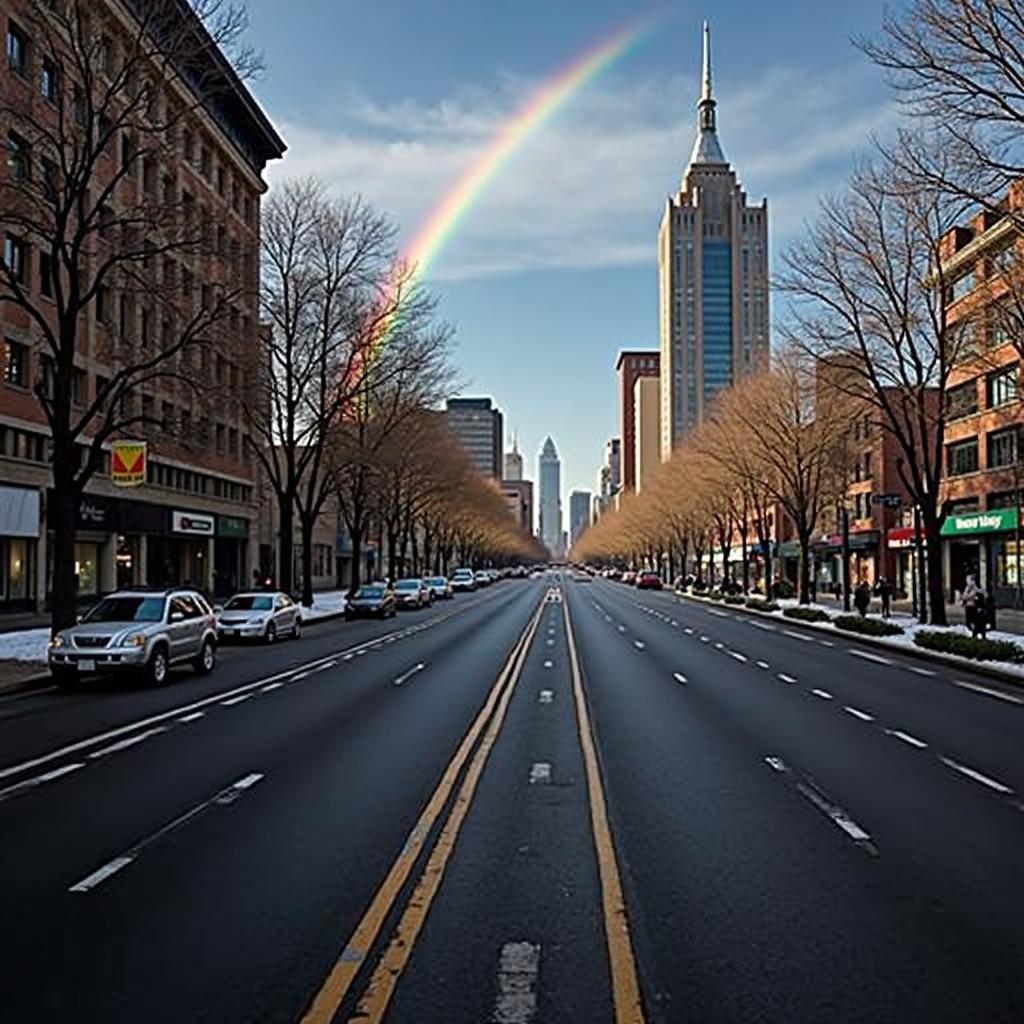} &
    \Thumb{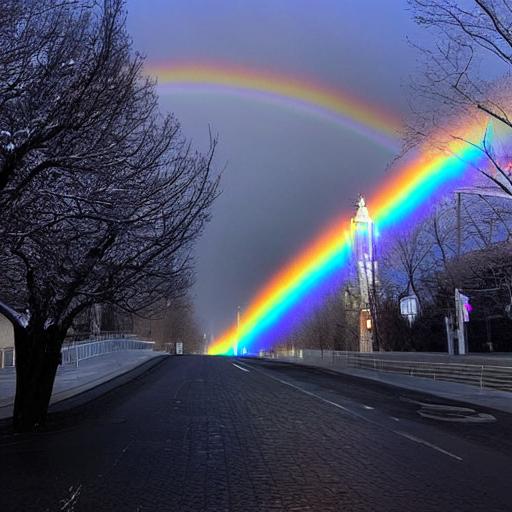}
    \aboveLegend
    \multicolumn{8}{c}{\scriptsize + rainbow and autumn $\rightarrow$ winter}
    \belowLegend

    \Thumb{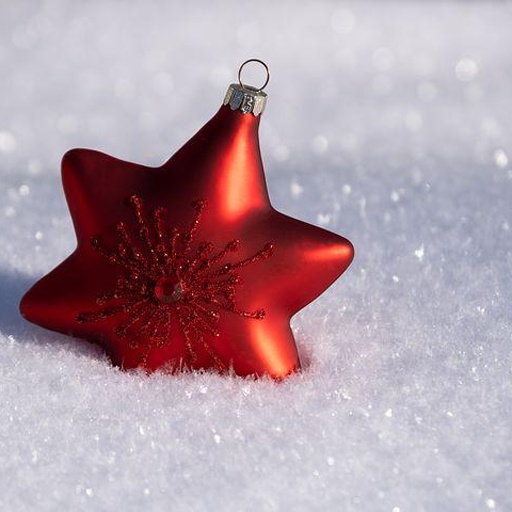} &
    \Thumb{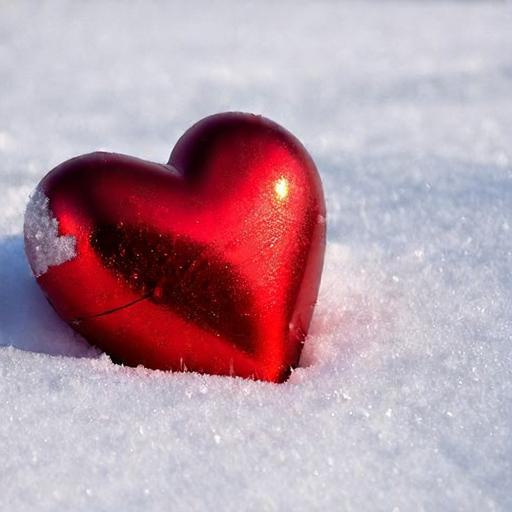} &
    \Thumb{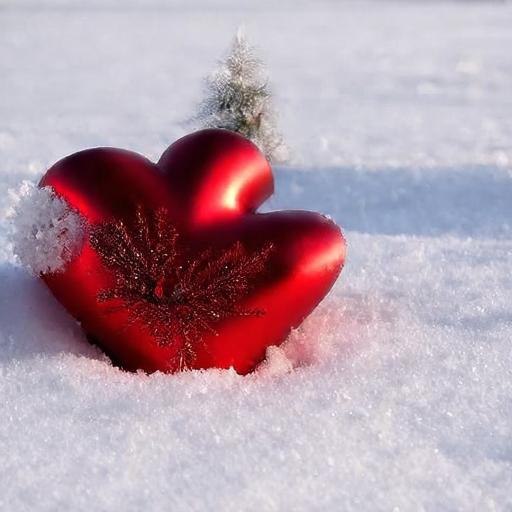} &
    \Thumb{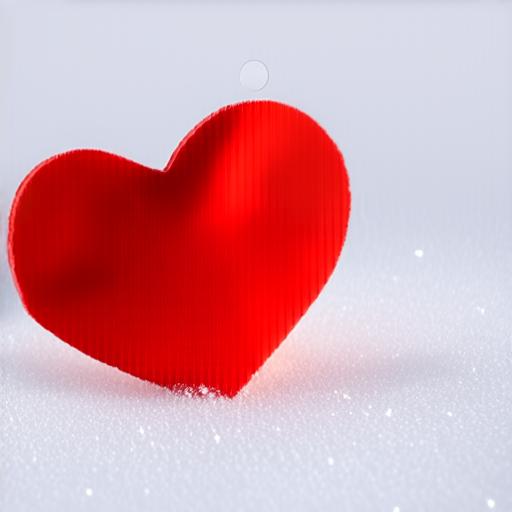} &
    \Thumb{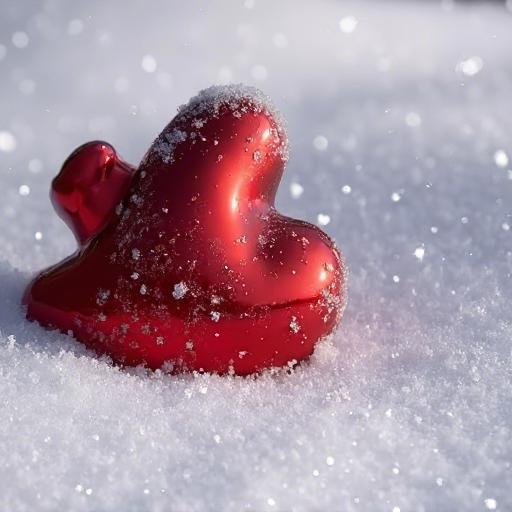} &
    \Thumb{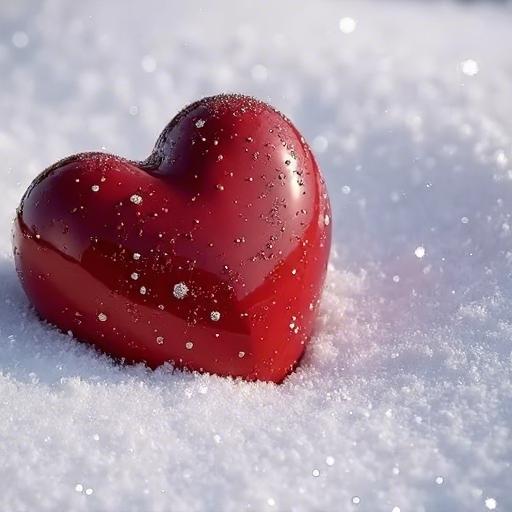} &
    \Thumb{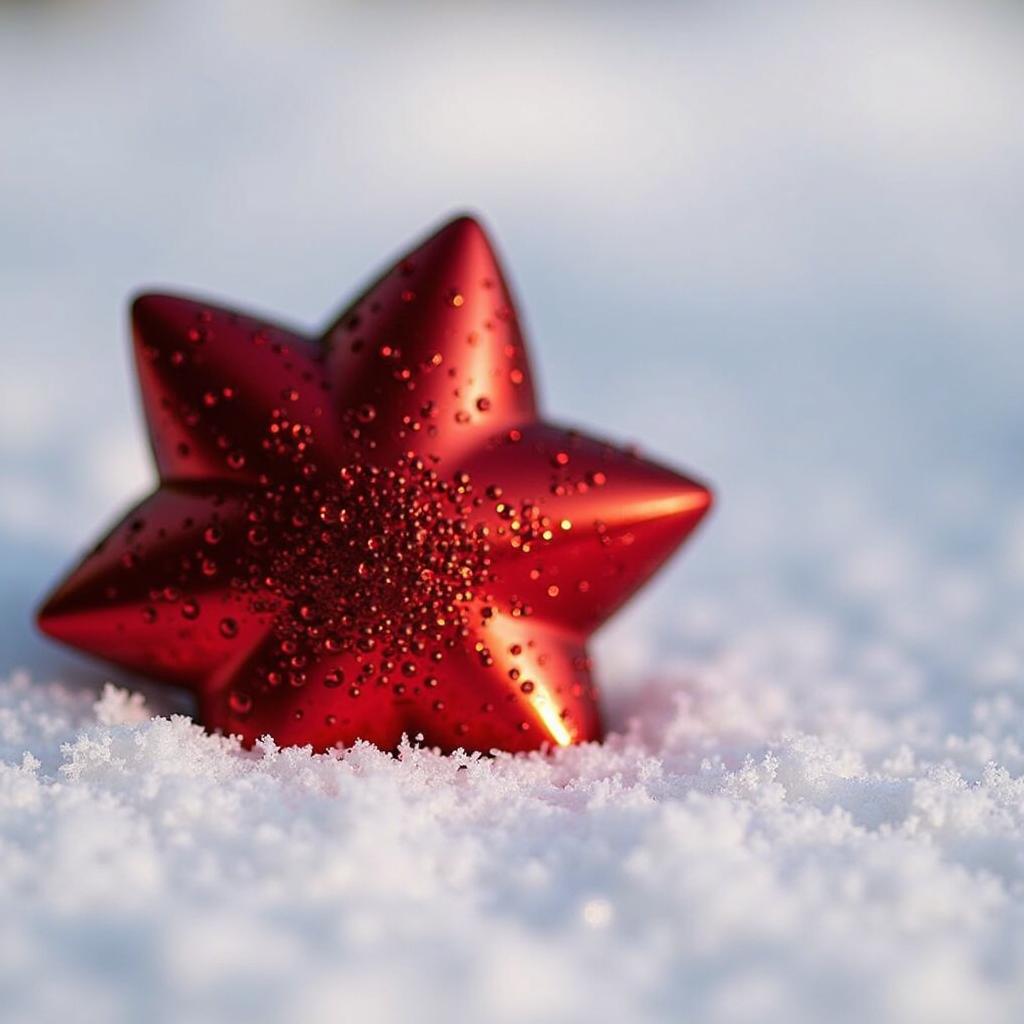} &
    \Thumb{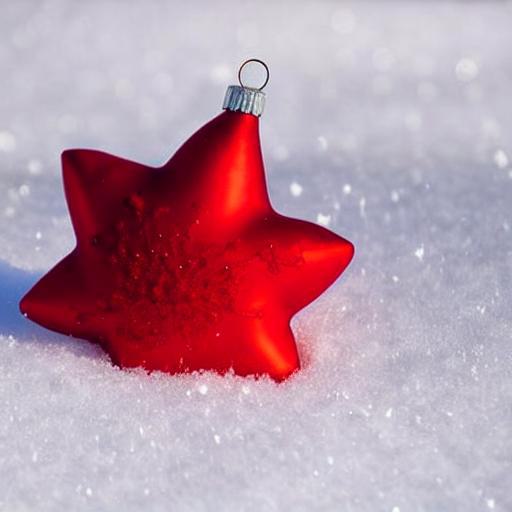}
    \aboveLegend
    \multicolumn{8}{c}{\scriptsize star $\rightarrow$ heart}
    \belowLegend

    \Thumb{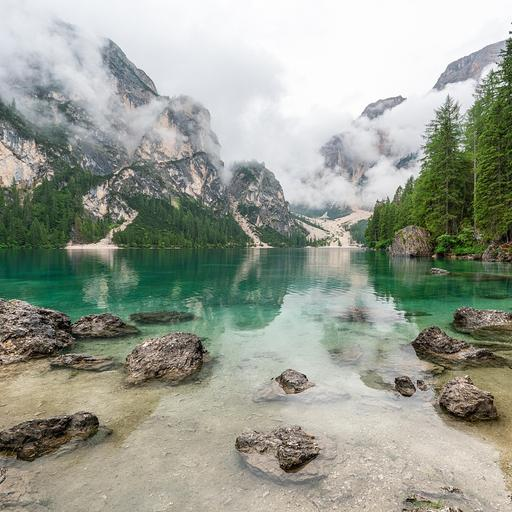} &
    \Thumb{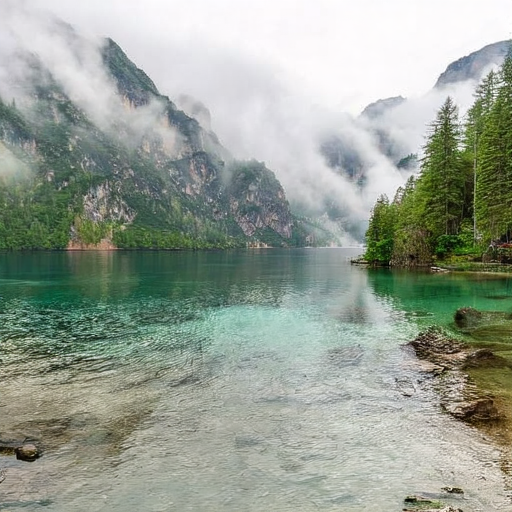} &
    \Thumb{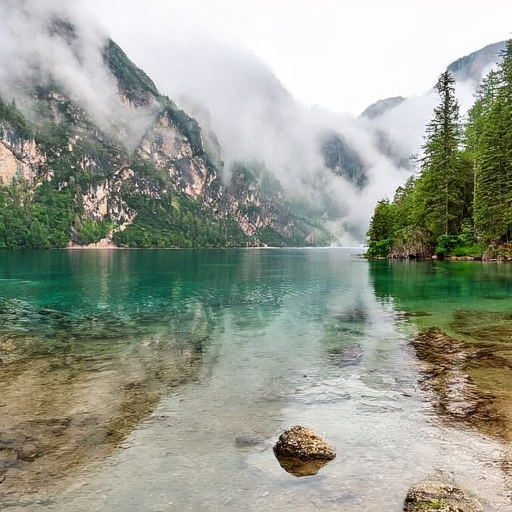} &
    \Thumb{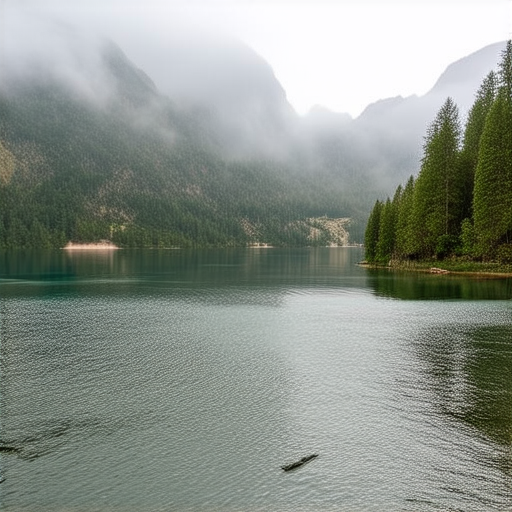} &
    \Thumb{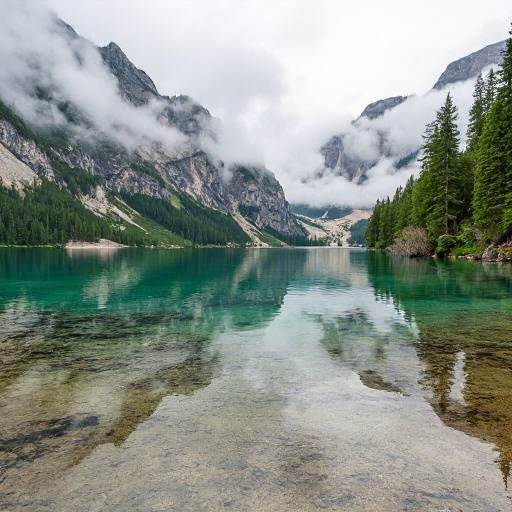} &
    \Thumb{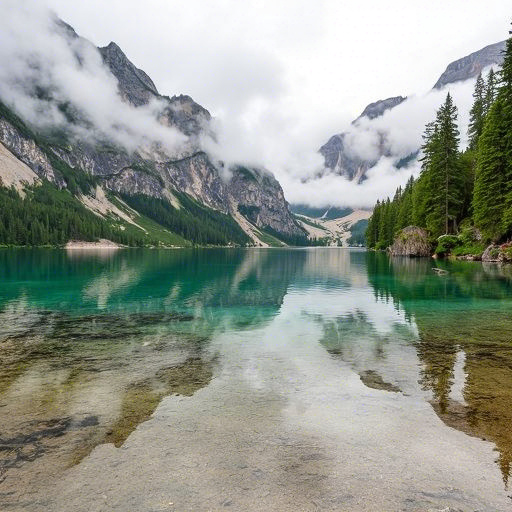} &
    \Thumb{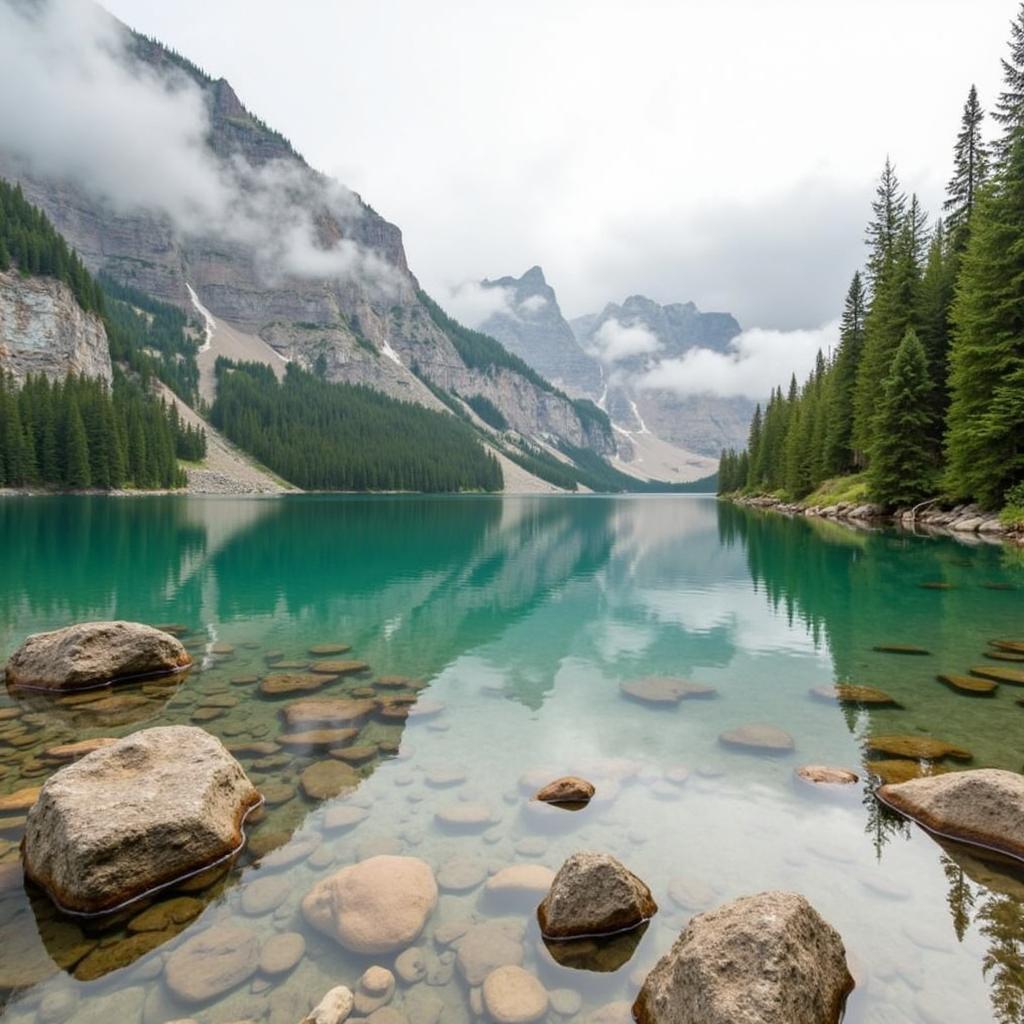} &
    \Thumb{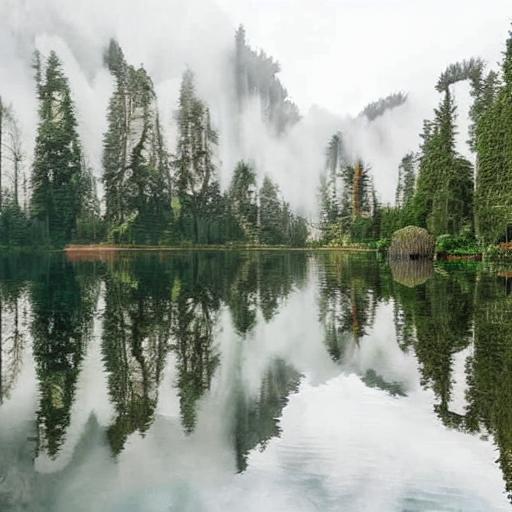}
    \aboveLegend
    \multicolumn{8}{c}{\scriptsize – with stones}
    \belowLegend

    \Thumb{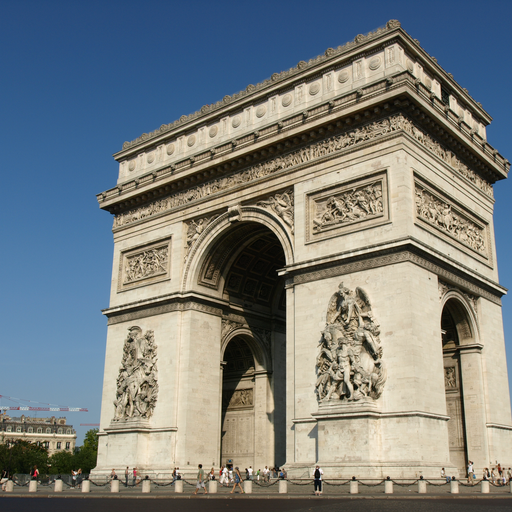} &
    \Thumb{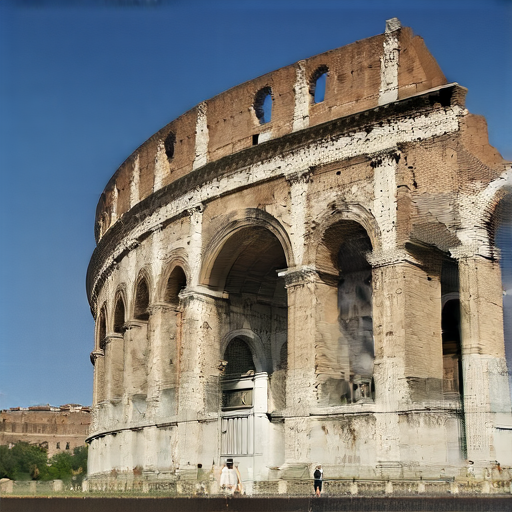} &
    \Thumb{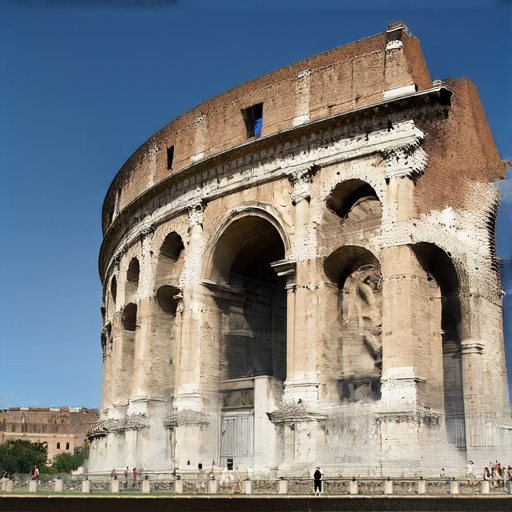} &
    \Thumb{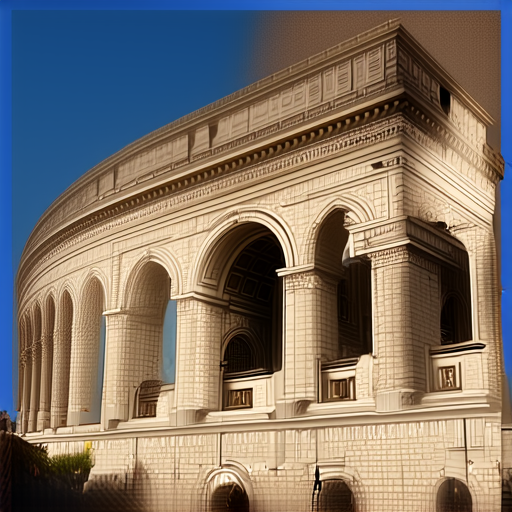} &
    \Thumb{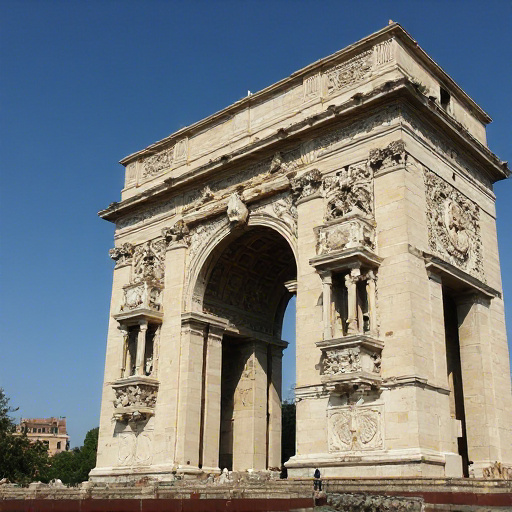} &
    \Thumb{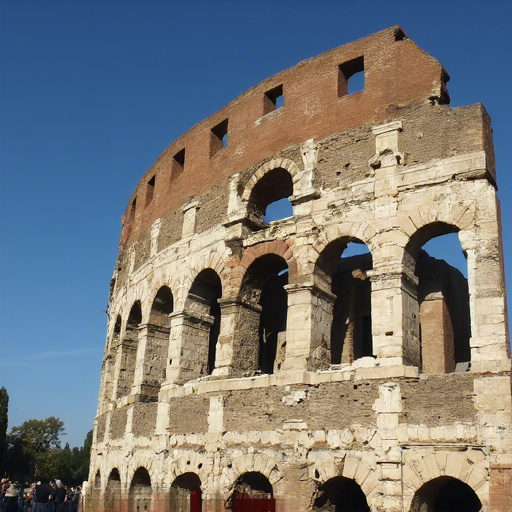} &
    \Thumb{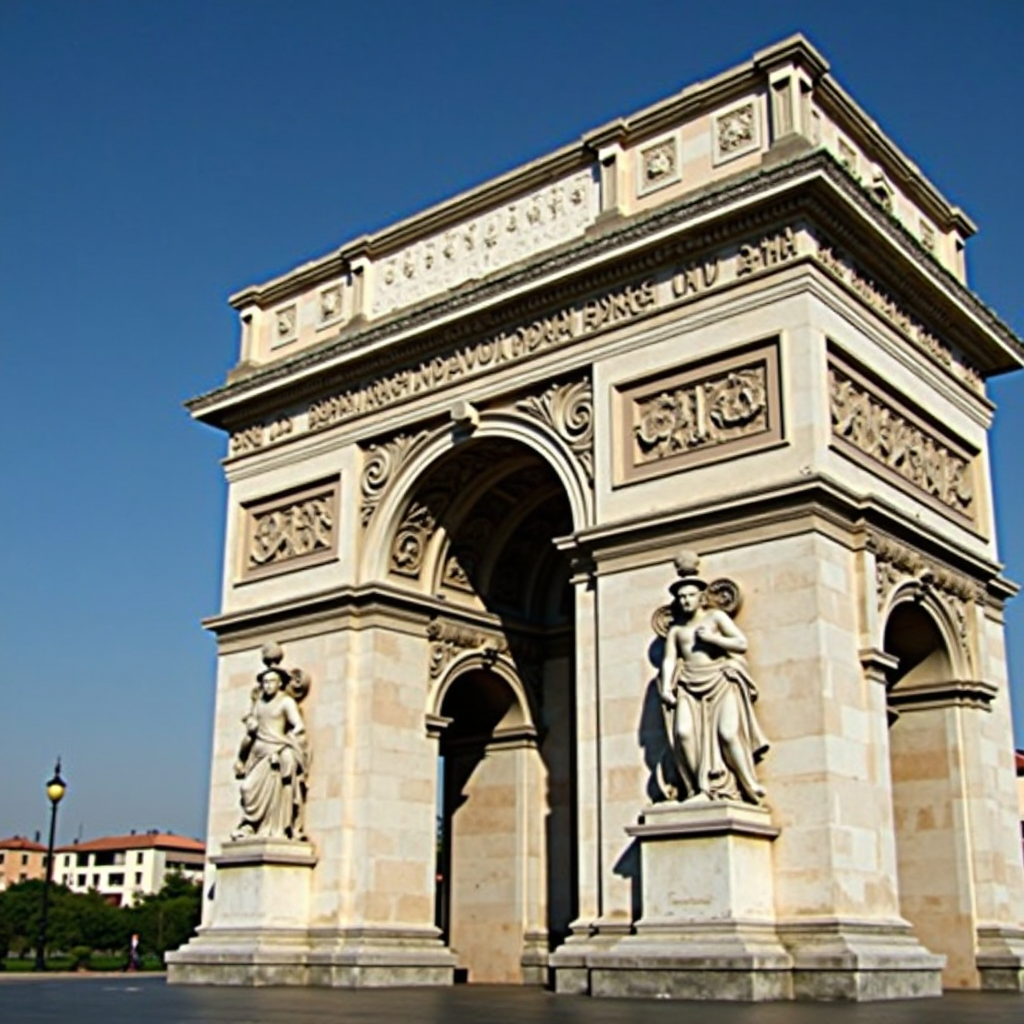} &
    \Thumb{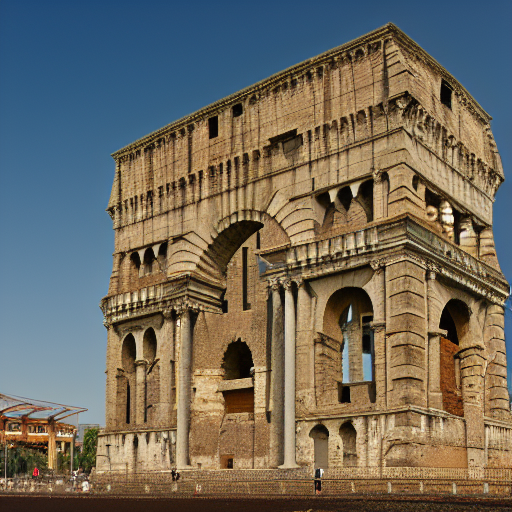}
    \aboveLegend
    \multicolumn{8}{c}{\scriptsize Arc‑de‑Triomphe $\rightarrow$ Colosseum}
    \belowLegend

  \end{tabular}
  \vspace{-1em}
  \caption{\textbf{Qualitative comparisons on images from our additional dataset and PIE benchmark.}}
  \label{fig:quali}
\end{figure*}

\subsection{Main Results}

\begin{figure*}[t]
  \centering

  \begin{subfigure}[t]{0.33\linewidth}
    \centering
    \begin{minipage}{\linewidth}
      \centering
      \includegraphics[width=0.49\linewidth]{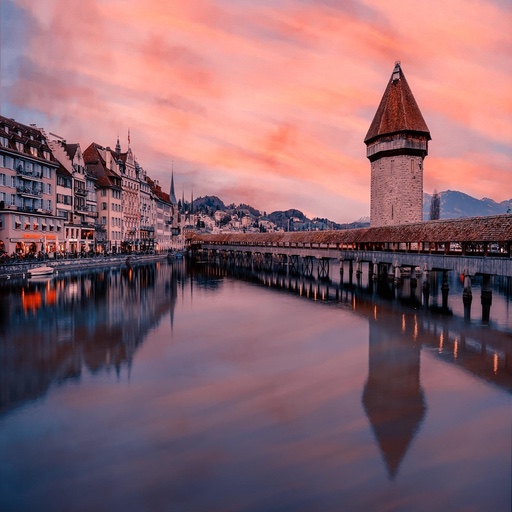}%
      \hspace{1pt}%
      \includegraphics[width=0.49\linewidth]{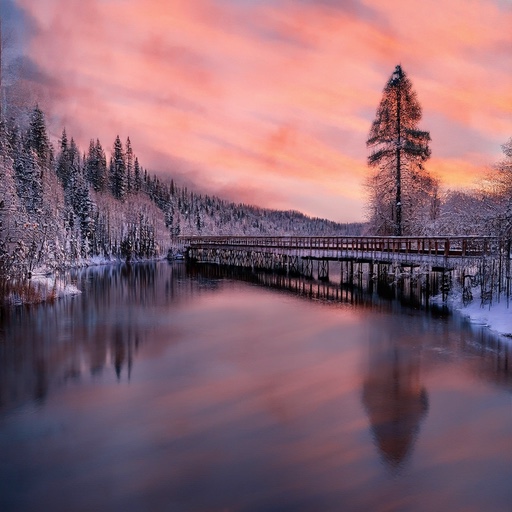}\\[2pt]
      \textbf{+ snow forest}
    \end{minipage}
  \end{subfigure}\hfill
  \begin{subfigure}[t]{0.33\linewidth}
    \centering
    \begin{minipage}{\linewidth}
      \centering
      \includegraphics[width=0.49\linewidth]{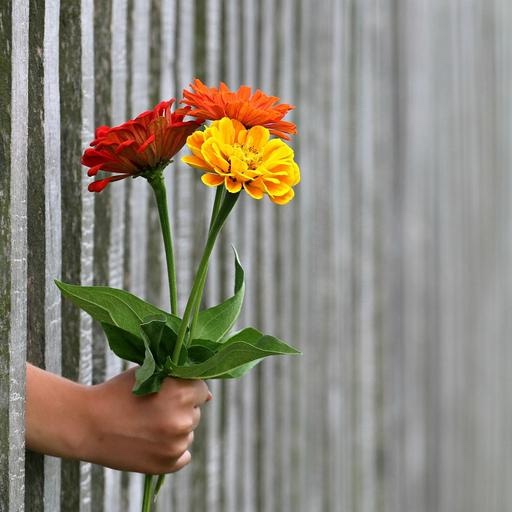}%
      \hspace{1pt}%
      \includegraphics[width=0.49\linewidth]{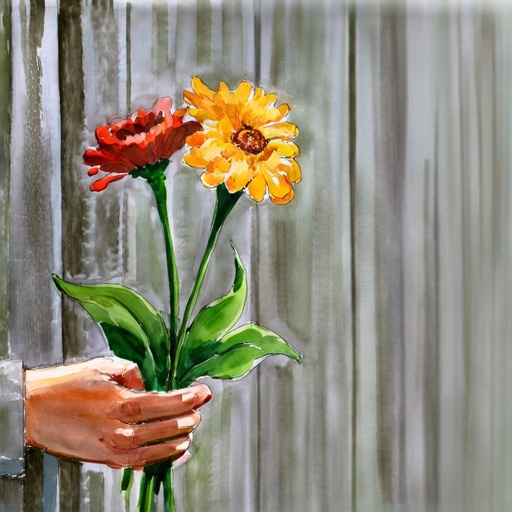}\\[2pt]
      \textbf{+ watercolor}
    \end{minipage}
  \end{subfigure}\hfill
  \begin{subfigure}[t]{0.33\linewidth}
    \centering
    \begin{minipage}{\linewidth}
      \centering
      \includegraphics[width=0.49\linewidth]{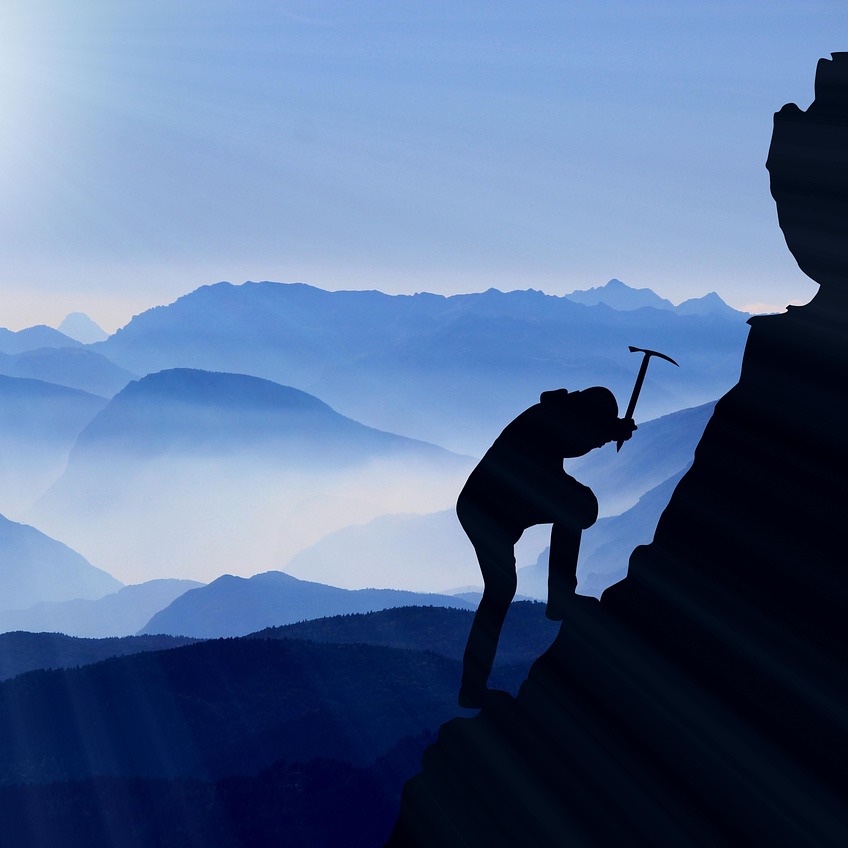}%
      \hspace{1pt}%
      \includegraphics[width=0.49\linewidth]{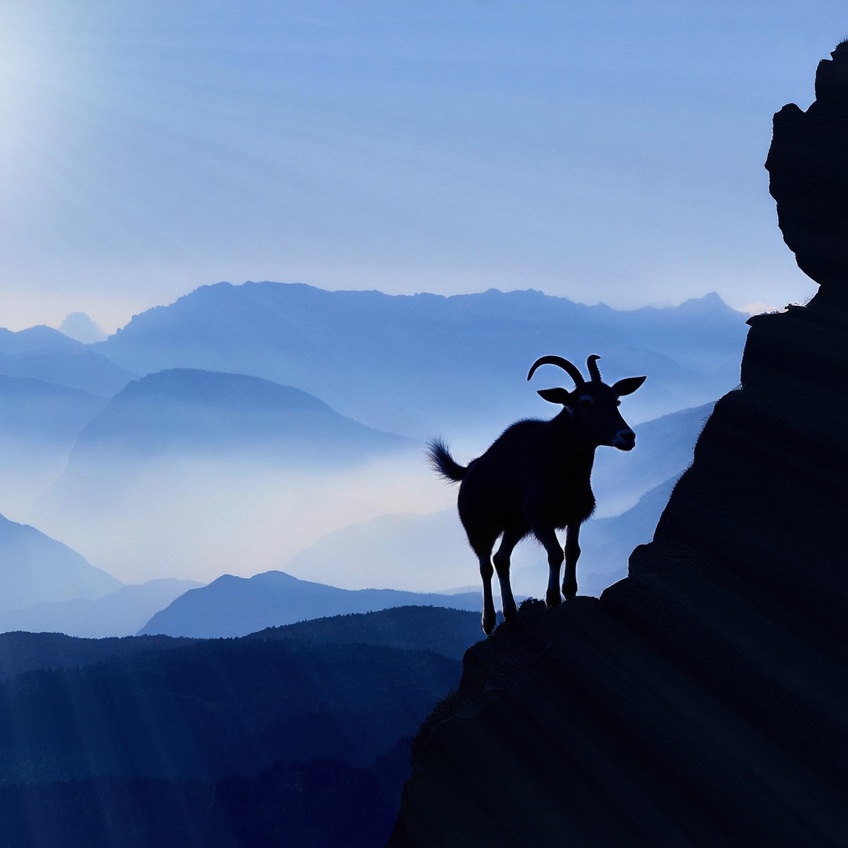}\\[2pt]
      \textbf{human $\to$ goat}
    \end{minipage}
  \end{subfigure}

  \vspace{6pt}

  \begin{subfigure}[t]{0.33\linewidth}
    \centering
    \begin{minipage}{\linewidth}
      \centering
      \includegraphics[width=0.49\linewidth]{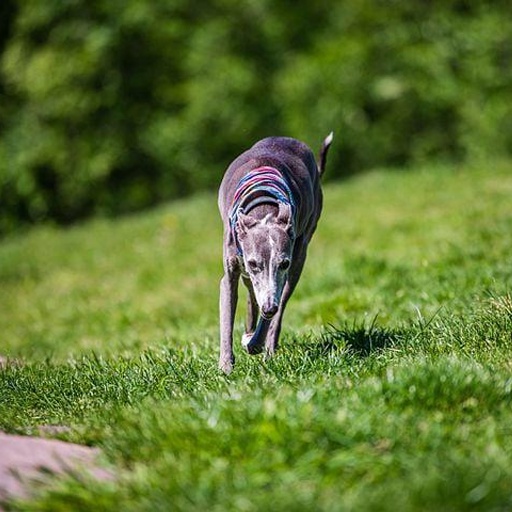}%
      \hspace{1pt}%
      \includegraphics[width=0.49\linewidth]{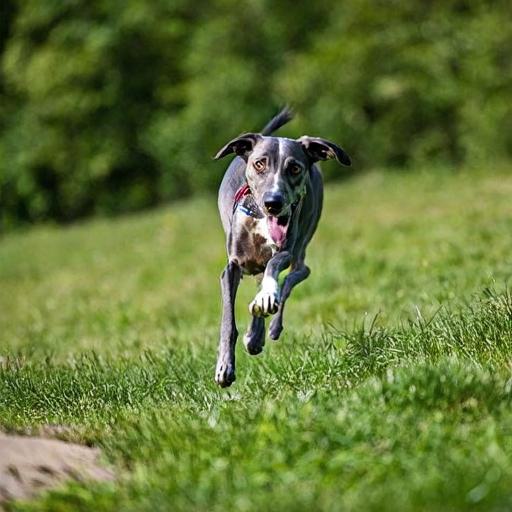}\\[2pt]
      \textbf{+ jumping}
    \end{minipage}
  \end{subfigure}\hfill
  \begin{subfigure}[t]{0.33\linewidth}
    \centering
    \begin{minipage}{\linewidth}
      \centering
      \includegraphics[width=0.49\linewidth]{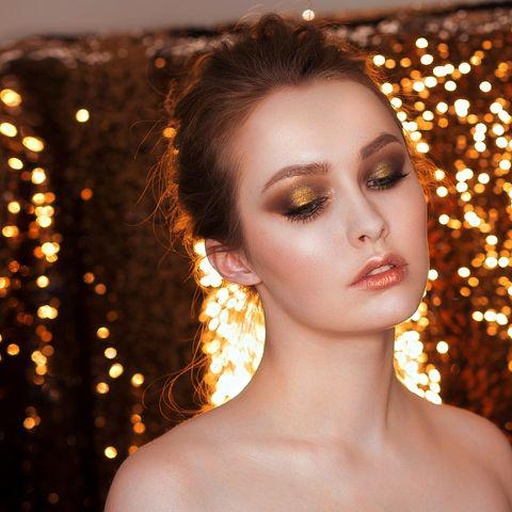}%
      \hspace{1pt}%
      \includegraphics[width=0.49\linewidth]{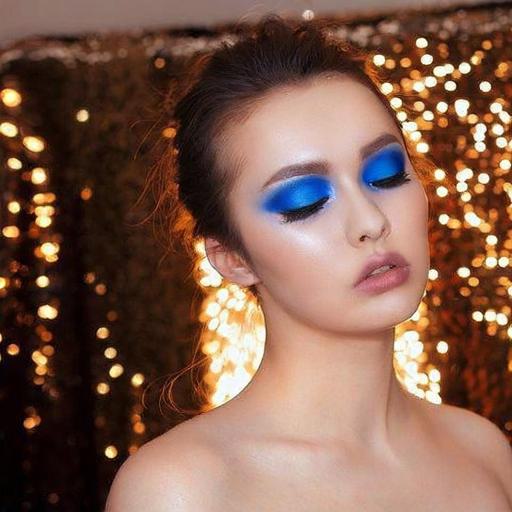}\\[2pt]
      \textbf{gold $\rightarrow$ blue}
    \end{minipage}
  \end{subfigure}\hfill
  \begin{subfigure}[t]{0.33\linewidth}
    \centering
    \begin{minipage}{\linewidth}
      \centering
      \includegraphics[width=0.49\linewidth]{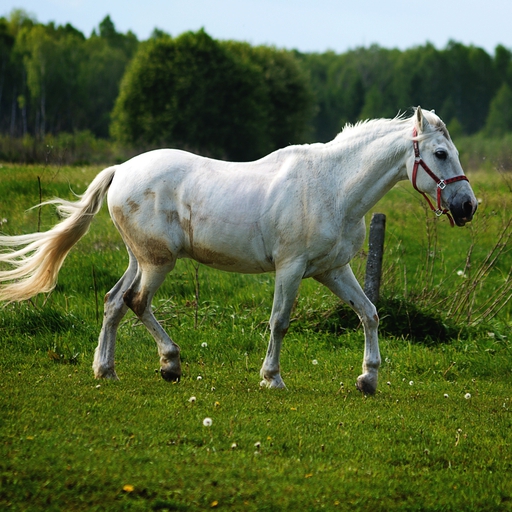}%
      \hspace{1pt}%
      \includegraphics[width=0.49\linewidth]{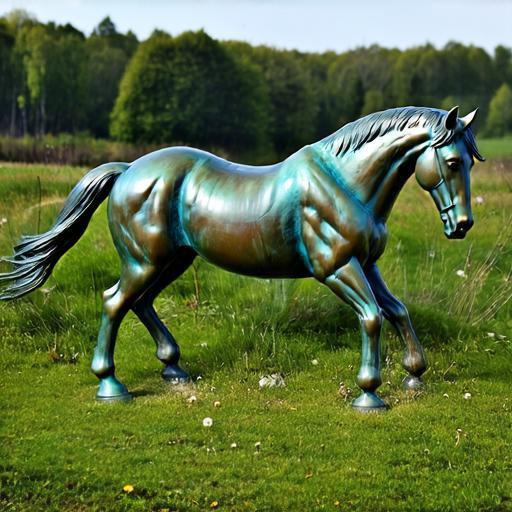}\\[2pt]
      \textbf{+ bronze}
    \end{minipage}
  \end{subfigure}

  \caption{\textbf{Qualitative edits produced by our DRFS.} Each pair indicates the source image (left) and edited result (right).}
  \label{fig:qualitative_pairs}
\end{figure*}

\paragraph{Main results.}
DRFS achieves the best overall semantic alignment, attaining the highest CLIP similarity for the edited prompts (23.83), which indicates closer adherence to the target prompt. It also surpasses RF-Inversion, FlowEdit, RF-Solver, and FireFlow on all metrics. Moreover, under the SD3 model, DRFS achieves the best structure-preservation and background-preservation performance. Compared with iRFDS, DRFS further yields more precise edits with higher visual quality while substantially improving background preservation (LPIPS: 93.81 vs.\ 186.39; MSE: 67.49 vs.\ 179.76; SSIM: 84.85 vs.\ 74.59). These results confirm that DRFS effectively reduces over-smoothing and irrelevant updates, making it the most balanced SD3-based method. Finally, on SD3.5, DRFS surpasses FlowEdit ~\cite{kulikov2024floweditinversionfreetextbasedediting} as well as the recent FTEdit and DNAEdit ~\cite{xu2025unveilinversioninvarianceflow, xie2025dnaedit} on almost all metrics.

\textbf{Qualitative results.} Figure ~\ref{fig:quali} presents editing results on challenging tasks on our additional dataset,  comparing DRFS with editing baselines. These tasks consist of color and texture changes, seasonal transformations, object removal, and large-scale landmark replacement. Across these challenging tasks, DRFS preserves global structure while applying the requested edits more faithfully than competing methods. 
We defer additional qualitative examples to the Appendix.

\subsection{Ablation Studies}

\paragraph{Effect of the additional shift term $c_t$.}
We perform an ablation in which we set  $c_t = 0$, $c_t\simeq(1-t)t$ (ours), or $c_t=t$,
using the same implementation as before with SD3.
Table~\ref{tab:abl_ct_pie_results} shows that a non‑zero $c_t$ improves semantic alignment with the target prompt (higher CLIP similarity on the edited region) at the cost of a reduced source fidelity %

This behaviour is explained by previous experiments%
, where we observed that larger $c_t$ values induce larger gradient updates and produce straighter latent trajectories.  
Geometrically, as illustrated in Fig.\,\ref{fig:dessins}, a higher $c_t$ pushes the path further from the source distribution and more directly toward the target, yielding stronger edits but slightly weaker background preservation. However, as explained before, our gradual shift %
avoids error amplification in early noisy steps.%

\begin{table*}[h]
    \caption{Ablation study on the corrected term $c_t$. The best is shown in bold.}
    \label{tab:abl_ct_pie_results}
    \vspace{-2mm}
    \centering
    \resizebox{\textwidth}{!}{%
    \setlength{\tabcolsep}{4.6pt}
    \begin{tabular}{l|c|c|cccc|cc}
        \toprule
        \multirow{2}{*}{\textbf{Method}} & \multicolumn{1}{c|}{\textbf{Structure}} & \multicolumn{4}{c|}{\textbf{Background Preservation}} & \multicolumn{2}{c}{\textbf{CLIP Similarity}} \\
        \cmidrule(lr){2-2}\cmidrule(lr){3-6}\cmidrule(lr){7-8}
        & Distance $_{\times 10^3} \downarrow$ & PSNR $\uparrow$ & LPIPS $_{\times 10^3} \downarrow$ & MSE $_{\times 10^4} \downarrow$ & SSIM $_{\times 10^2} \uparrow$ & Whole $\uparrow$ & Edited $\uparrow$ \\
        \midrule
        \textbf{DRFS ($c_t\simeq(1-t)t$)} & 23.05 & 23.38 & 93.81 & 67.49 & 84.85 & 26.90 & \best{23.83} \\
        \textbf{DRFS ($c_t=t$)}     & 37.28 & 20.71 & 143.06 & 122.55 & 80.27 & \best{27.09} & 23.21 \\
        \textbf{DRFS ($c_t=0$)}     & \best{8.35} & \best{28.63} & \best{44.66} & \best{21.91} & \best{90.52} & 25.67 & 22.53 \\
        \bottomrule
    \end{tabular}
    }
\end{table*}

\begin{figure}[H]
  \centering
  \begin{subfigure}[t]{0.33\linewidth}
    \centering
    \includegraphics[width=0.32\linewidth]{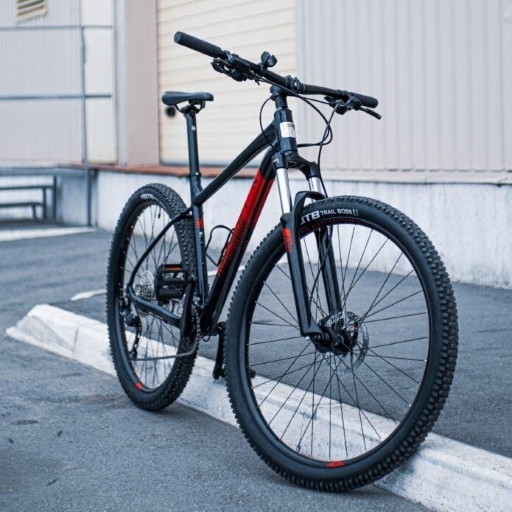}%
    \hspace{1pt}%
    \includegraphics[width=0.32\linewidth]{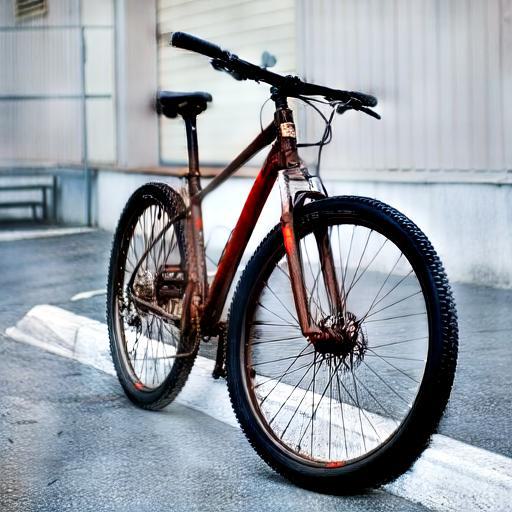}%
    \hspace{1pt}%
    \includegraphics[width=0.32\linewidth]{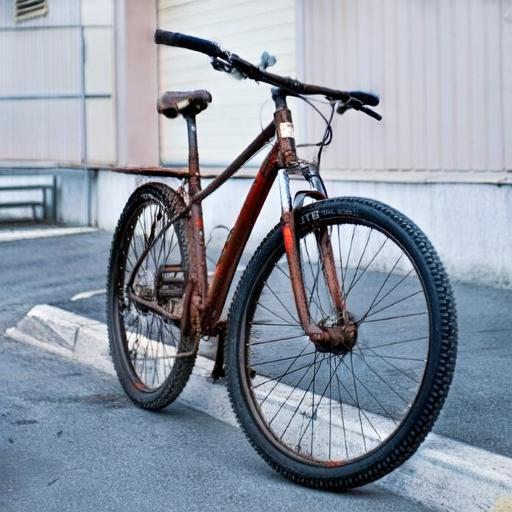}
    \caption*{\textbf{+ rusty}}
  \end{subfigure}\hfill
  \begin{subfigure}[t]{0.33\linewidth}
    \centering
    \includegraphics[width=0.32\linewidth]{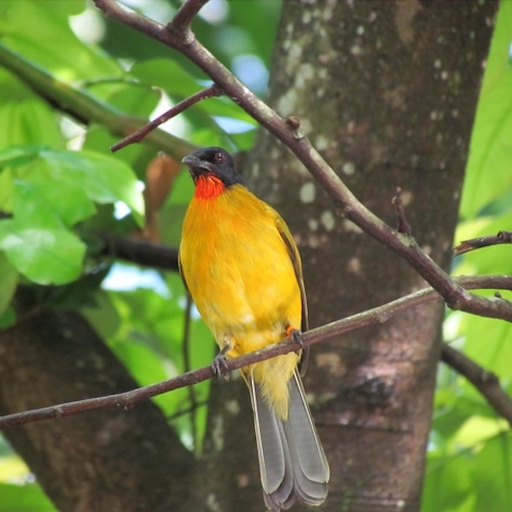}%
    \hspace{1pt}%
    \includegraphics[width=0.32\linewidth]{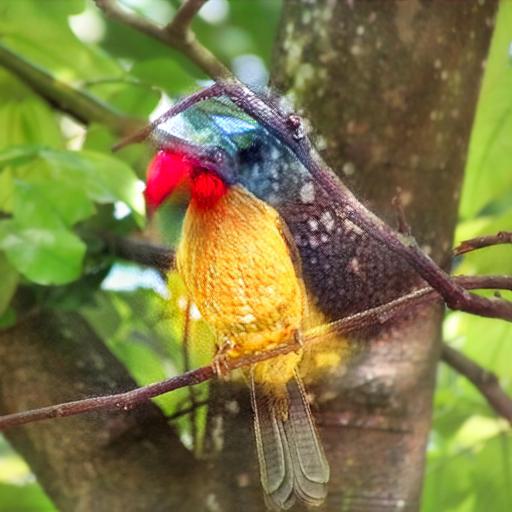}%
    \hspace{1pt}%
    \includegraphics[width=0.32\linewidth]{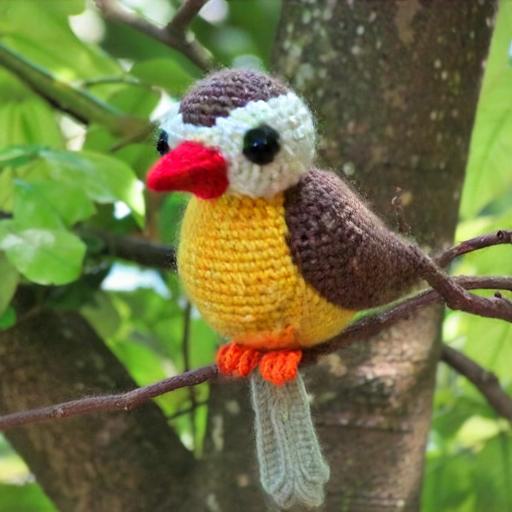}
    \caption*{\textbf{+ crochet}}
  \end{subfigure}\hfill
  \begin{subfigure}[t]{0.33\linewidth}
    \centering
    \includegraphics[width=0.32\linewidth]{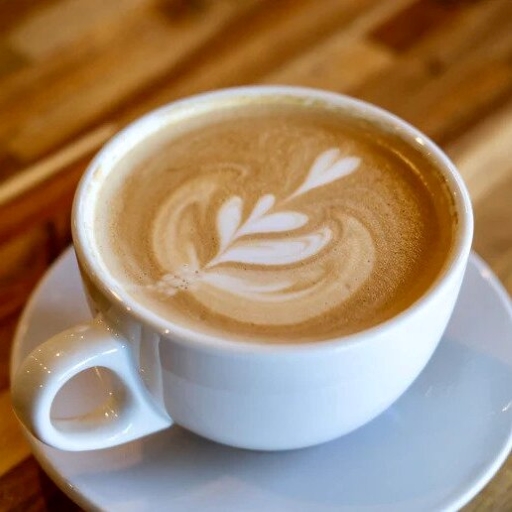}%
    \hspace{1pt}%
    \includegraphics[width=0.32\linewidth]{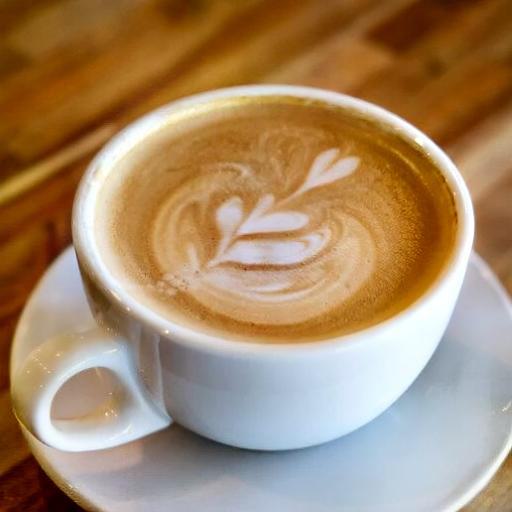}%
    \hspace{1pt}%
    \includegraphics[width=0.32\linewidth]{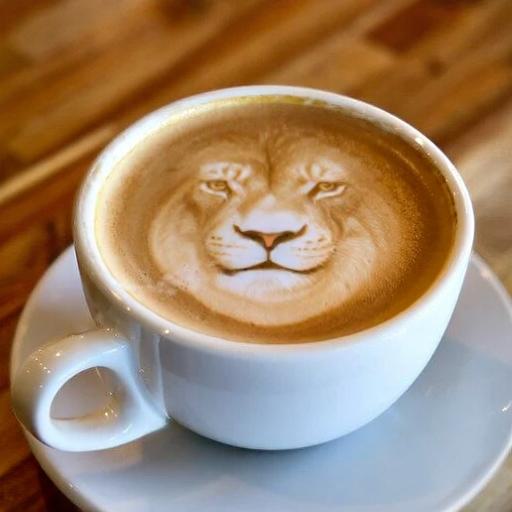}
    \caption*{\textbf{tulip $\rightarrow$ lion}}
  \end{subfigure}

  \caption{Qualitative edits produced by our DRFS with different schedulers. For each triplet: left = source, center = random scheduler, right = descending scheduler.}
  \label{fig:qualitative_triplets}
\end{figure}

\paragraph{Effect of the time-steps scheduler strategy.}Across all our experiments we found that a descending timestep scheduler yields more consistent edits than sampling timesteps uniformly at random (Fig.\,\ref{fig:qualitative_triplets}).  
The intuition is that descending timesteps realise a coarse-to-fine optimisation: early, high-noise steps permit large geometric changes (e.g.\ shape or pose), whereas the final, low-noise steps refine colours and texture.  
In contrast, a random scheduler interleaves coarse and fine updates, often introducing visible artefacts.  
To isolate the scheduler effect, we set $c_t = 0$ and keep all other hyper-parameters fixed.

\section{Related Work}

\paragraph{Text-guided Inversion and Editing.}
Image editing methods can broadly be categorized into training-based and training-free approaches.
Training-based methods fine-tune generative models using triplets—source image, editing instruction, and target image~\cite{brooks2023instructpix2pixlearningfollowimage, zhang2023magicbrush, Geng23instructdiff}—or source image and prompt pairs to reduce reconstruction errors~\cite{kawar2023imagictextbasedrealimage}. 
Training-free methods often rely on inversion for diffusion models~\cite{ju2023directinversionboostingdiffusionbased}. DDIM inversion~\cite{song2021denoising} introduced this idea, later refined by text embedding optimization~\cite{mokady2022nulltextinversioneditingreal}, negative prompting~\cite{miyake2024negativepromptinversionfastimage}, and other strategies \citep{hubermanspiegelglas2024editfriendlyddpmnoise, 30a8b7f8d8ee493ebdb7738c03394510, wallace2022edictexactdiffusioninversion, li2023stylediffusion, lin2024scheduleeditsimpleeffective, brack2024leditslimitlessimageediting}. In addition,~\cite{xu2023inversionfreeimageeditingnatural} introduced an inversion-free editing scheme.

Inverting Rectified Flow (RF) models is more challenging due to higher reconstruction errors~\cite{wang2024qwen2vlenhancingvisionlanguagemodels, rout2025semantic, li2025five}. Existing methods address this via dynamic control~\cite{rout2025semantic}, high-order solvers~\cite{wang2025taming, deng2025fireflow}, noisy latent optimization~\cite{patel2024steeringrectifiedflowmodels}, or fixed-step refinements~\cite{xu2025unveilinversioninvarianceflow}. Similar to diffusion editing~\cite{qi2023fatezerofusingattentionszeroshot, hertz2022prompttopromptimageeditingcross, tumanyan2022plugandplaydiffusionfeaturestextdriven}, RF methods also apply invariance controls such as attention injection~\cite{wang2025taming} or AdaLN feature injection~\cite{xu2025unveilinversioninvarianceflow}. FlowEdit~\cite{kulikov2024floweditinversionfreetextbasedediting} is the only inversion-free RF method, enabling efficient editing without explicit latent recovery. More recently,~\cite{xie2025dnaedit} refine Gaussian noise by aligning predicted and expected velocity fields, while~\cite{yoon2025splitflow} decompose the editing prompt using an LLM and aggregate the resulting flows. Although promising, the latter relies on an additional LLM and requires more NFEs.

\paragraph{Distillation-based Methods.} Score Distillation Sampling~\cite{poole2023dreamfusion}, i.e. leveraging the priors of diffusion models, has been widely studied over the past years \citep{poole2023dreamfusion, wang2023prolificdreamer, katzir2024noisefree, mcallister2024rethinking}. 
Several extensions~\cite{Hertz_2023_ICCV, Nam_2024_CVPR} refine the SDS objective to improve image editing quality. RFDS~\cite{yang2025texttoimage} introduced a distillation framework based on RF models for text-to-3D synthesis. Its variant, iRFDS, was adapted for image editing by first inverting the source image through noise optimization, followed by forward sampling guided by the target prompt.

\section{Conclusion}
We propose Delta Rectified Flow Sampling (DRFS), a new distillation-based method for text-to-image editing that is both inversion-free and training-free. DRFS minimizes the discrepancy between source and target velocities through an energy function, mitigating the over-smoothing artifacts of vanilla RFDS. A time-dependent shift term further guides noisy latents toward the correct semantic trajectory, reducing model-data mismatch during optimization. We establish theoretical connections to DDS and FlowEdit, providing a unified view of score-based, flow-based optimization and ODE editing. Overall, DRFS enables efficient editing without architectural changes and offers a principled perspective on plug-and-play image editing.

\medskip
{\small
\bibliographystyle{ieeenat_fullname}
\bibliography{references}

\begin{thebibliography}{52}
\providecommand{\natexlab}[1]{#1}
\providecommand{\url}[1]{\texttt{#1}}
\expandafter\ifx\csname urlstyle\endcsname\relax
  \providecommand{\doi}[1]{doi: #1}\else
  \providecommand{\doi}{doi: \begingroup \urlstyle{rm}\Url}\fi

\bibitem[Agustsson and Timofte(2017)]{8014884}
Eirikur Agustsson and Radu Timofte.
\newblock Ntire 2017 challenge on single image super-resolution: Dataset and study.
\newblock In \emph{2017 IEEE Conference on Computer Vision and Pattern Recognition Workshops (CVPRW)}, pages 1122--1131, July 2017.
\newblock \doi{10.1109/CVPRW.2017.150}.

\bibitem[Albergo et~al.(2023)Albergo, Boffi, and Vanden-Eijnden]{albergo2023stochasticinterpolantsunifyingframework}
Michael~S. Albergo, Nicholas~M. Boffi, and Eric Vanden-Eijnden.
\newblock Stochastic interpolants: A unifying framework for flows and diffusions, 2023.
\newblock URL \url{https://arxiv.org/abs/2303.08797}.

\bibitem[Brack et~al.(2024)Brack, Friedrich, Kornmeier, Tsaban, Schramowski, Kersting, and Passos]{brack2024leditslimitlessimageediting}
Manuel Brack, Felix Friedrich, Katharina Kornmeier, Linoy Tsaban, Patrick Schramowski, Kristian Kersting, and Apolinário Passos.
\newblock Ledits++: Limitless image editing using text-to-image models, 2024.
\newblock URL \url{https://arxiv.org/abs/2311.16711}.

\bibitem[Brooks et~al.(2023)Brooks, Holynski, and Efros]{brooks2023instructpix2pixlearningfollowimage}
Tim Brooks, Aleksander Holynski, and Alexei~A. Efros.
\newblock Instructpix2pix: Learning to follow image editing instructions, 2023.
\newblock URL \url{https://arxiv.org/abs/2211.09800}.

\bibitem[Deng et~al.(2025)Deng, He, Mei, Wang, and Tang]{deng2025fireflow}
Yingying Deng, Xiangyu He, Changwang Mei, Peisong Wang, and Fan Tang.
\newblock Fireflow: Fast inversion of rectified flow for image semantic editing.
\newblock In \emph{Forty-second International Conference on Machine Learning}, 2025.
\newblock URL \url{https://openreview.net/forum?id=JFafMSAjUm}.

\bibitem[Esser et~al.(2024)Esser, Kulal, Blattmann, Entezari, Müller, Saini, Levi, Lorenz, Sauer, Boesel, Podell, Dockhorn, English, Lacey, Goodwin, Marek, and Rombach]{esser2024scalingrectifiedflowtransformers}
Patrick Esser, Sumith Kulal, Andreas Blattmann, Rahim Entezari, Jonas Müller, Harry Saini, Yam Levi, Dominik Lorenz, Axel Sauer, Frederic Boesel, Dustin Podell, Tim Dockhorn, Zion English, Kyle Lacey, Alex Goodwin, Yannik Marek, and Robin Rombach.
\newblock Scaling rectified flow transformers for high-resolution image synthesis, 2024.
\newblock URL \url{https://arxiv.org/abs/2403.03206}.

\bibitem[Geng et~al.(2023)Geng, Yang, Hang, Li, Gu, Zhang, Bao, Zhang, Hu, Chen, and Guo]{Geng23instructdiff}
Zigang Geng, Binxin Yang, Tiankai Hang, Chen Li, Shuyang Gu, Ting Zhang, Jianmin Bao, Zheng Zhang, Han Hu, Dong Chen, and Baining Guo.
\newblock Instructdiffusion: {A} generalist modeling interface for vision tasks.
\newblock \emph{CoRR}, abs/2309.03895, 2023.
\newblock \doi{10.48550/arXiv.2309.03895}.
\newblock URL \url{https://doi.org/10.48550/arXiv.2309.03895}.

\bibitem[Hertz et~al.(2022)Hertz, Mokady, Tenenbaum, Aberman, Pritch, and Cohen-Or]{hertz2022prompttopromptimageeditingcross}
Amir Hertz, Ron Mokady, Jay Tenenbaum, Kfir Aberman, Yael Pritch, and Daniel Cohen-Or.
\newblock Prompt-to-prompt image editing with cross attention control, 2022.
\newblock URL \url{https://arxiv.org/abs/2208.01626}.

\bibitem[Hertz et~al.(2023)Hertz, Aberman, and Cohen-Or]{Hertz_2023_ICCV}
Amir Hertz, Kfir Aberman, and Daniel Cohen-Or.
\newblock Delta denoising score.
\newblock In \emph{Proceedings of the IEEE/CVF International Conference on Computer Vision (ICCV)}, pages 2328--2337, October 2023.

\bibitem[Ho and Salimans(2021)]{ho2021classifierfree}
Jonathan Ho and Tim Salimans.
\newblock Classifier-free diffusion guidance.
\newblock In \emph{NeurIPS 2021 Workshop on Deep Generative Models and Downstream Applications}, 2021.
\newblock URL \url{https://openreview.net/forum?id=qw8AKxfYbI}.

\bibitem[Huang et~al.(2025)Huang, Liao, Hu, Lin, Wu, Li, Tan, Liu, Liu, Zang, Yu, and Lei]{Huang_2025_CVPR}
Yufei Huang, Bangyan Liao, Yuqi Hu, Haitao Lin, Lirong Wu, Siyuan Li, Cheng Tan, Zicheng Liu, Yunfan Liu, Zelin Zang, Chang Yu, and Zhen Lei.
\newblock Dacapo: Score distillation as stacked bridge for fast and high-quality 3d editing.
\newblock In \emph{Proceedings of the IEEE/CVF Conference on Computer Vision and Pattern Recognition (CVPR)}, pages 16304--16313, June 2025.

\bibitem[Huberman-Spiegelglas et~al.(2024)Huberman-Spiegelglas, Kulikov, and Michaeli]{hubermanspiegelglas2024editfriendlyddpmnoise}
Inbar Huberman-Spiegelglas, Vladimir Kulikov, and Tomer Michaeli.
\newblock An edit friendly ddpm noise space: Inversion and manipulations, 2024.
\newblock URL \url{https://arxiv.org/abs/2304.06140}.

\bibitem[Ju et~al.(2023)Ju, Zeng, Bian, Liu, and Xu]{ju2023directinversionboostingdiffusionbased}
Xuan Ju, Ailing Zeng, Yuxuan Bian, Shaoteng Liu, and Qiang Xu.
\newblock Direct inversion: Boosting diffusion-based editing with 3 lines of code, 2023.
\newblock URL \url{https://arxiv.org/abs/2310.01506}.

\bibitem[Katzir et~al.(2024)Katzir, Patashnik, Cohen-Or, and Lischinski]{katzir2024noisefree}
Oren Katzir, Or~Patashnik, Daniel Cohen-Or, and Dani Lischinski.
\newblock Noise-free score distillation.
\newblock In \emph{The Twelfth International Conference on Learning Representations}, 2024.
\newblock URL \url{https://openreview.net/forum?id=dlIMcmlAdk}.

\bibitem[Kawar et~al.(2023)Kawar, Zada, Lang, Tov, Chang, Dekel, Mosseri, and Irani]{kawar2023imagictextbasedrealimage}
Bahjat Kawar, Shiran Zada, Oran Lang, Omer Tov, Huiwen Chang, Tali Dekel, Inbar Mosseri, and Michal Irani.
\newblock Imagic: Text-based real image editing with diffusion models, 2023.
\newblock URL \url{https://arxiv.org/abs/2210.09276}.

\bibitem[Koo et~al.(2025)Koo, Yoon, Hong, and Yoo]{30a8b7f8d8ee493ebdb7738c03394510}
Gwanhyeong Koo, Sunjae Yoon, \{Ji Woo\} Hong, and \{Chang D.\} Yoo.
\newblock Flexiedit: Frequency-aware latent refinement for enhanced non-rigid editing.
\newblock In Ale{\v s} Leonardis, Elisa Ricci, Stefan Roth, Olga Russakovsky, Torsten Sattler, and G{\"u}l Varol, editors, \emph{Computer Vision – ECCV 2024 - 18th European Conference, Proceedings}, Lecture Notes in Computer Science (including subseries Lecture Notes in Artificial Intelligence and Lecture Notes in Bioinformatics), pages 363--379. Springer Science and Business Media Deutschland GmbH, 2025.
\newblock ISBN 9783031730351.
\newblock \doi{10.1007/978-3-031-73036-8\_21}.
\newblock Publisher Copyright: {\textcopyright} The Author(s), under exclusive license to Springer Nature Switzerland AG 2025.; 18th European Conference on Computer Vision, ECCV 2024 ; Conference date: 29-09-2024 Through 04-10-2024.

\bibitem[Koo et~al.(2024)Koo, Park, and Sung]{Koo:2024PDS}
Juil Koo, Chanho Park, and Minhyuk Sung.
\newblock Posterior distillation sampling.
\newblock In \emph{CVPR}, 2024.

\bibitem[Kulikov et~al.(2024)Kulikov, Kleiner, Huberman-Spiegelglas, and Michaeli]{kulikov2024floweditinversionfreetextbasedediting}
Vladimir Kulikov, Matan Kleiner, Inbar Huberman-Spiegelglas, and Tomer Michaeli.
\newblock Flowedit: Inversion-free text-based editing using pre-trained flow models, 2024.
\newblock URL \url{https://arxiv.org/abs/2412.08629}.

\bibitem[Labs(2024)]{flux2024}
Black~Forest Labs.
\newblock Flux.
\newblock \url{https://github.com/black-forest-labs/flux}, 2024.

\bibitem[Li et~al.(2025)Li, Xie, Wu, Zhang, and Wang]{li2025five}
Minghan Li, Chenxi Xie, Yichen Wu, Lei Zhang, and Mengyu Wang.
\newblock Five: A fine-grained video editing benchmark for evaluating emerging diffusion and rectified flow models.
\newblock \emph{arXiv preprint arXiv:2503.13684}, 2025.

\bibitem[Li et~al.(2023)Li, van~de Weijer, Hu, Khan, Hou, Wang, and Yang]{li2023stylediffusion}
Senmao Li, Joost van~de Weijer, Taihang Hu, Fahad~Shahbaz Khan, Qibin Hou, Yaxing Wang, and Jian Yang.
\newblock Stylediffusion: Prompt-embedding inversion for text-based editing.
\newblock \emph{arXiv preprint arXiv:2303.15649}, 2023.

\bibitem[Lin et~al.(2024)Lin, Wang, Wang, An, Chen, Liu, Tian, Dai, Wang, and Wang]{lin2024scheduleeditsimpleeffective}
Haonan Lin, Mengmeng Wang, Jiahao Wang, Wenbin An, Yan Chen, Yong Liu, Feng Tian, Guang Dai, Jingdong Wang, and Qianying Wang.
\newblock Schedule your edit: A simple yet effective diffusion noise schedule for image editing, 2024.
\newblock URL \url{https://arxiv.org/abs/2410.18756}.

\bibitem[Lipman et~al.(2023)Lipman, Chen, Ben-Hamu, Nickel, and Le]{lipman2023flow}
Yaron Lipman, Ricky T.~Q. Chen, Heli Ben-Hamu, Maximilian Nickel, and Matthew Le.
\newblock Flow matching for generative modeling.
\newblock In \emph{The Eleventh International Conference on Learning Representations}, 2023.
\newblock URL \url{https://openreview.net/forum?id=PqvMRDCJT9t}.

\bibitem[Liu et~al.(2023)Liu, Gong, and Liu]{DBLPLiuG023}
Xingchao Liu, Chengyue Gong, and Qiang Liu.
\newblock Flow straight and fast: Learning to generate and transfer data with rectified flow.
\newblock In \emph{ICLR}, 2023.
\newblock URL \url{https://openreview.net/forum?id=XVjTT1nw5z}.

\bibitem[McAllister et~al.(2024)McAllister, Ge, Huang, Jacobs, Efros, Holynski, and Kanazawa]{mcallister2024rethinking}
David McAllister, Songwei Ge, Jia-Bin Huang, David~W. Jacobs, Alexei~A Efros, Aleksander Holynski, and Angjoo Kanazawa.
\newblock Rethinking score distillation as a bridge between image distributions.
\newblock In \emph{The Thirty-eighth Annual Conference on Neural Information Processing Systems}, 2024.
\newblock URL \url{https://openreview.net/forum?id=I8PkICj9kM}.

\bibitem[Miyake et~al.(2024)Miyake, Iohara, Saito, and Tanaka]{miyake2024negativepromptinversionfastimage}
Daiki Miyake, Akihiro Iohara, Yu~Saito, and Toshiyuki Tanaka.
\newblock Negative-prompt inversion: Fast image inversion for editing with text-guided diffusion models, 2024.
\newblock URL \url{https://arxiv.org/abs/2305.16807}.

\bibitem[Mokady et~al.(2022)Mokady, Hertz, Aberman, Pritch, and Cohen-Or]{mokady2022nulltextinversioneditingreal}
Ron Mokady, Amir Hertz, Kfir Aberman, Yael Pritch, and Daniel Cohen-Or.
\newblock Null-text inversion for editing real images using guided diffusion models, 2022.
\newblock URL \url{https://arxiv.org/abs/2211.09794}.

\bibitem[Nam et~al.(2024)Nam, Kwon, Park, and Ye]{Nam_2024_CVPR}
Hyelin Nam, Gihyun Kwon, Geon~Yeong Park, and Jong~Chul Ye.
\newblock Contrastive denoising score for text-guided latent diffusion image editing.
\newblock In \emph{Proceedings of the IEEE/CVF Conference on Computer Vision and Pattern Recognition (CVPR)}, pages 9192--9201, June 2024.

\bibitem[Patel et~al.(2024)Patel, Wen, Metaxas, and Yang]{patel2024steeringrectifiedflowmodels}
Maitreya Patel, Song Wen, Dimitris~N. Metaxas, and Yezhou Yang.
\newblock Steering rectified flow models in the vector field for controlled image generation, 2024.
\newblock URL \url{https://arxiv.org/abs/2412.00100}.

\bibitem[{Pixabay}()]{pixabay2025}
{Pixabay}.
\newblock Pixabay.
\newblock \url{https://pixabay.com/}.
\newblock License: CC0; accessed 15 May 2025.

\bibitem[Poole et~al.(2023)Poole, Jain, Barron, and Mildenhall]{poole2023dreamfusion}
Ben Poole, Ajay Jain, Jonathan~T. Barron, and Ben Mildenhall.
\newblock Dreamfusion: Text-to-3d using 2d diffusion.
\newblock In \emph{The Eleventh International Conference on Learning Representations}, 2023.
\newblock URL \url{https://openreview.net/forum?id=FjNys5c7VyY}.

\bibitem[Qi et~al.(2023)Qi, Cun, Zhang, Lei, Wang, Shan, and Chen]{qi2023fatezerofusingattentionszeroshot}
Chenyang Qi, Xiaodong Cun, Yong Zhang, Chenyang Lei, Xintao Wang, Ying Shan, and Qifeng Chen.
\newblock Fatezero: Fusing attentions for zero-shot text-based video editing, 2023.
\newblock URL \url{https://arxiv.org/abs/2303.09535}.

\bibitem[Radford et~al.(2021)Radford, Kim, Hallacy, Ramesh, Goh, Agarwal, Sastry, Askell, Mishkin, Clark, Krueger, and Sutskever]{pmlr-v139-radford21a}
Alec Radford, Jong~Wook Kim, Chris Hallacy, Aditya Ramesh, Gabriel Goh, Sandhini Agarwal, Girish Sastry, Amanda Askell, Pamela Mishkin, Jack Clark, Gretchen Krueger, and Ilya Sutskever.
\newblock Learning transferable visual models from natural language supervision.
\newblock In Marina Meila and Tong Zhang, editors, \emph{Proceedings of the 38th International Conference on Machine Learning}, volume 139 of \emph{Proceedings of Machine Learning Research}, pages 8748--8763. PMLR, 18--24 Jul 2021.
\newblock URL \url{https://proceedings.mlr.press/v139/radford21a.html}.

\bibitem[Rombach et~al.(2022)Rombach, Blattmann, Lorenz, Esser, and Ommer]{Rombach_2022_CVPR_sd1_5}
Robin Rombach, Andreas Blattmann, Dominik Lorenz, Patrick Esser, and Bj\"orn Ommer.
\newblock High-resolution image synthesis with latent diffusion models, June 2022.

\bibitem[Rout et~al.(2025)Rout, Chen, Ruiz, Caramanis, Shakkottai, and Chu]{rout2025semantic}
Litu Rout, Yujia Chen, Nataniel Ruiz, Constantine Caramanis, Sanjay Shakkottai, and Wen-Sheng Chu.
\newblock Semantic image inversion and editing using rectified stochastic differential equations.
\newblock In \emph{The Thirteenth International Conference on Learning Representations}, 2025.
\newblock URL \url{https://openreview.net/forum?id=Hu0FSOSEyS}.

\bibitem[Sauer et~al.(2024)Sauer, Boesel, Dockhorn, Blattmann, Esser, and Rombach]{sauer2024sd3.5}
Axel Sauer, Frederic Boesel, Tim Dockhorn, Andreas Blattmann, Patrick Esser, and Robin Rombach.
\newblock Fast high-resolution image synthesis with latent adversarial diffusion distillation.
\newblock In \emph{SIGGRAPH Asia 2024 Conference Papers}, pages 1--11, 2024.

\bibitem[Song et~al.(2021)Song, Meng, and Ermon]{song2021denoising}
Jiaming Song, Chenlin Meng, and Stefano Ermon.
\newblock Denoising diffusion implicit models.
\newblock In \emph{International Conference on Learning Representations}, 2021.
\newblock URL \url{https://openreview.net/forum?id=St1giarCHLP}.

\bibitem[Su et~al.(2023)Su, Song, Meng, and Ermon]{su2023dual}
Xuan Su, Jiaming Song, Chenlin Meng, and Stefano Ermon.
\newblock Dual diffusion implicit bridges for image-to-image translation.
\newblock In \emph{The Eleventh International Conference on Learning Representations}, 2023.
\newblock URL \url{https://openreview.net/forum?id=5HLoTvVGDe}.

\bibitem[Tumanyan et~al.(2022)Tumanyan, Geyer, Bagon, and Dekel]{tumanyan2022plugandplaydiffusionfeaturestextdriven}
Narek Tumanyan, Michal Geyer, Shai Bagon, and Tali Dekel.
\newblock Plug-and-play diffusion features for text-driven image-to-image translation, 2022.
\newblock URL \url{https://arxiv.org/abs/2211.12572}.

\bibitem[Wallace et~al.(2022)Wallace, Gokul, and Naik]{wallace2022edictexactdiffusioninversion}
Bram Wallace, Akash Gokul, and Nikhil Naik.
\newblock Edict: Exact diffusion inversion via coupled transformations, 2022.
\newblock URL \url{https://arxiv.org/abs/2211.12446}.

\bibitem[Wang et~al.(2025)Wang, Pu, Qi, Guo, Ma, Huang, Chen, Li, and Shan]{wang2025taming}
Jiangshan Wang, Junfu Pu, Zhongang Qi, Jiayi Guo, Yue Ma, Nisha Huang, Yuxin Chen, Xiu Li, and Ying Shan.
\newblock Taming rectified flow for inversion and editing.
\newblock In \emph{Forty-second International Conference on Machine Learning}, 2025.
\newblock URL \url{https://openreview.net/forum?id=uDreZphNky}.

\bibitem[Wang et~al.(2024)Wang, Bai, Tan, Wang, Fan, Bai, Chen, Liu, Wang, Ge, Fan, Dang, Du, Ren, Men, Liu, Zhou, Zhou, and Lin]{wang2024qwen2vlenhancingvisionlanguagemodels}
Peng Wang, Shuai Bai, Sinan Tan, Shijie Wang, Zhihao Fan, Jinze Bai, Keqin Chen, Xuejing Liu, Jialin Wang, Wenbin Ge, Yang Fan, Kai Dang, Mengfei Du, Xuancheng Ren, Rui Men, Dayiheng Liu, Chang Zhou, Jingren Zhou, and Junyang Lin.
\newblock Qwen2-vl: Enhancing vision-language model's perception of the world at any resolution, 2024.
\newblock URL \url{https://arxiv.org/abs/2409.12191}.

\bibitem[Wang et~al.(2004)Wang, Bovik, Sheikh, and Simoncelli]{ssim}
Z.~Wang, A.C. Bovik, H.R. Sheikh, and E.P. Simoncelli.
\newblock Image quality assessment: From error visibility to structural similarity.
\newblock \emph{IEEE Transactions on Image Processing}, page 600–612, Apr 2004.
\newblock \doi{10.1109/tip.2003.819861}.
\newblock URL \url{http://dx.doi.org/10.1109/tip.2003.819861}.

\bibitem[Wang et~al.(2023)Wang, Lu, Wang, Bao, Li, Su, and Zhu]{wang2023prolificdreamer}
Zhengyi Wang, Cheng Lu, Yikai Wang, Fan Bao, Chongxuan Li, Hang Su, and Jun Zhu.
\newblock Prolificdreamer: High-fidelity and diverse text-to-3d generation with variational score distillation.
\newblock In \emph{Thirty-seventh Conference on Neural Information Processing Systems}, 2023.
\newblock URL \url{https://openreview.net/forum?id=ppJuFSOAnM}.

\bibitem[Xie et~al.(2025)Xie, Li, Li, Wu, Yi, and Zhang]{xie2025dnaedit}
Chenxi Xie, Minghan Li, Shuai Li, Yuhui Wu, Qiaosi Yi, and Lei Zhang.
\newblock {DNAE}dit: Direct noise alignment for text-guided rectified flow editing.
\newblock In \emph{The Thirty-ninth Annual Conference on Neural Information Processing Systems}, 2025.
\newblock URL \url{https://openreview.net/forum?id=JxBA9OJExP}.

\bibitem[Xu et~al.(2025)Xu, Jiang, Hu, Luo, He, Zhang, Wang, Wu, Ling, and Wang]{xu2025unveilinversioninvarianceflow}
Pengcheng Xu, Boyuan Jiang, Xiaobin Hu, Donghao Luo, Qingdong He, Jiangning Zhang, Chengjie Wang, Yunsheng Wu, Charles Ling, and Boyu Wang.
\newblock Unveil inversion and invariance in flow transformer for versatile image editing, 2025.
\newblock URL \url{https://arxiv.org/abs/2411.15843}.

\bibitem[Xu et~al.(2023)Xu, Huang, Pan, Ma, and Chai]{xu2023inversionfreeimageeditingnatural}
Sihan Xu, Yidong Huang, Jiayi Pan, Ziqiao Ma, and Joyce Chai.
\newblock Inversion-free image editing with natural language, 2023.
\newblock URL \url{https://arxiv.org/abs/2312.04965}.

\bibitem[Yang et~al.(2025)Yang, Cheng, Yang, Liu, and Lin]{yang2025texttoimage}
Xiaofeng Yang, Chen Cheng, Xulei Yang, Fayao Liu, and Guosheng Lin.
\newblock Text-to-image rectified flow as plug-and-play priors.
\newblock In \emph{The Thirteenth International Conference on Learning Representations}, 2025.
\newblock URL \url{https://openreview.net/forum?id=SzPZK856iI}.

\bibitem[Yoon et~al.(2025)Yoon, Li, Beaudouin, Wen, Azhar, and Wang]{yoon2025splitflow}
Sung-Hoon Yoon, Minghan Li, Gaspard Beaudouin, Congcong Wen, Muhammad~Rafay Azhar, and Mengyu Wang.
\newblock Splitflow: Flow decomposition for inversion-free text-to-image editing.
\newblock In \emph{The Thirty-ninth Annual Conference on Neural Information Processing Systems}, 2025.
\newblock URL \url{https://openreview.net/forum?id=lRs6qSMKH1}.

\bibitem[Zhang et~al.(2023)Zhang, Mo, Chen, Sun, and Su]{zhang2023magicbrush}
Kai Zhang, Lingbo Mo, Wenhu Chen, Huan Sun, and Yu~Su.
\newblock Magicbrush: A manually annotated dataset for instruction-guided image editing.
\newblock In \emph{Thirty-seventh Conference on Neural Information Processing Systems Datasets and Benchmarks Track}, 2023.
\newblock URL \url{https://openreview.net/forum?id=ZsDB2GzsqG}.

\bibitem[Zhang et~al.(2018)Zhang, Isola, Efros, Shechtman, and Wang]{lpips}
Richard Zhang, Phillip Isola, Alexei~A. Efros, Eli Shechtman, and Oliver Wang.
\newblock The unreasonable effectiveness of deep features as a perceptual metric.
\newblock In \emph{2018 IEEE/CVF Conference on Computer Vision and Pattern Recognition}, Jun 2018.
\newblock \doi{10.1109/cvpr.2018.00068}.
\newblock URL \url{http://dx.doi.org/10.1109/cvpr.2018.00068}.

\bibitem[Zheng et~al.(2023)Zheng, Le, Shaul, Lipman, Grover, and Chen]{zheng2023guidedflowsgenerativemodeling}
Qinqing Zheng, Matt Le, Neta Shaul, Yaron Lipman, Aditya Grover, and Ricky T.~Q. Chen.
\newblock Guided flows for generative modeling and decision making, 2023.
\newblock URL \url{https://arxiv.org/abs/2311.13443}.

\end{thebibliography}
}

\clearpage
\appendix

\renewcommand{\thefigure}{S\arabic{figure}}
\renewcommand{\thetable}{S\arabic{table}}
\renewcommand{\thealgorithm}{S\arabic{algorithm}}
\renewcommand{\theequation}{S\arabic{equation}}
\setcounter{figure}{0}
\setcounter{table}{0}
\setcounter{algorithm}{0}
\setcounter{equation}{0}

\twocolumn[
\begin{center}
    {\LARGE  \textbf{Delta Rectified Flow Sampling for Text-to-Image Editing }\\
    Appendix}
\end{center}
]

\section{Diffusion Models}
\label{appendix:secA}
\subsection{Diffusion models background}
Diffusion models define a forward process that gradually adds Gaussian noise to a clean image (or its latent) \(x_0\) and a reverse (denoising) process that recovers \(x_0\) from a noisy sample. 

Given noise schedulers \(a_t\) and \(b_t\), the forward process is defined as:
\begin{align} \label{eqn::diffusion_forward}
   x_t = a_t x_0 + b_t \varepsilon, \quad x_0 \sim p_{\text{0}},\, \varepsilon \sim \mathcal{N}(0,I). 
\end{align}
In practice, schedulers are chosen as
\[
a_t = \sqrt{\bar{\alpha}_t},\quad b_t = \sqrt{1 - \bar{\alpha}_t},
\]
where $\bar \alpha_0=1,\bar \alpha_T=0$. A neural network \(\varepsilon_\theta\) is then trained to predict the noise \(\varepsilon\) from a noisy input using the loss function:
\small
\begin{multline}
\mathcal{L}_{\text{DDPM}}(\theta) = \mathbb{E}_{t \sim \mathcal{U}\{1,\dots,T\},\, x_0 \sim p_0,\, \varepsilon \sim \mathcal{N}(0,I)} \Bigl[ \\
\left\| \varepsilon_\theta(a_t x_0 + b_t \varepsilon,\, t) - \varepsilon \right\|^2
\Bigr].
\end{multline}

Once we have a pretrained neural network $\epsilon_\theta$, clean images can be sampled in various ways. For simplicity, below, we explain a deterministic sampling process (DDIM \cite{song2021denoising}). From an initial gaussian noise \(x_T \sim \mathcal{N}(0,I)\), $\{ x_{t} \}_{t=0}^{T}$ are recursively defined as:
\begin{align}
    x_{t-1} = a_{t-1}\,\left(\frac{x_t - b_t\,\varepsilon_\theta(x_t, t)}{a_t}\right)
+ b_{t-1}\,\varepsilon_\theta(x_t, t).
\end{align}
\paragraph{Score function.} Note that the forward process Eq. \eqref{eqn::diffusion_forward} induces a marginal distribution of $x_t$, which we denote as $p_t$. The score function of $p_t$ is defined as the gradient of the log-density: $s(x_t,t)\colon = \nabla_{x_t}\log p_t(x_t)$.
The ground truth score function and the truth noise prediction exhibit the following connection:
\begin{align*}
    s(x_t,t) = \nabla_{x_t} \log p_t(x_t) = -\frac{\mathbb{E}[\epsilon \mid x_t]}{b_t}.
\end{align*}
Therefore, a diffusion model that predicts noise can be interpreted as a model that predicts the score function with the relation of $s_\theta(x_t,t) = -\frac{\varepsilon_\theta(x_t,t) }{b_t}$.

\subsection{Reconstruction errors}
\label{appendix:secA:recon}

We compare the reconstruction accuracy of (i) a diffusion model inverted with DDIM and (ii) a rectified-flow model inverted with a first-order (Euler) solver.  For each method we run an inversion then reconstruction and measure the full reconstruction error at every time-step \(t\).

\begin{algorithm}[H]
\caption{DDIM inversion — reconstruction error}
\begin{algorithmic}[1]
\Require Source image \(x_0\); total steps \(T\)
\For{\(t = 0\) \textbf{to} \(T-1\)}  \Comment{DDIM inversion}
    \State \(x_{t+1} \gets a_{t+1}\!\left(\dfrac{x_t - b_t \varepsilon_\theta(x_t,t)}{a_t}\right)
              + b_{t+1}\,\varepsilon_\theta(x_t,t)\)
\EndFor
\State \(\tilde{x}_T \gets x_T\)
\For{\(t = T-1\) \textbf{down to} \(0\)}  \Comment{DDIM reconstruction}
    \State \(\tilde{x}_{t} \gets a_t\!\left(\dfrac{\tilde{x}_{t+1} - b_{t+1}\varepsilon_\theta(\tilde{x}_{t+1},t+1)}{a_{t+1}}\right)
              + b_t\,\varepsilon_\theta(\tilde{x}_{t+1},t+1)\)
\EndFor
\State \(e_t \gets \lVert \tilde{x}_{t}-x_{t}\rVert^{}_2\) for all \(t\in\{0,\dots,T-1\}\)
\State \Return \(e=(e_0,\dots,e_{T-1})\)
\end{algorithmic}
\end{algorithm}

The per-step approximation used is \(\varepsilon_\theta(x_{t},t)\approx\varepsilon_\theta(x_{t+1},t)\).

\begin{algorithm}[H]
\caption{Rectified-flow inversion — reconstruction error}
\begin{algorithmic}[1]
\Require Source image \(x_0\); total steps \(T\)
\For{\(i = 0\) \textbf{to} \(T-1\)}  \Comment{Euler inversion}
    \State \(x_{t_{i+1}} \gets x_{t_i} - (t_i - t_{i+1})\,v_\theta(x_{t_i},t_i)\)
\EndFor
\State \(\tilde{x}_{t_T} \gets x_{t_T}\)
\For{\(i = T-1\) \textbf{down to} \(0\)}  \Comment{Euler reconstruction}
    \State \(\tilde{x}_{t_i} \gets \tilde{x}_{t_{i+1}} + (t_i - t_{i+1})\,v_\theta(\tilde{x}_{t_{i+1}},t_{i+1})\)
\EndFor
\State \(e_{t_i} \gets \lVert \tilde{x}_{t_i} - x_{t_i}\rVert^{}_2\) for all \(i\in\{0,\dots,T-1\}\)
\State \Return \(e=(e_{t_0},\dots,e_{t_{T-1}})\)
\end{algorithmic}
\end{algorithm}

Here the per-step approximation is \(v_\theta(x_{t_i},t_i)\approx v_\theta(x_{t_{i+1}},t_i)\).

\begin{figure}[h]
  \centering
  \resizebox{0.5\linewidth}{!}{\input{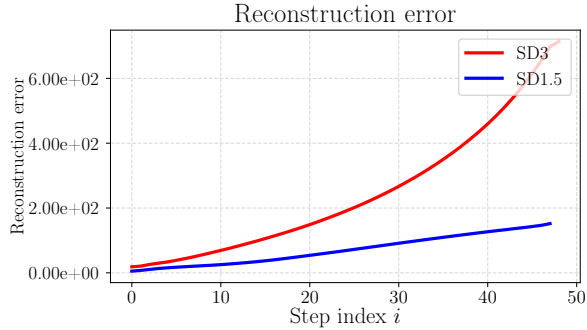}}
  \caption{Reconstruction-error comparison between diffusion and rectified-flow models.}
  \label{fig:rec-errors}
\end{figure}

\vspace{.5em}
\noindent
Figure~\ref{fig:rec-errors} plots the resulting error curves for a diffusion model (SD1.5) and a rectified-flow model (SD3). 

The gap can stem from two key architectural differences: diffusion models employ the scheduler
\((a_t,b_t)=(\sqrt{\bar\alpha_t},\sqrt{1-\bar\alpha_t})\), whereas rectified-flow models use the linear scheduler \((1 - t,\,t)\); and diffusion models directly predict the noise \(\varepsilon\), while rectified-flow models predict the velocity \(u = \varepsilon - x_0\).

\textbf{The rectified-flow reconstruction error is significantly higher across all steps, motivating our investigation of inversion-free editing methods.}

\subsection{Distillation Sampling paradigm}
\begin{figure}[h]
  \centering
  \includegraphics[width=\linewidth]{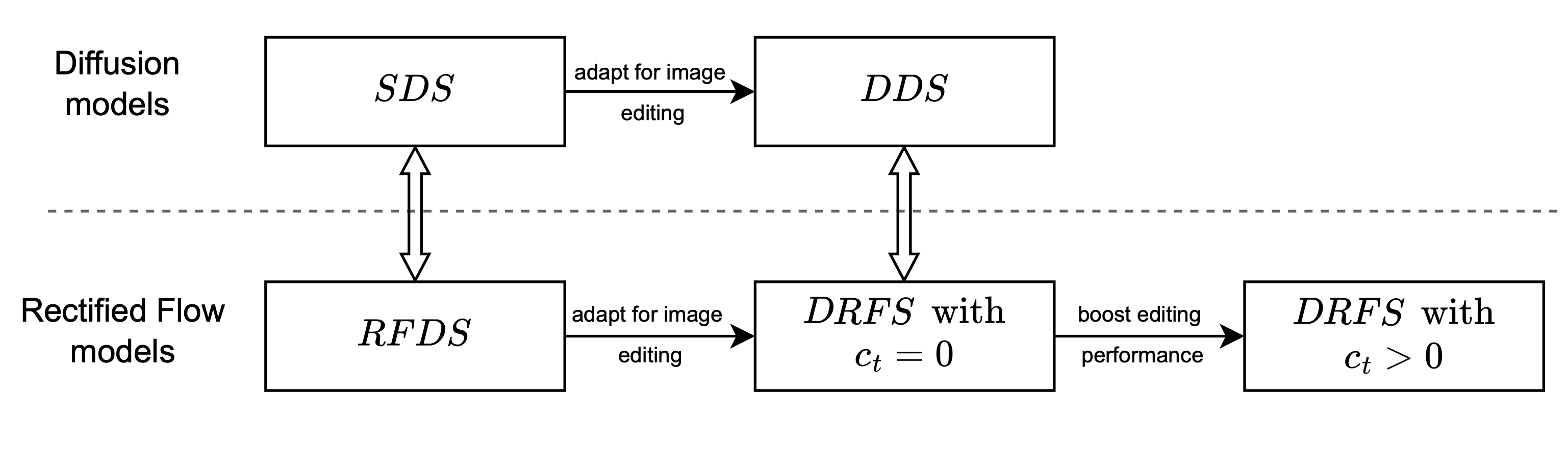}
  \caption{DRFS in the Distillation Sampling paradigm}
  \label{fig:DS_figure}
\end{figure}

Figure \eqref{fig:DS_figure} represents the position of our DRFS in the Distillation Sampling paradigm.

\section{Additional implementation details}
\label{appendix:secB}

An implementation of our method is available in the code appendix.

\subsection{Stable Diffusion 3}
We used a batch size of 1 and a unit weighting function, following \citep{Hertz_2023_ICCV}. We set the source and target CFG values to 6 and 16.5, respectively, and performed 50 optimization steps using the Stable Diffusion descending time-steps scheduler.

We used this setting for all figures plotted in the paper unless stated otherwise.

During the very noisy early time-steps we apply a small learning rate to avoid drifting too far from the source image while still permitting non-rigid edits.  
The rate is increased in the second half of optimization, where latents are cleaner and substantive edits are easier to apply, and then gently decayed at the final few steps, when further changes are unlikely to be beneficial.

\begin{figure}[h]
  \centering
  \resizebox{0.5\linewidth}{!}{\input{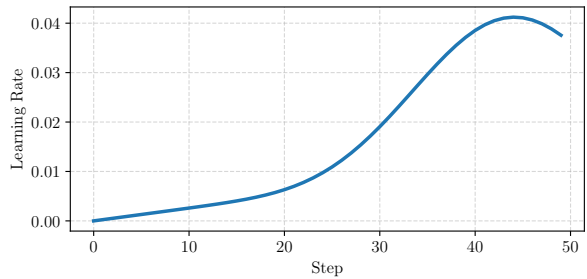}} %
  \caption{Learning rate used in our optimization.}
  \label{fig:lr-schedule}
\end{figure}

\subsection{Stable Diffusion 3.5}
We used the same hyperparameters  as SD3, except for CFG, where we set the source and target CFG values to 5.5 and 13.5 , respectively.

\subsection{Efficiency and computational cost}

All experiments were conducted on a single NVIDIA RTX A6000 GPU (48 GB VRAM), with 16 CPU cores and 64 GB of RAM. On average,  on SD3, one edit takes 7.3 seconds using our DRFS method, compared to 4.9 seconds with FlowEdit and 2 minutes and 26 seconds with the distillation-based method iRFDS (averaged over 700 edits from the PIE benchmark).
\begin{table}[h]
\centering
\small
\caption{\textbf{Efficiency on PIE (700 edits) on SD3.} Mean $\pm$ std over images.}
\label{tab:runtime}
\resizebox{\columnwidth}{!}{
\setlength{\tabcolsep}{4.0pt}
\begin{tabular}{lccc}
\toprule
Method & NFE $\downarrow$ & Peak VRAM (GB) $\downarrow$ & Time (s/edit) $\downarrow$ \\
\midrule
FlowEdit (official ODE) & 33 & 18.52 $\pm$ 0.01 & 4.9 $\pm$ 0.4 \\
iRFDS & 2800 & 37.86 $\pm$ 0.01 & 145.3 $\pm$ 2.7 \\
\rowcolor{gray!10}
DRFS (ours) & 50 & 18.53 $\pm$ 0.03 & 7.3 $\pm$ 1.3 \\
\bottomrule
\end{tabular}}
\end{table}

\subsection{On the shift schedule}
\label{app:shift}
Under a descending timestep schedule, early iterations operate at high noise ($t \simeq 1$), where gradients are noisier and the current estimate $x_0^k$ is least accurate.
A large early shift can \emph{reinject} these errors into future velocity queries and cause drift.

Let $\hat{x}_t(x_0)$ denote the shifted forward state used to query target velocities, let $x_0^\star$ be an ideal edited latent, and define the current optimization error $e_k := x_0^k - x_0^\star$.
With rectified-flow forward $(1-t)x_0 + t\varepsilon$, introducing the shift yields
\[
\hat{x}_t(x_0) = (1-t)x_0 + t\varepsilon + c_t(x_0 - x_0^{src}).
\]
Define $A_t := (1-t) + c_t$. Then the shift amplifies the optimization error as
\[
\hat{x}_t(x_0^k) - \hat{x}_t(x_0^\star) = A_t\, e_k,
\]
so \textbf{$A_t$ directly controls error reinjection}.

Moreover, if we assume that $v_\theta(\cdot,t,\varphi)$ is $L_t$-Lipschitz in $x$, the velocity mismatch is bounded by
\[
\|v_\theta(\hat{x}_t(x_0^k),t,\varphi^\text{tgt}) - v_\theta(x_t^\star,t,\varphi^\text{tgt})\|
\le L_t\bigl(A_t\|e_k\| + c_t\|\Delta\|\bigr),
\]
where $\Delta := x_0^\star - x_0^{src}$ and $x_t^\star := (1-t)x_0^\star + t\varepsilon$.

Since $\|e_k\|$ is largest early, we require $A_t$ to be small near $t\simeq 1$ while keeping a non-zero shift at mid-noise to improve target alignment.

This motivates $c_t \rightarrow 0$ as $t\rightarrow 0$ and $t\rightarrow 1$, and our simple choice $c_t=t(1-t)$.
In contrast, FlowEdit uses $c_t=t$, which gives $A_t\equiv 1$ (no attenuation), while $c_t=t(1-t)$ yields $A_t=1-t^2 \rightarrow 0$ as $t\rightarrow 1$.

\subsection{Additional details}

For Fig. \ref{fig:oversmooth}, we took the official implementation of RFDS \cite{yang2025texttoimage}, adapted to start the optimization from the source image, and slightly reduced the target CFG. %

For Fig. \ref{fig:subfig:S_R} and \ref{fig:subfig:grad_norms_eta}, we used 20 images, source and target prompts from PIE benchmark.

For Fig. \ref{fig:qualitative_triplets}, we used $c_t=0$, 40 optimisation steps and simply a constant learning rate of $0.02$, and same CFG values as before, to highlight and isolate the impact of the scheduler strategy. To try a good optimization set-up, we tested some target CFG values between between 12.5 and 18.5.

The results from the table \ref{tab:pie_results} of PIE benchmark of diffusion based methods were taken from \cite{ju2023directinversionboostingdiffusionbased}. The results from FireFlow, RFSolver and RF-Inv, were taken from the FireFlow paper, and completed on the the LPIPS and MSE metrics by running the evaluation ourselves with each official implementation. For FlowEdit and iRFDS, we also used the official implementations, to run the benchmark.

When computing the metrics on our additional dataset (Fig. ~\ref{fig:lpips-clip}), we also took the official implementation of each model.

\subsection{Additional dataset generation details}
\label{sec:add_dataset}
We used the Qwen2.5-VL-7B-Instruct \cite{wang2024qwen2vlenhancingvisionlanguagemodels} model to generate more than 300 source captions and target prompts. 

\begin{itemize}[leftmargin=*,topsep=0pt,itemsep=1ex]
  \item \textbf{Prompt used to caption source images:}
    \begin{adjustwidth}{1em}{1em}
      \begin{Verbatim}
Describe simply this image in just very few words
      \end{Verbatim}
    \end{adjustwidth}

  \item \textbf{Prompt used to generate target prompts:}
    \begin{adjustwidth}{1em}{1em}
      \begin{Verbatim}
Given the original description {source_prompt}
Generate a new description by modifying the most important object, attribute:
the description should have a completely different meaning, by just and only 
changing a (or 2 MAXIMUM) word(s). You can also add or remove a new object, attribute, etc. in the description. 
The change should totally alter the meaning or visual content of the description. 
All other words MUST remain the same and in the same order. Return only the new description.
      \end{Verbatim}
    \end{adjustwidth}
\end{itemize}

\section{Additional results}
\label{appendix:secC}
\subsection{Effective gradient cancellation in irrelevant parts}

 \begin{figure*}[h]
  \centering
  \newcommand{\blockwid}{0.56\textwidth}   %
  \newcommand{\onewid}{0.14\textwidth}     %

  \begin{subfigure}[t]{\blockwid}
    \centering
    \newcommand{\trajwid}{0.24\linewidth}
    \includegraphics[width=\trajwid]{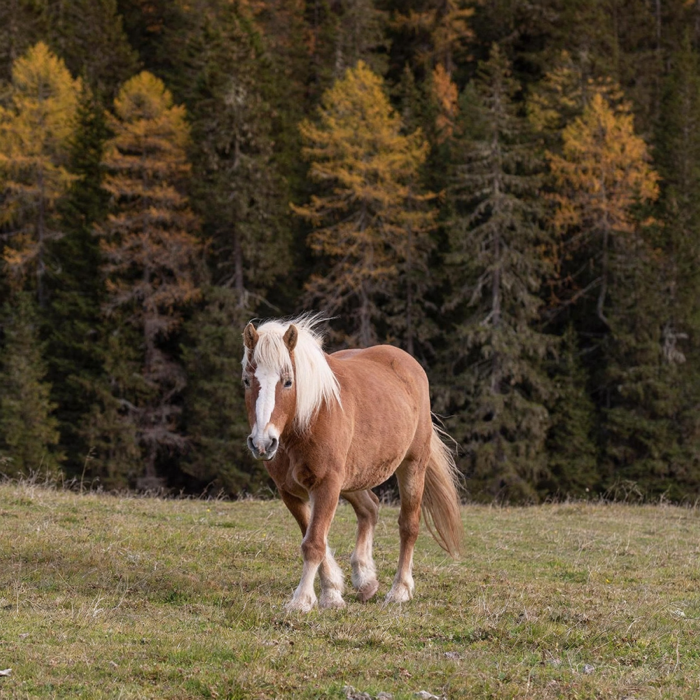}\hfill
    \includegraphics[width=\trajwid]{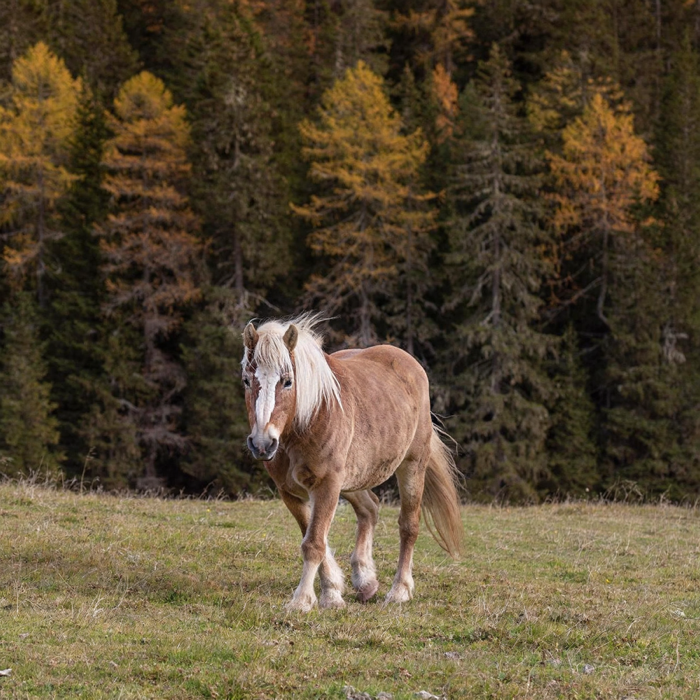}\hfill
    \includegraphics[width=\trajwid]{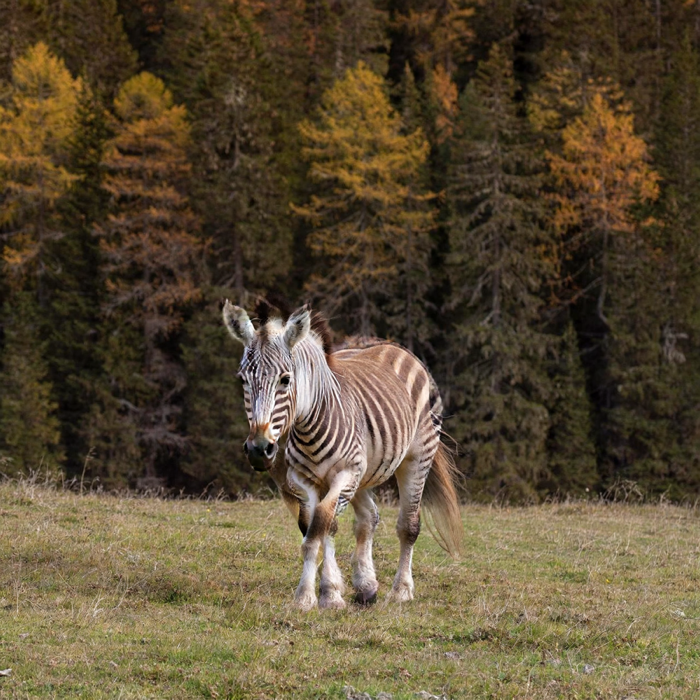}\hfill
    \includegraphics[width=\trajwid]{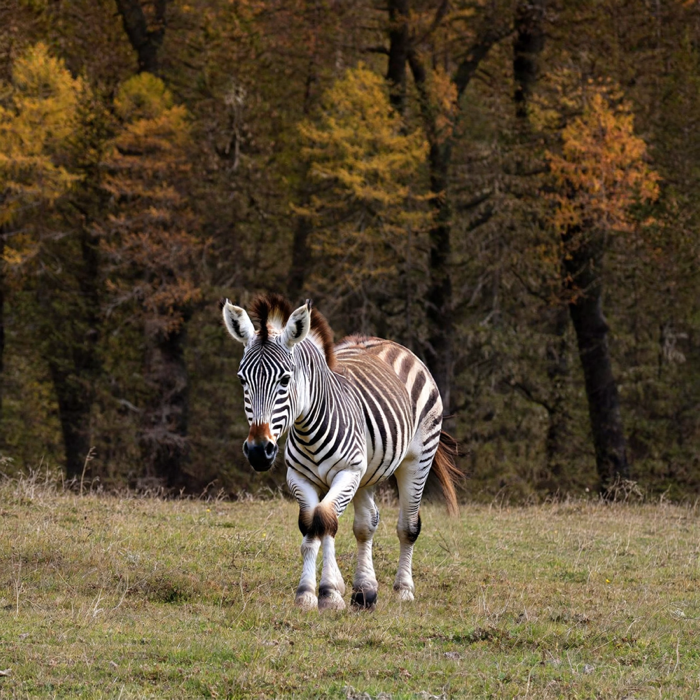}
    \\[0.3em]
    \textit{"Brown horse"} $\rightarrow$ \textit{"Zebra"}
  \end{subfigure}\hfill
  \begin{subfigure}[t]{\onewid}
    \centering
    \includegraphics[width=0.95\linewidth]{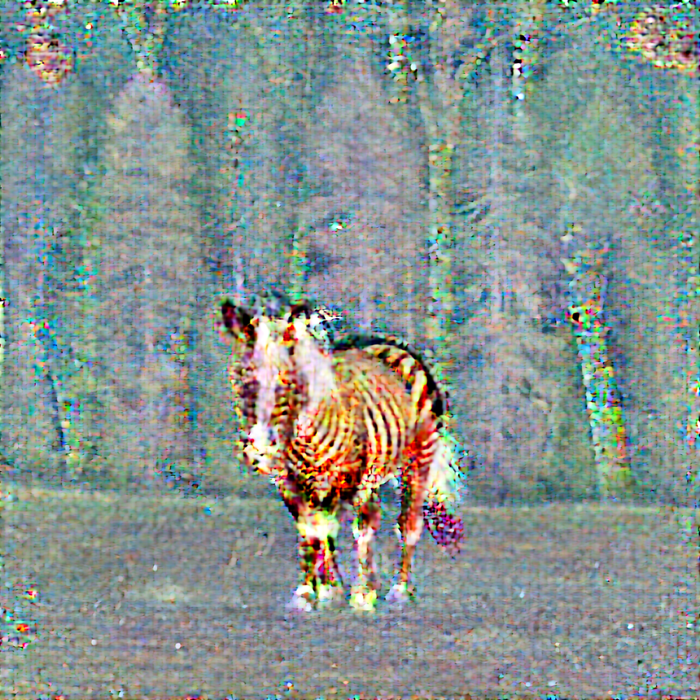}
    \caption*{\gradcap{$\nabla \mathcal{E}_{\text{DRFS}}$}}
  \end{subfigure}\hfill
  \begin{subfigure}[t]{\onewid}
    \centering
    \includegraphics[width=0.95\linewidth]{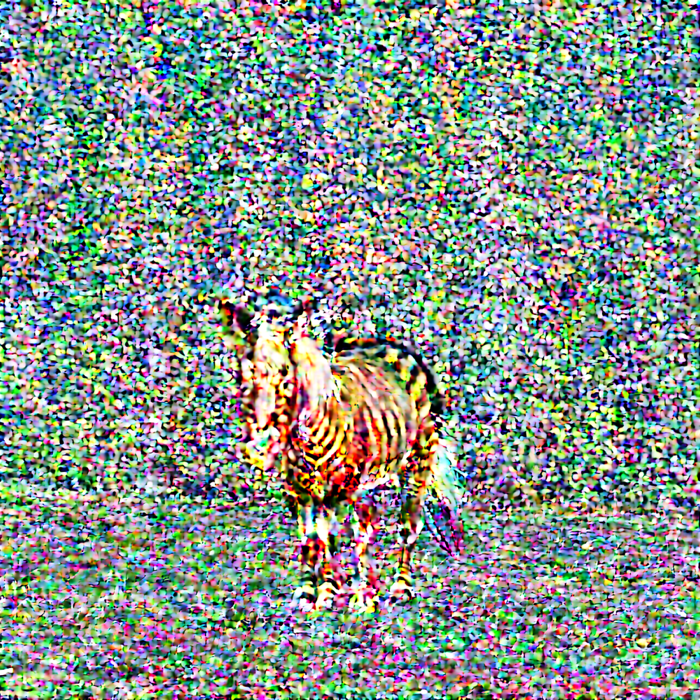}
    \caption*{\gradcap{$\nabla \mathcal{E}_{\text{RFDS}}(\hat x_t^\text{tgt},\varphi^\text{tgt})$}}
  \end{subfigure}\hfill
  \begin{subfigure}[t]{\onewid}
    \centering
    \includegraphics[width=0.95\linewidth]{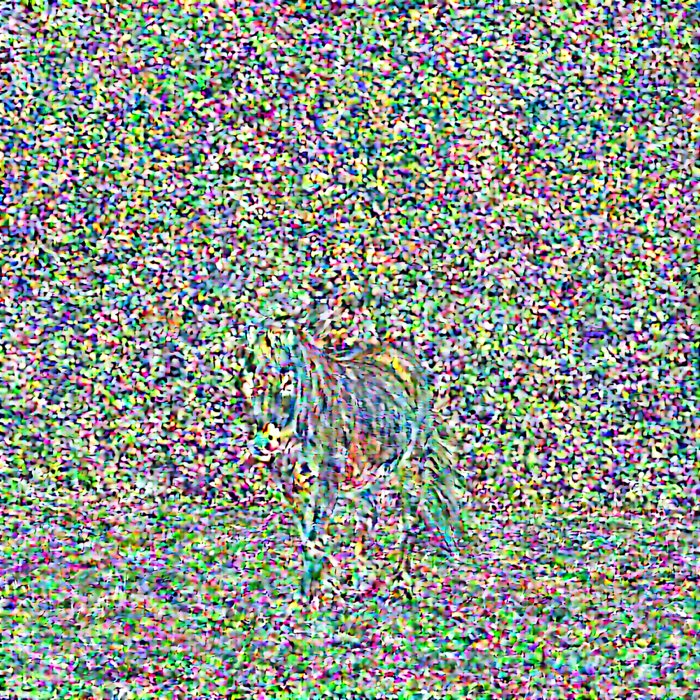}
    \caption*{\gradcap{ $\nabla \mathcal{E}_{\text{RFDS}}( x_t^{src},\varphi^{src})$}}
  \end{subfigure}

  \vspace{0.6em}
  \caption{DRFS gradients. DRFS gradients cancel out in irrelevant parts of the image.}
  \label{fig:dds-gradients}
\end{figure*}

Rewritting Eq. \ref{eq:DRFS_grad}, we have
\small
\begin{equation}
\begin{split}
\nabla_\Theta \mathcal{E}_{\text{DRFS}}
&= \mathbb{E}_{t,\varepsilon}\bigl[
      w_{\text{DRFS}}(t)\bigl(v_\theta(\hat{x}_t^{\text{tgt}})
      - v_\theta(x_t^{\text{src}})\bigr)
    \bigr] \\
 &=- \mathbb{E}_{t,\varepsilon}\bigl[
      w_{\text{DRFS}}(t)\bigl(\dot{\hat{x}}_t^{\text{tgt}}
      - \dot{x}_t^{\text{src}}\bigr)
    \bigr] \\
&= \mathbb{E}_{t,\varepsilon}\bigl[
      w_{\text{DRFS}}(t)\bigl(v_\theta(\hat{x}_t^{\text{tgt}})
      - \dot{\hat{x}}_t^{\text{tgt}}\bigr)
    \bigr] \\
&\quad - \mathbb{E}_{t,\varepsilon}\bigl[
      w_{\text{DRFS}}(t)\bigl(v_\theta(x_t^{\text{src}})
      - \dot{x}_t^{\text{src}}\bigr)
    \bigr].
\end{split}
\label{eq:DRFS_RFDS}
\end{equation}

We use the notation $\nabla \mathcal{E}_{\text{RFDS}}( x_t^{{src}},\varphi^{\text{src}})$ and $\nabla \mathcal{E}_{\text{RFDS}}( x_t^{\text{src}},\varphi^{\text{src}})$ for respectively the first and the second term in Equation \ref{eq:DRFS_RFDS}, when using only one $(t, \varepsilon)$ pair to compute the gradients.

Figure \ref{fig:dds-gradients} visualize the DRFS gradients, and its differential nature, cancelling irrelevant gradients in irrelevant areas. This propery clearly echoes DDS \cite{Hertz_2023_ICCV}. However, the introduction of the $c_t$ term in $\hat x_t^\text{tgt}$, gives straighter paths and larger gradient updates, so a much more effective editing. DDS official implementation uses 200 optimization, while we use 50 optimization steps with DRFS.

\subsection{Result on our additional dataset for different CFG values}

\begin{figure}[h]
  \centering
  \includegraphics[width=0.95\linewidth]{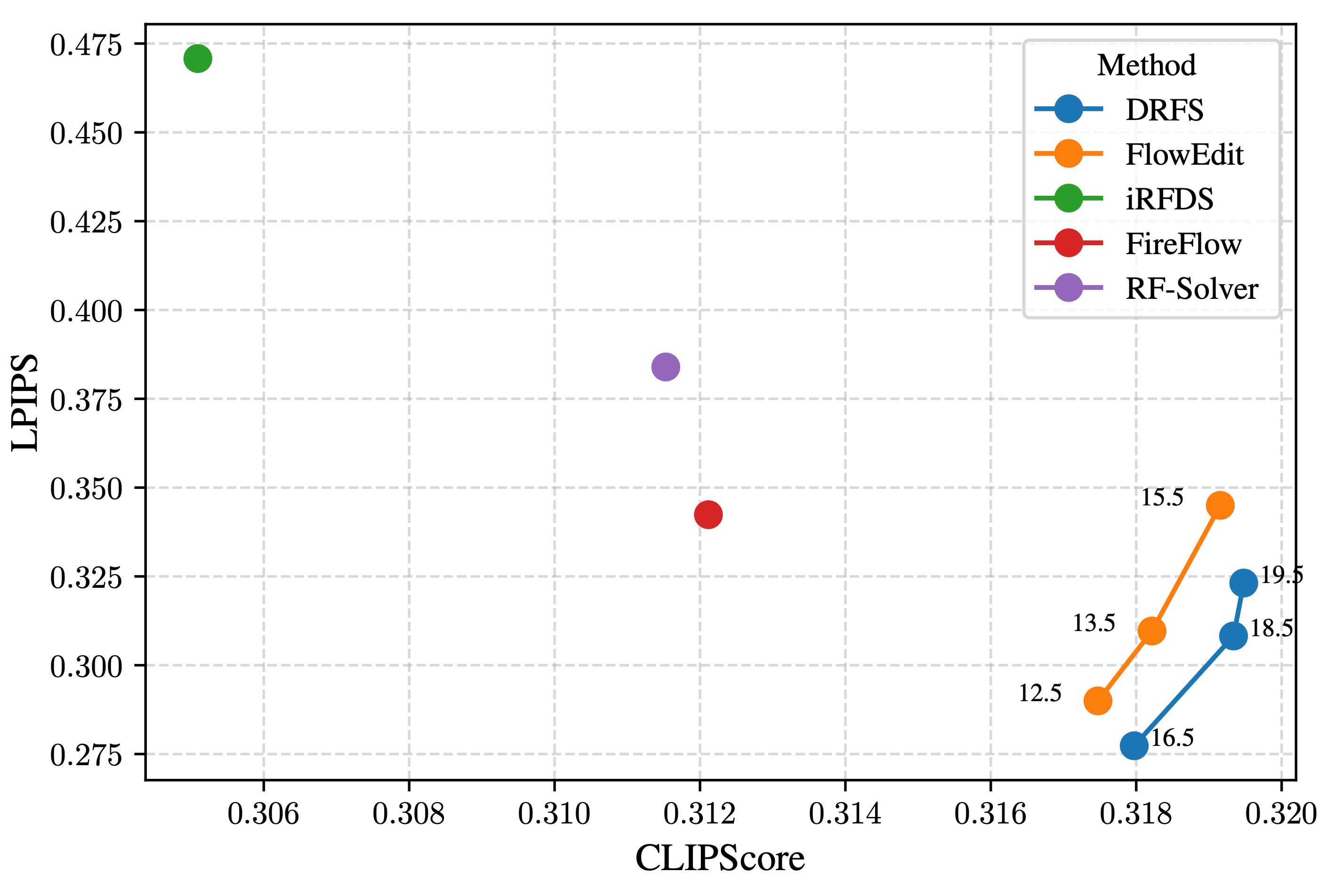}
  \caption{Comparison between LPIPS and CLIPScore on 340 source images and editing prompts. Higher CLIPScore indicates better semantic alignment with the target prompt, while lower LPIPS indicates better perceptual similarity to the reference image.}
    \label{fig:lpips-clip}

\end{figure}

On the additional dataset described in \ref{sec:add_dataset}, we measured LPIPS and CLIP similarity for several methods across a range of target CFG scales (directly written on Fig. \ref{fig:lpips-clip}). As the plot shows, DRFS consistently outperforms all baselines, striking the best balance between background preservation and adherence to the target prompt.

\subsection{More qualitative results}

\begin{figure*}[ht]
  \centering
  \begin{subfigure}[t]{0.33\linewidth} %
    \centering
    \includegraphics[width=0.49\linewidth]{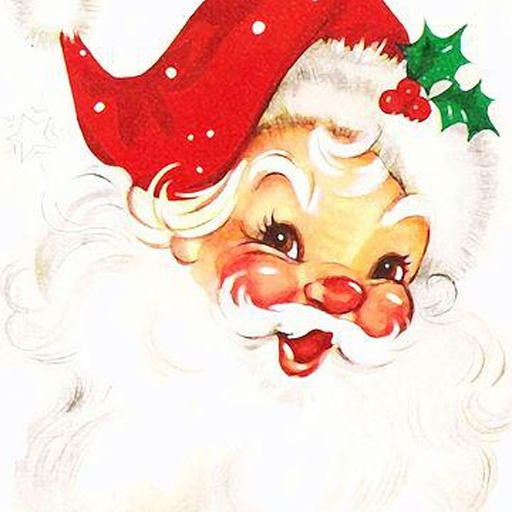}%
    \hspace{1pt}%
    \includegraphics[width=0.49\linewidth]{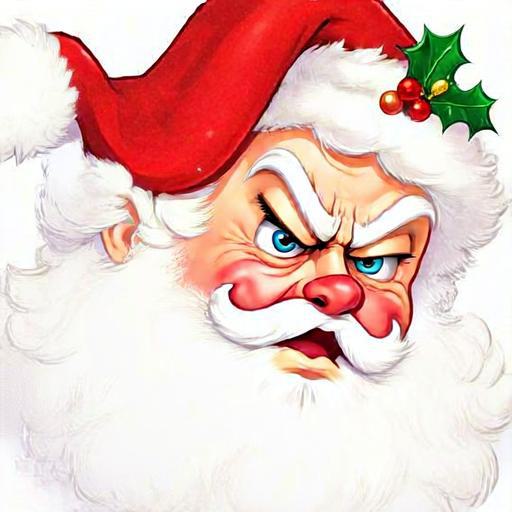}
    \caption*{\textbf{laughing face $\to$ angry face}}
  \end{subfigure}\hfill %
  \begin{subfigure}[t]{0.33\linewidth}
    \centering
    \includegraphics[width=0.49\linewidth]{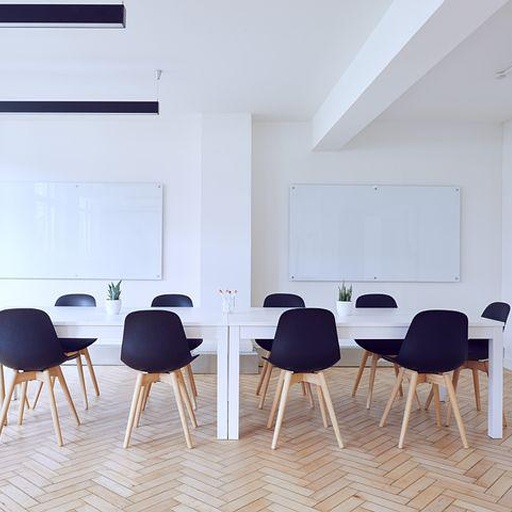}%
    \hspace{1pt}%
    \includegraphics[width=0.49\linewidth]{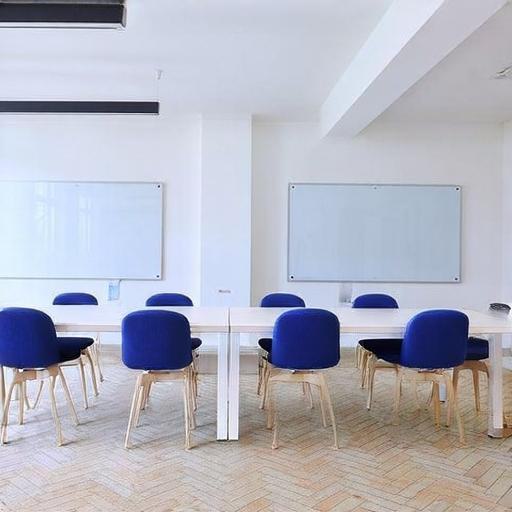}
    \caption*{\textbf{black chairs $\to$ blue chairs}}
  \end{subfigure}\hfill
    \begin{subfigure}[t]{0.33\linewidth}
    \centering
    \includegraphics[width=0.49\linewidth]{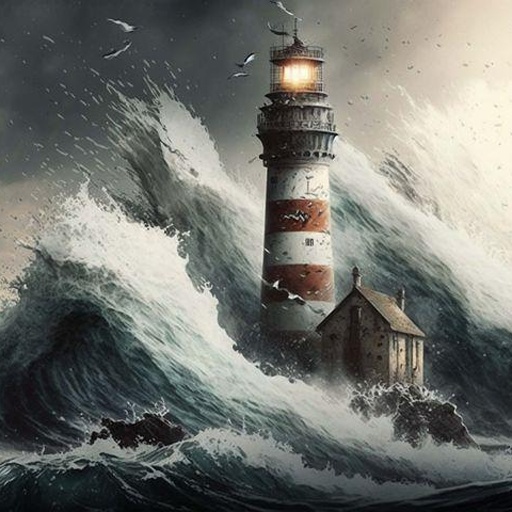}%
    \hspace{1pt}%
    \includegraphics[width=0.49\linewidth]{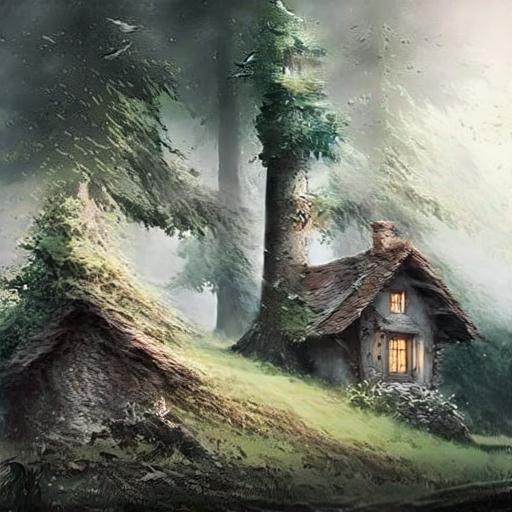}
    \caption*{\textbf{sea and house $\to$ forest and house}}
  \end{subfigure}

  \vspace{4pt} %

    \begin{subfigure}[t]{0.33\linewidth}
    \centering
    \includegraphics[width=0.49\linewidth]{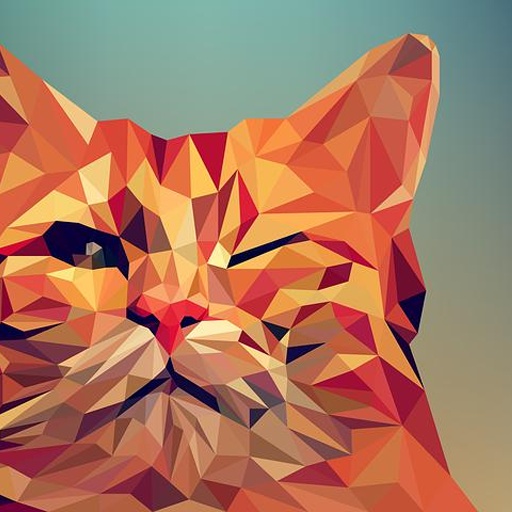}%
    \hspace{1pt}%
    \includegraphics[width=0.49\linewidth]{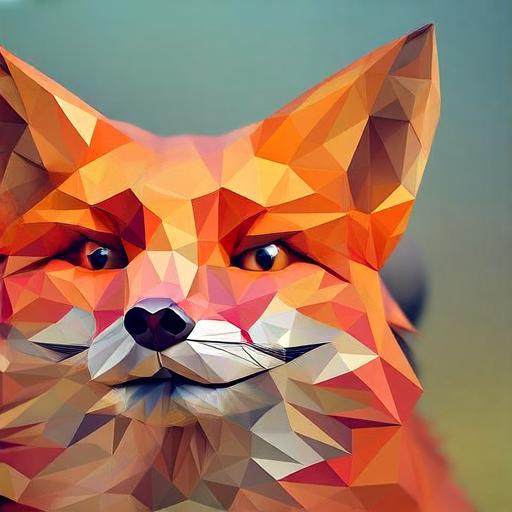}
    \caption*{\textbf{cat $\to$ fox}}
  \end{subfigure}\hfill
    \begin{subfigure}[t]{0.33\linewidth}
    \centering
    \includegraphics[width=0.49\linewidth]{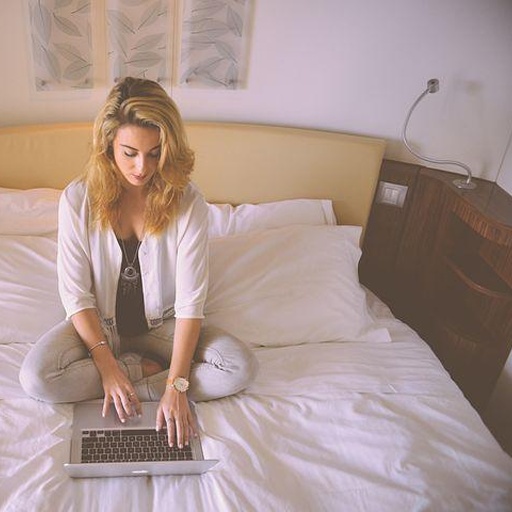}%
    \hspace{1pt}%
    \includegraphics[width=0.49\linewidth]{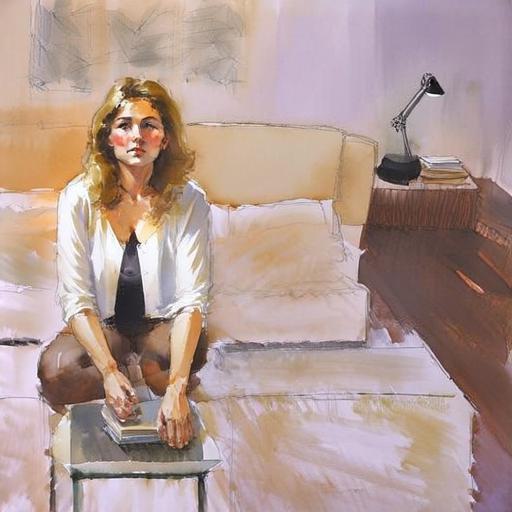}
    \caption*{\textbf{+ watercolor}}
  \end{subfigure}
  \begin{subfigure}[t]{0.33\linewidth}
    \centering
    \includegraphics[width=0.49\linewidth]{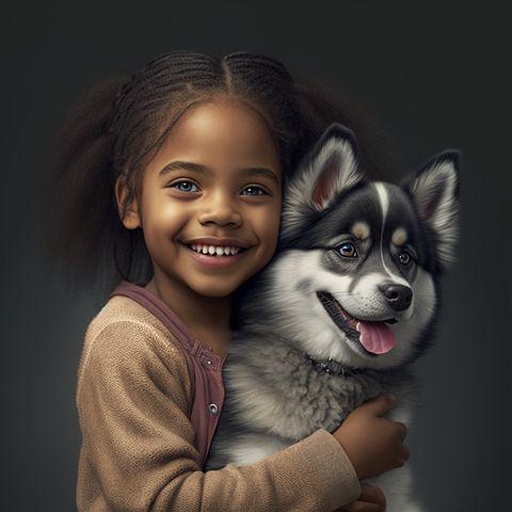}%
    \hspace{1pt}%
    \includegraphics[width=0.48\linewidth]{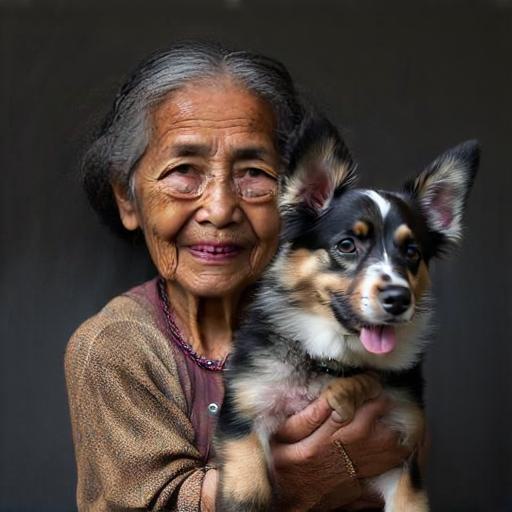}
    \caption*{\textbf{young $\to$ old}}
  \end{subfigure}

    \vspace{4pt} %

    \begin{subfigure}[t]{0.33\linewidth}
    \centering
    \includegraphics[width=0.49\linewidth]{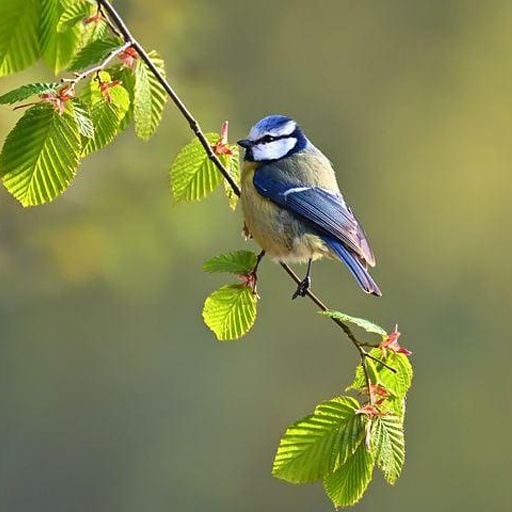}%
    \hspace{1pt}%
    \includegraphics[width=0.49\linewidth]{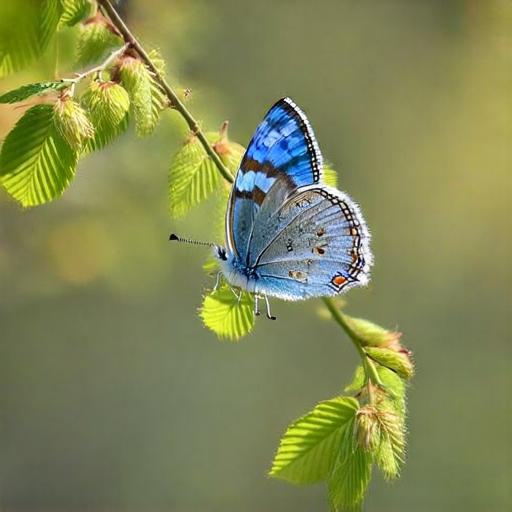}
    \caption*{\textbf{bird $\to$ butterfly}}
  \end{subfigure}\hfill
    \begin{subfigure}[t]{0.33\linewidth}
    \centering
    \includegraphics[width=0.49\linewidth]{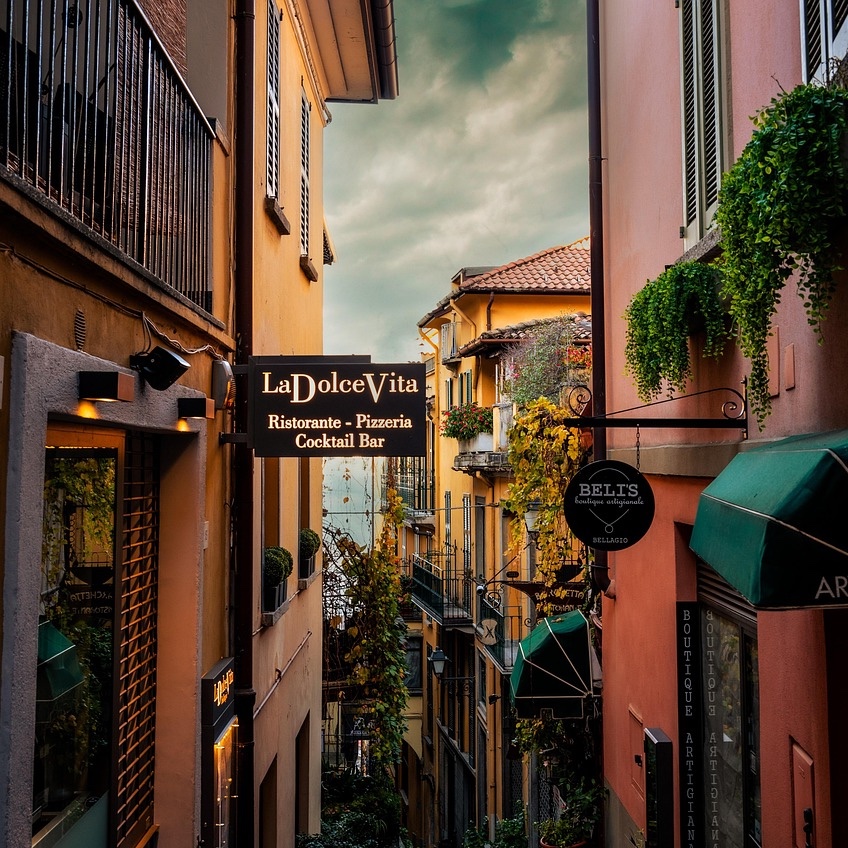}%
    \hspace{1pt}%
    \includegraphics[width=0.49\linewidth]{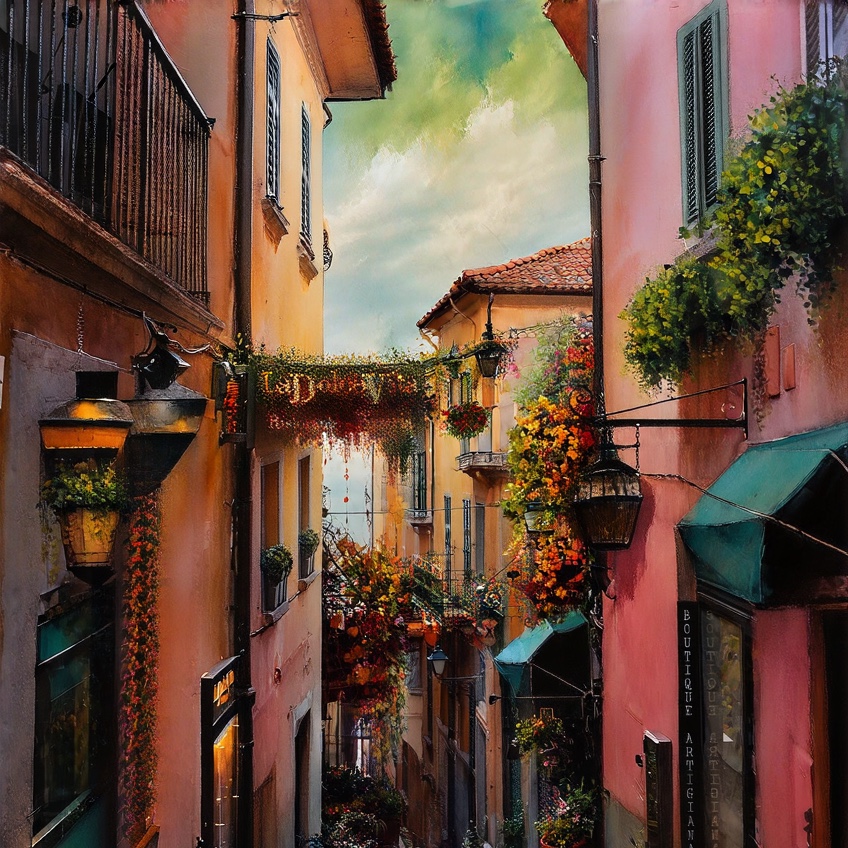}
    \caption*{\textbf{+ watercolor}}
  \end{subfigure}
  \begin{subfigure}[t]{0.33\linewidth}
    \centering
    \includegraphics[width=0.49\linewidth]{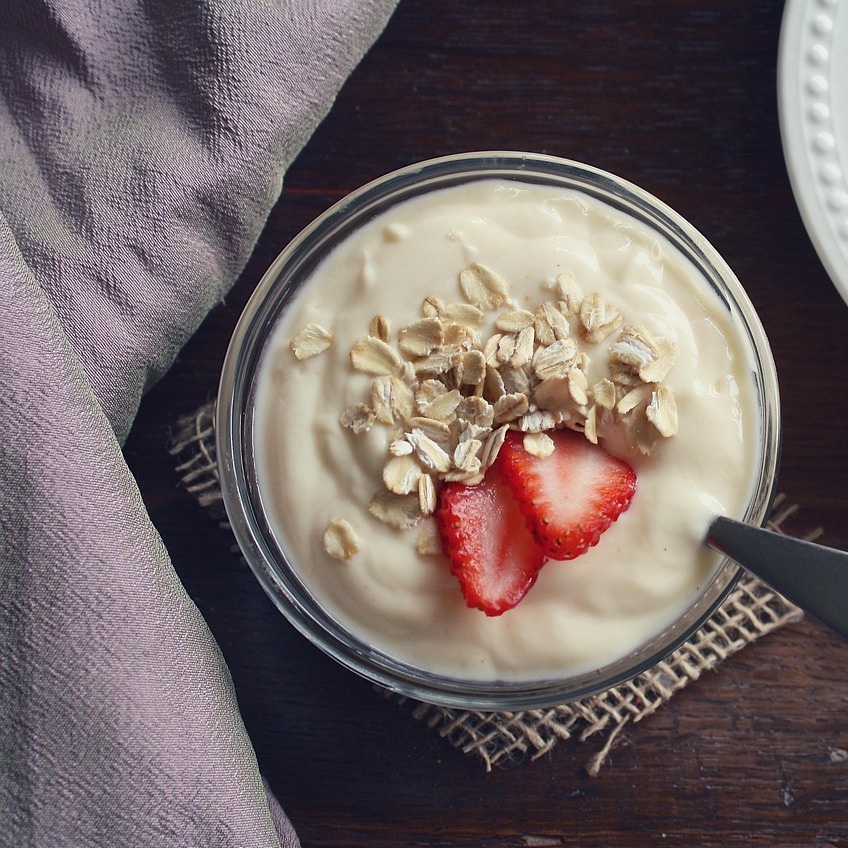}%
    \hspace{1pt}%
    \includegraphics[width=0.48\linewidth]{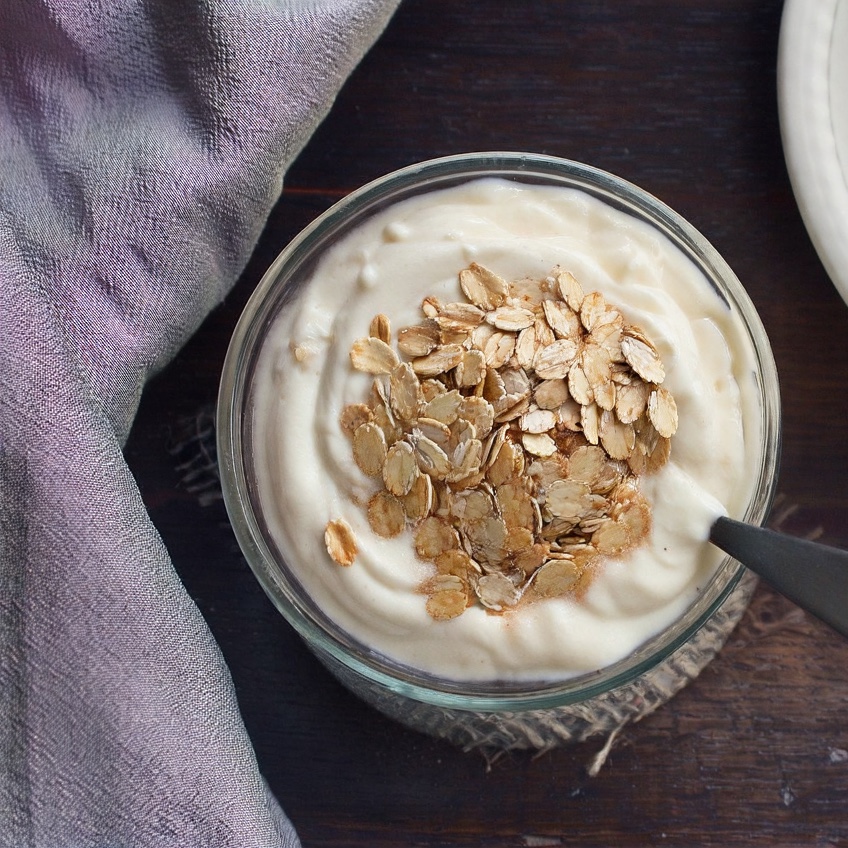}
    \caption*{\textbf{- strawberry}}
  \end{subfigure}\hfill

    \vspace{4pt} %
    \begin{subfigure}[t]{0.33\linewidth}
    \centering
    \includegraphics[width=0.49\linewidth]{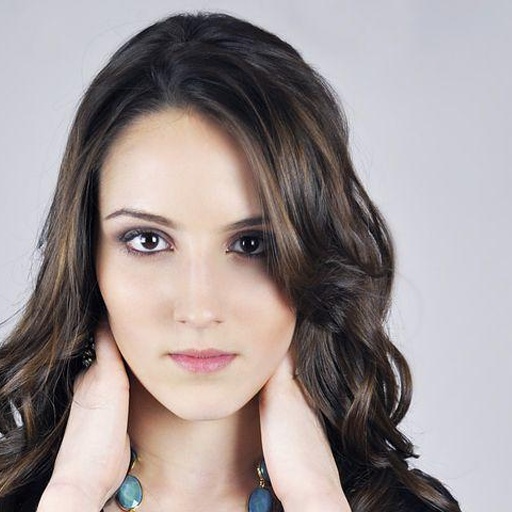}%
    \hspace{1pt}%
    \includegraphics[width=0.49\linewidth]{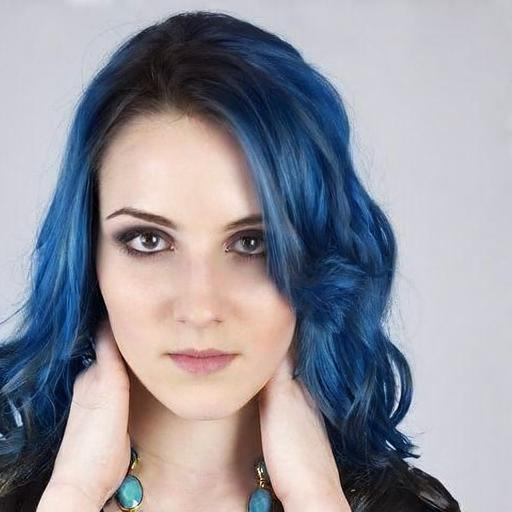}
    \caption*{\textbf{brown hair $\to$ blue hair}}
  \end{subfigure}\hfill
    \begin{subfigure}[t]{0.33\linewidth}
    \centering
    \includegraphics[width=0.49\linewidth]{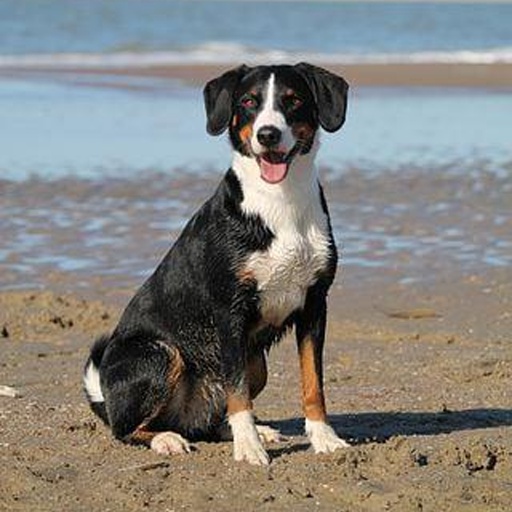}%
    \hspace{1pt}%
    \includegraphics[width=0.49\linewidth]{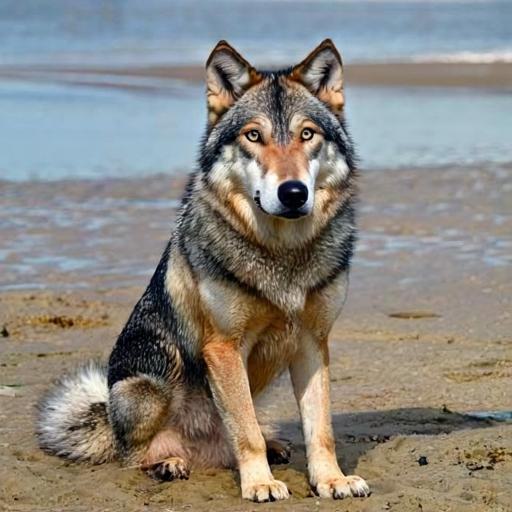}
    \caption*{\textbf{dog $\to$ wolf}}
  \end{subfigure}
  \begin{subfigure}[t]{0.33\linewidth}
    \centering
    \includegraphics[width=0.49\linewidth]{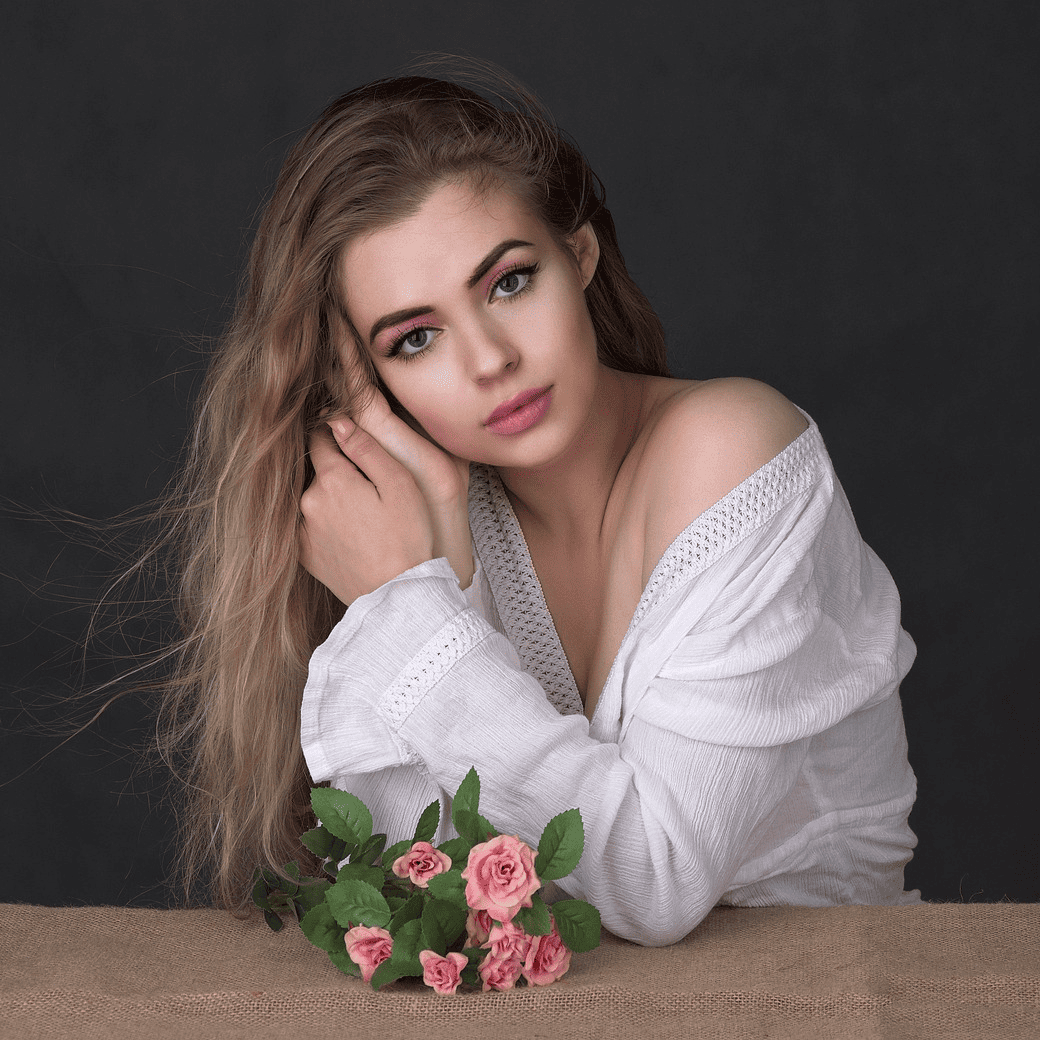}%
    \hspace{1pt}%
    \includegraphics[width=0.48\linewidth]{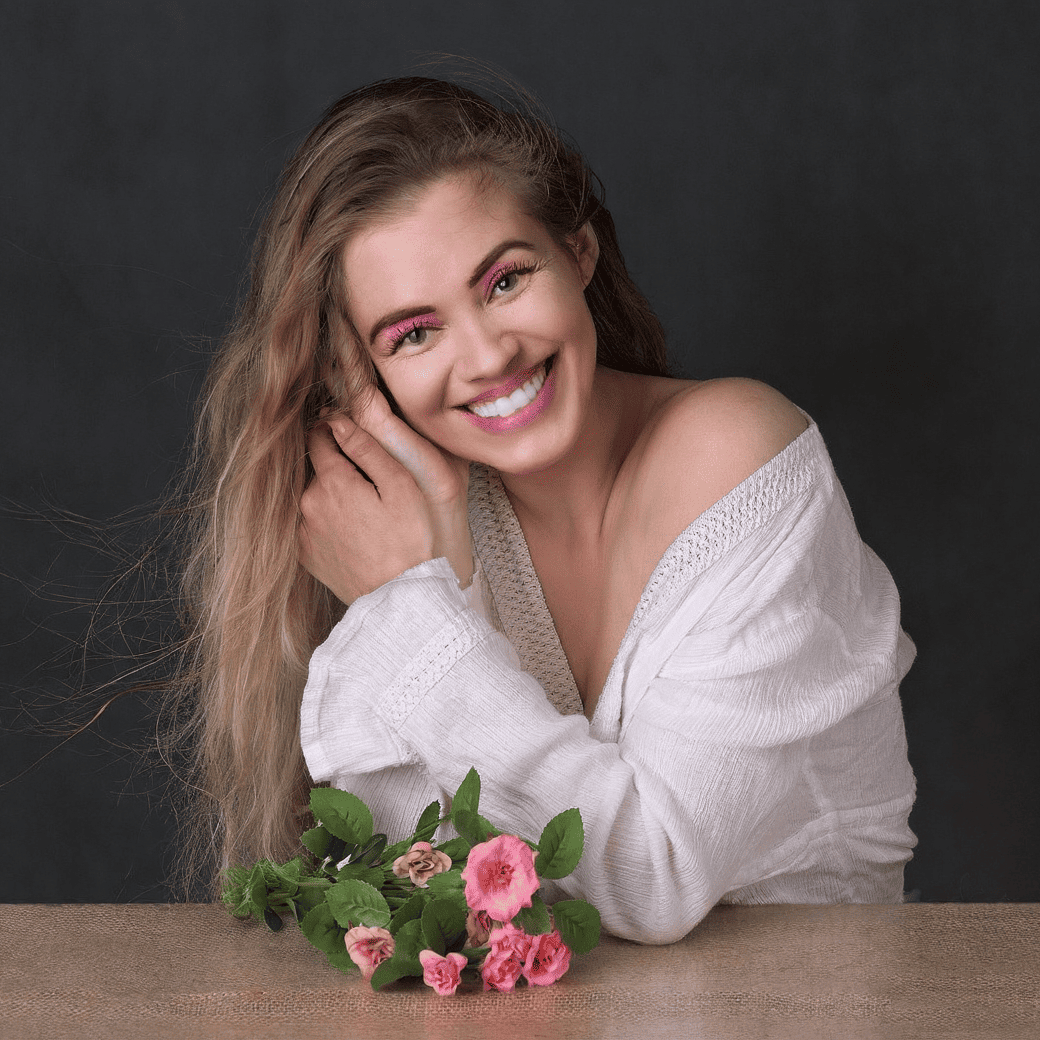}
    \caption*{\textbf{+ smile}}
  \end{subfigure}\hfill
    \vspace{4pt} %
    \begin{subfigure}[t]{0.33\linewidth}
    \centering
    \includegraphics[width=0.49\linewidth]{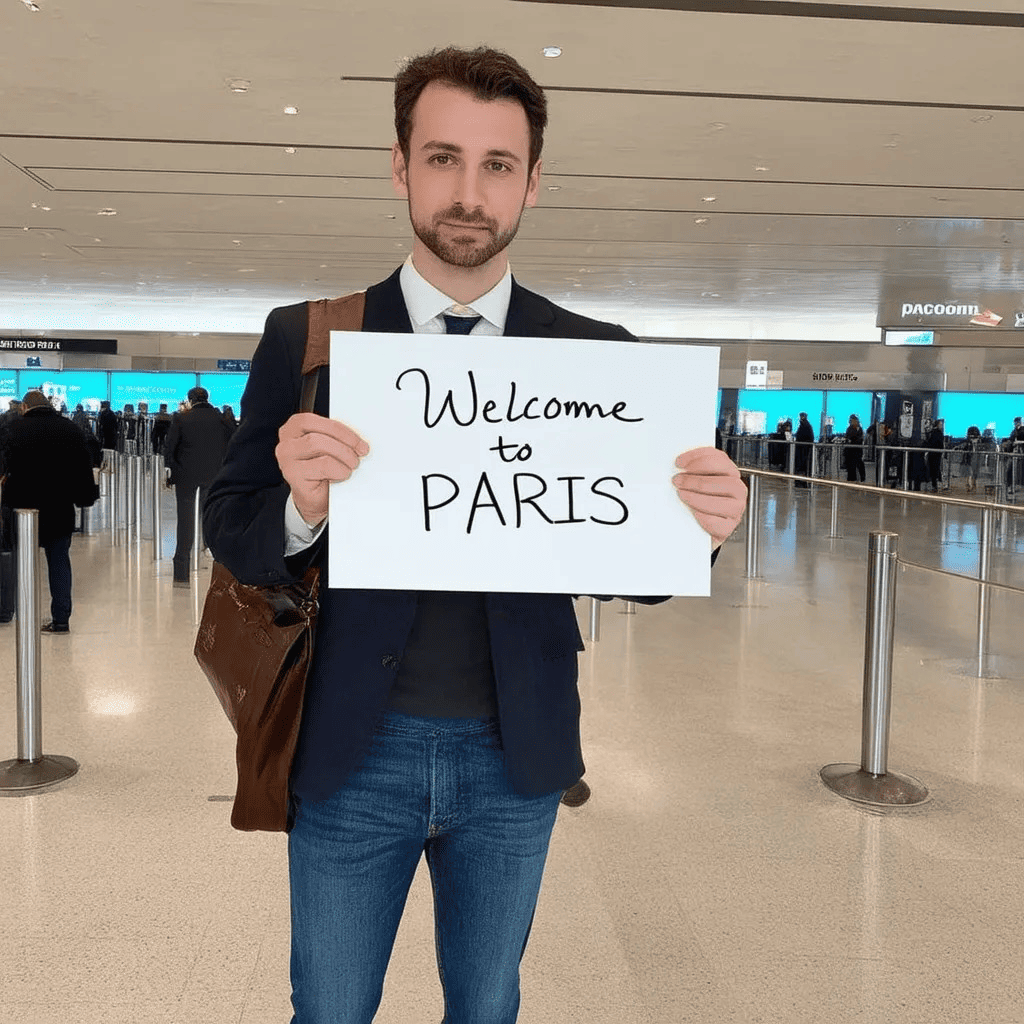}%
    \hspace{1pt}%
    \includegraphics[width=0.49\linewidth]{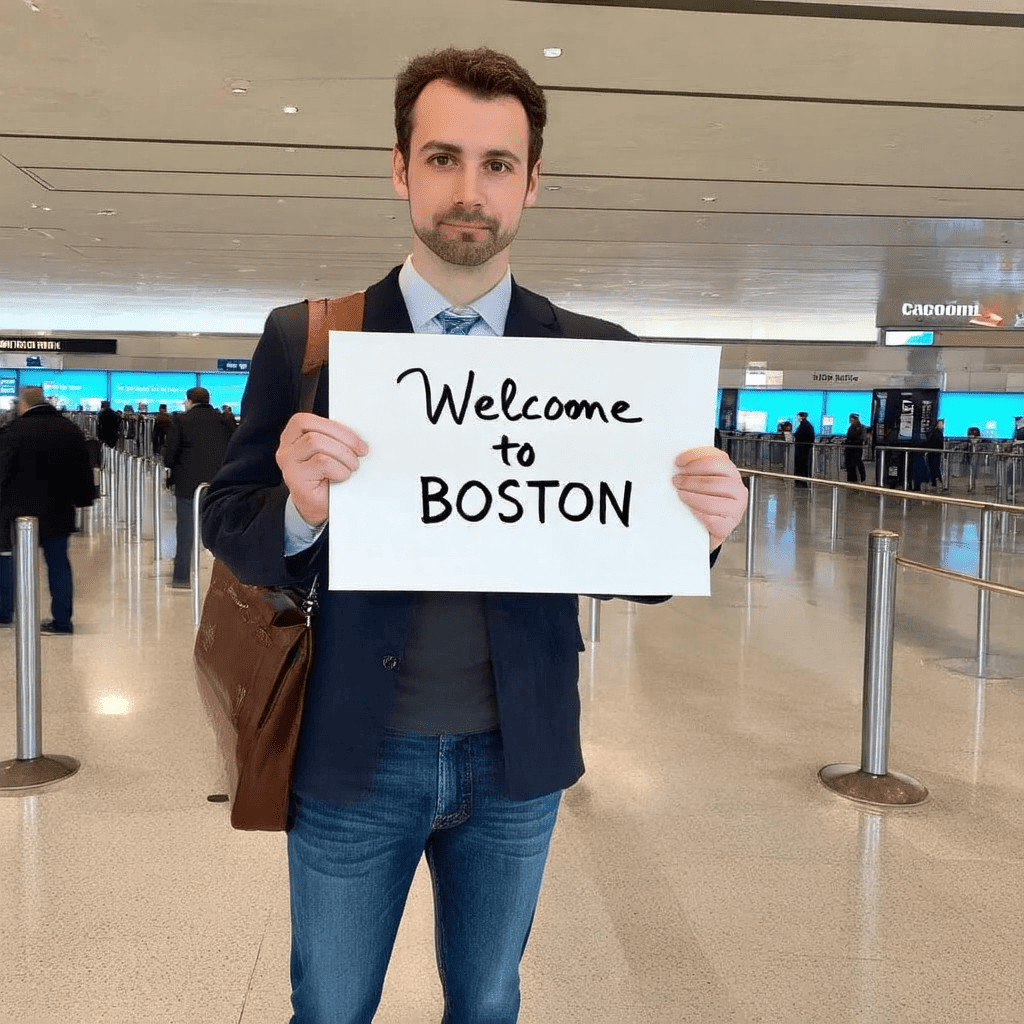}
    \caption*{\textbf{Paris $\to$ Boston}}
  \end{subfigure}\hfill
    \begin{subfigure}[t]{0.33\linewidth}
    \centering
    \includegraphics[width=0.49\linewidth]{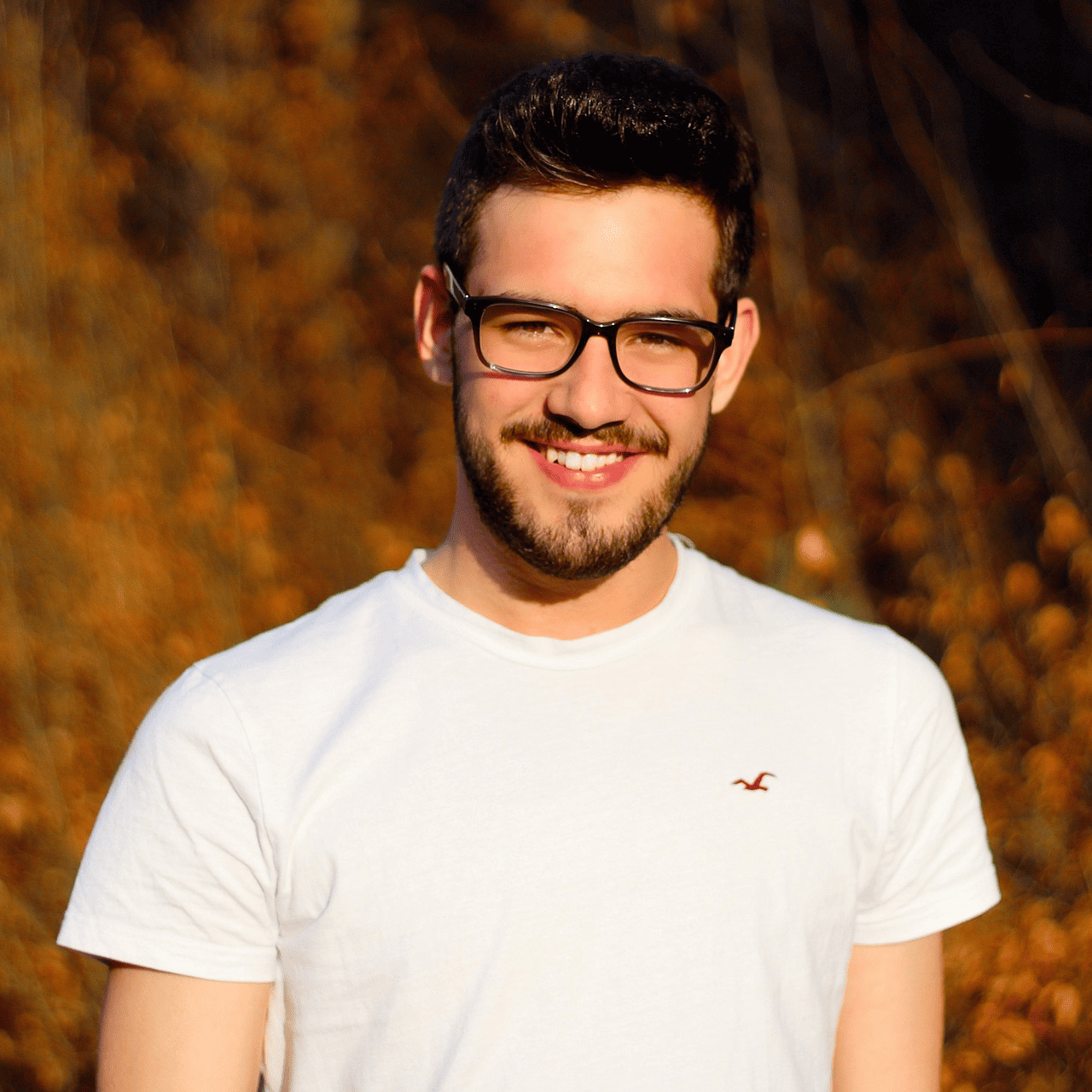}%
    \hspace{1pt}%
    \includegraphics[width=0.49\linewidth]{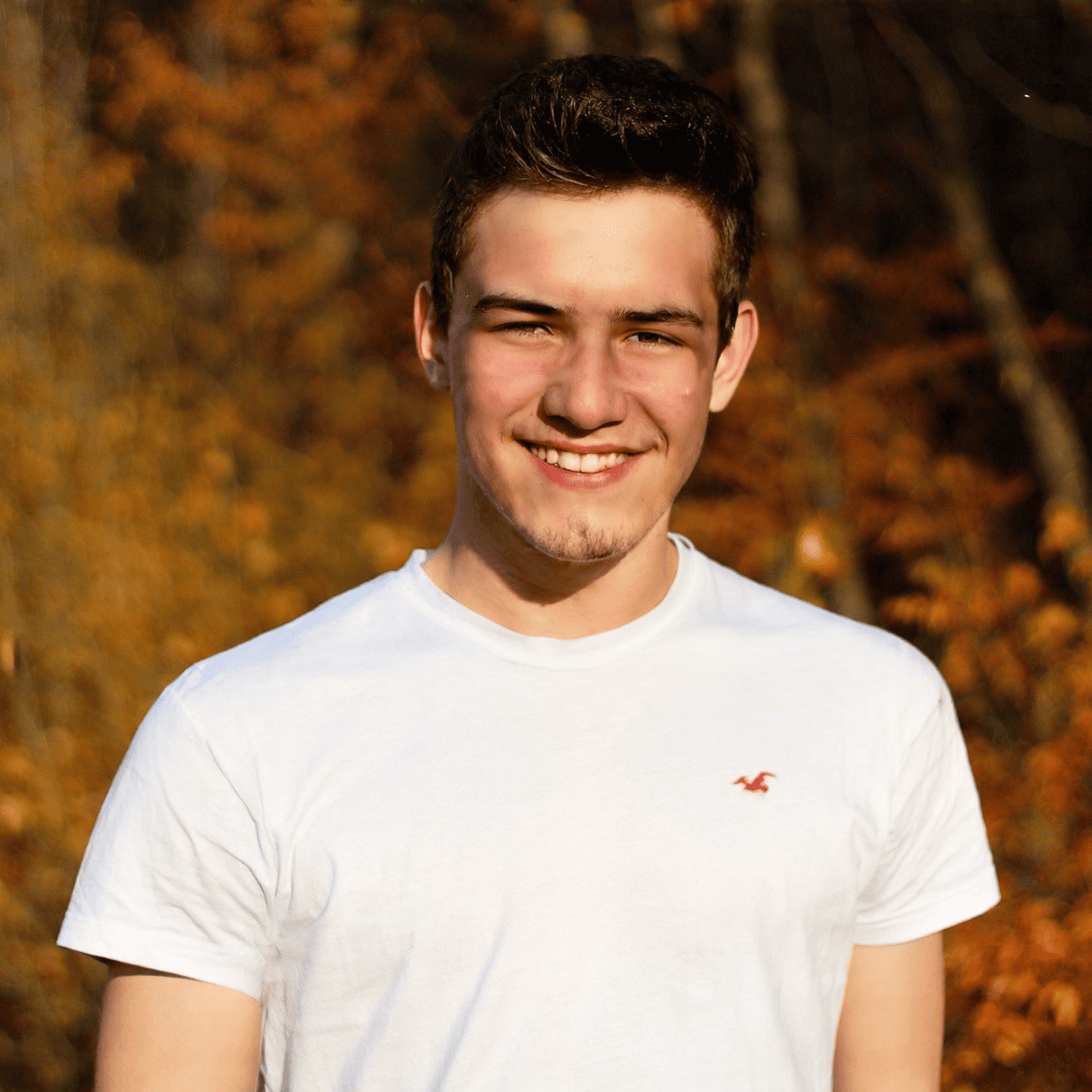}
    \caption*{\textbf{- glasses - beard}}
  \end{subfigure}
  \begin{subfigure}[t]{0.33\linewidth}
    \centering
    \includegraphics[width=0.49\linewidth]{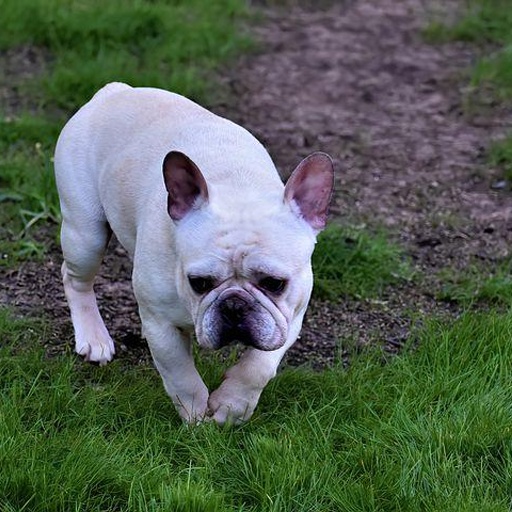}%
    \hspace{1pt}%
    \includegraphics[width=0.48\linewidth]{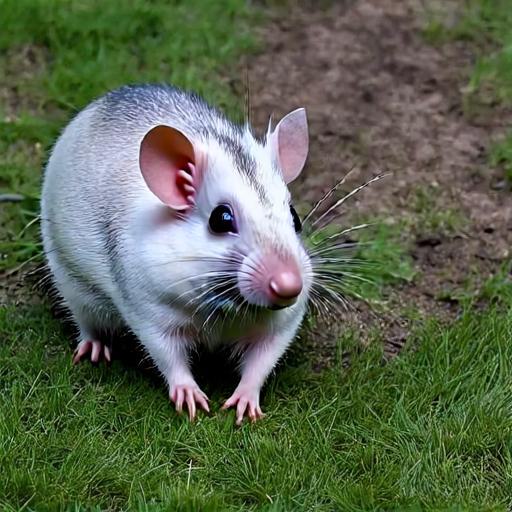}
    \caption*{\textbf{white bulldog $\to$ white rat}}
  \end{subfigure}\hfill
  \vspace{4pt} %
    \begin{subfigure}[t]{0.33\linewidth}
    \centering
    \includegraphics[width=0.49\linewidth]{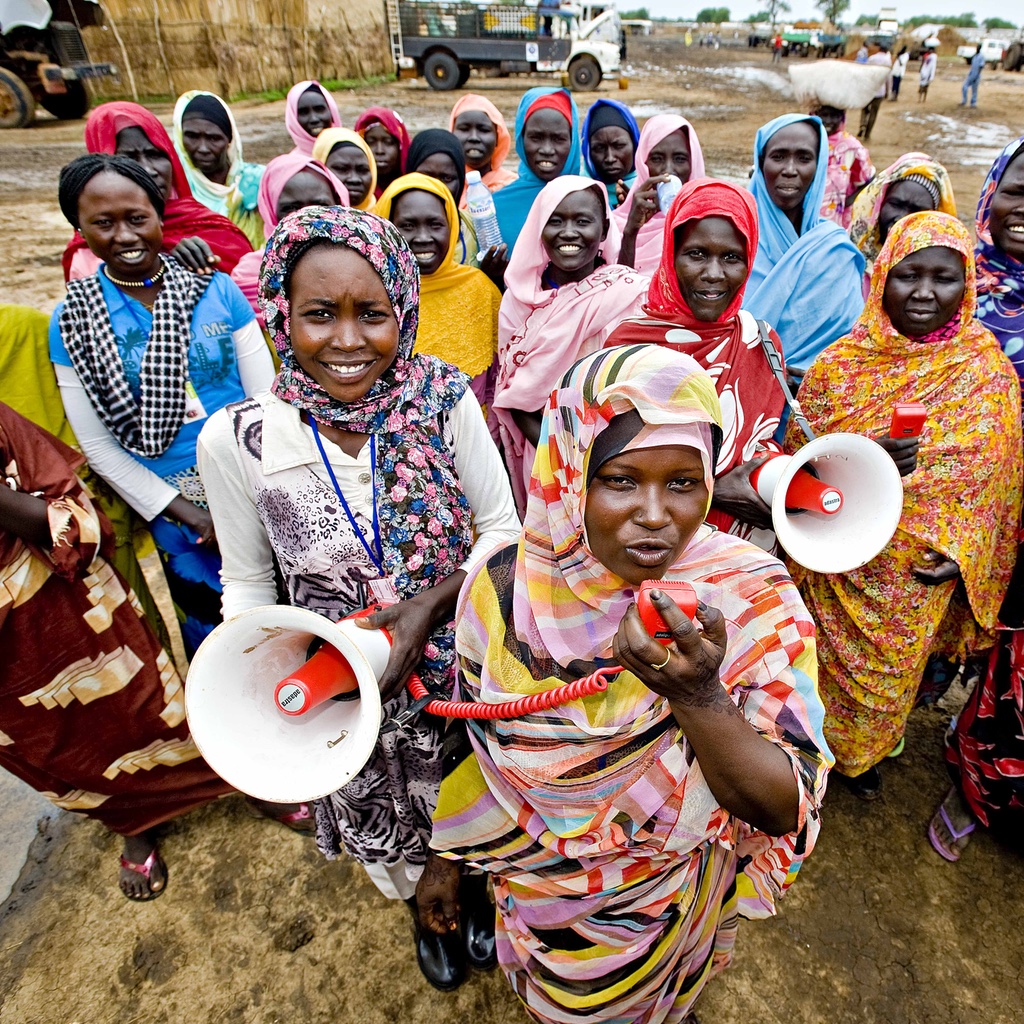}%
    \hspace{1pt}%
    \includegraphics[width=0.49\linewidth]{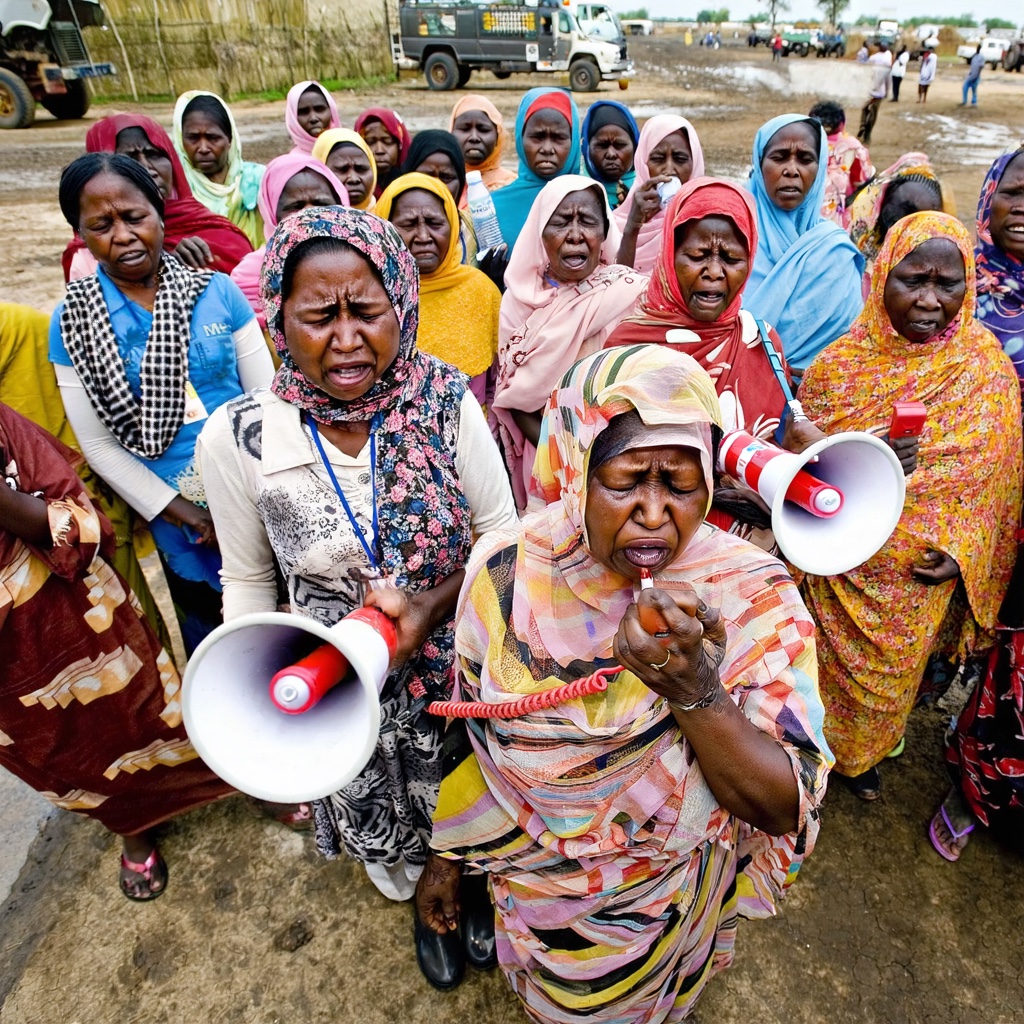}
    \caption*{\textbf{smiling $\to$ crying}}
  \end{subfigure}\hfill
    \begin{subfigure}[t]{0.33\linewidth}
    \centering
    \includegraphics[width=0.49\linewidth]{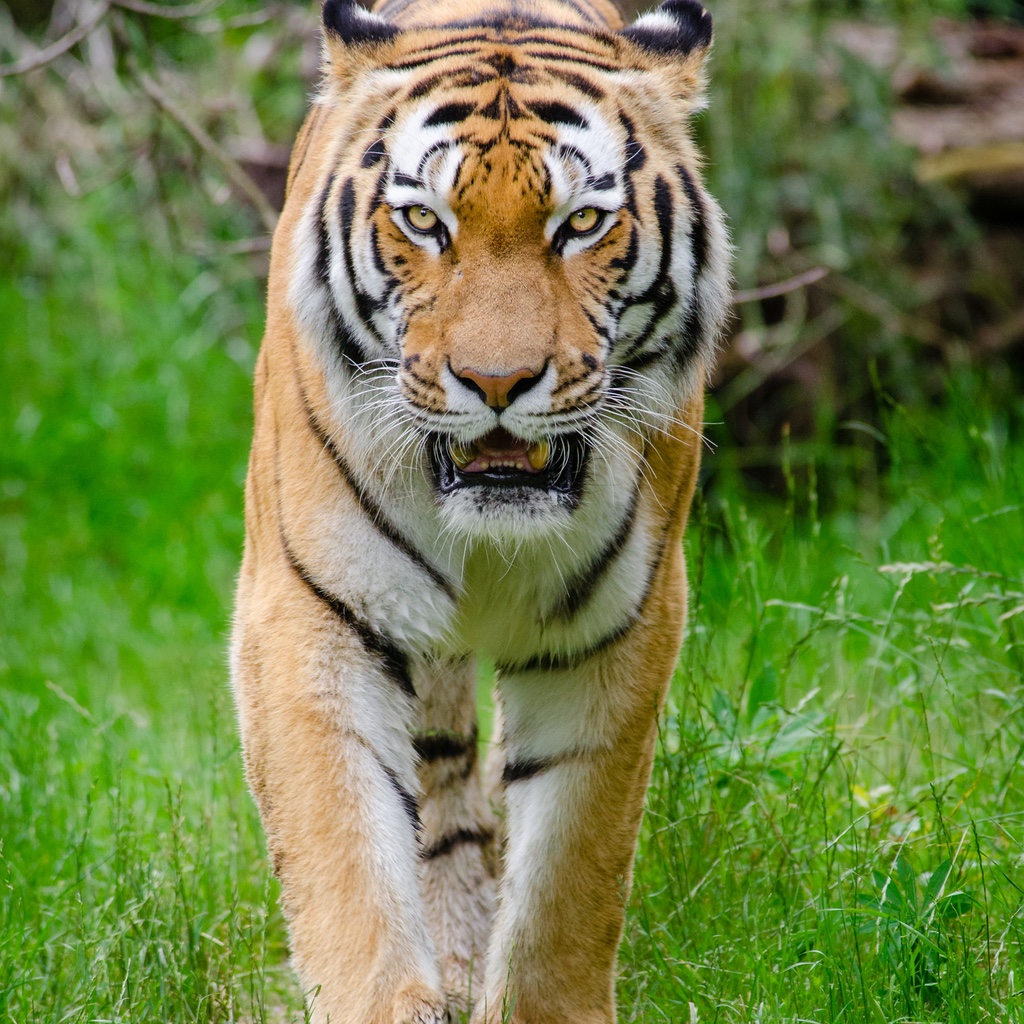}%
    \hspace{1pt}%
    \includegraphics[width=0.48\linewidth]{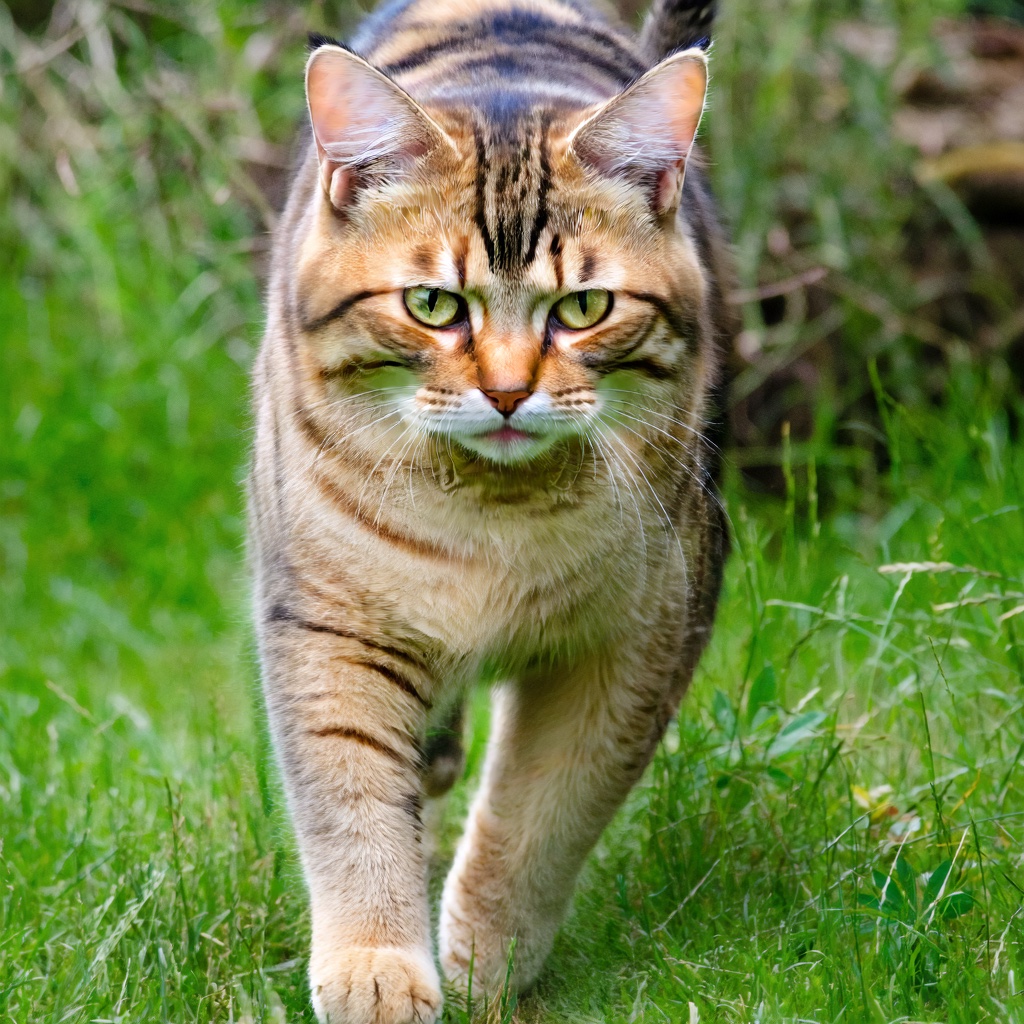}
    \caption*{\textbf{lion $\to$ cat}}
  \end{subfigure}
  \begin{subfigure}[t]{0.33\linewidth}
    \centering
    \includegraphics[width=0.49\linewidth]{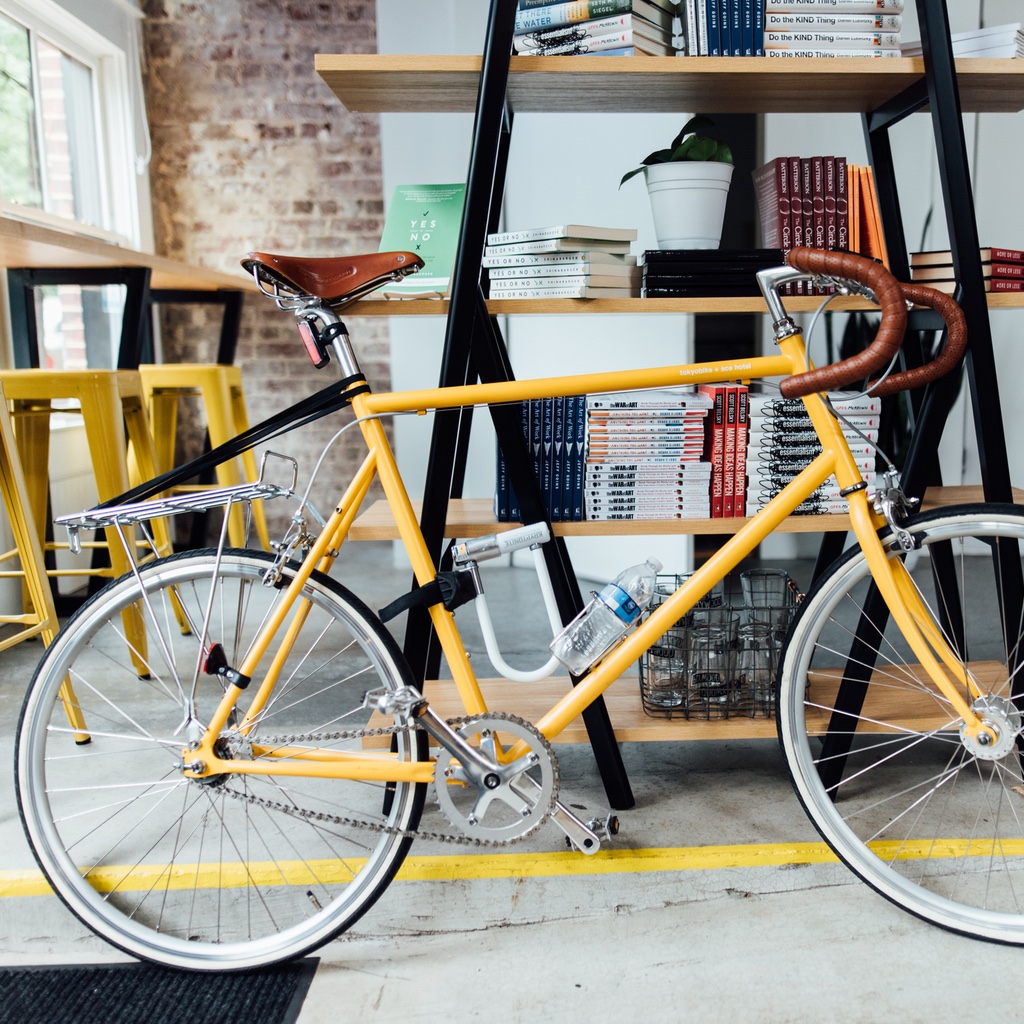}%
    \hspace{1pt}%
    \includegraphics[width=0.49\linewidth]{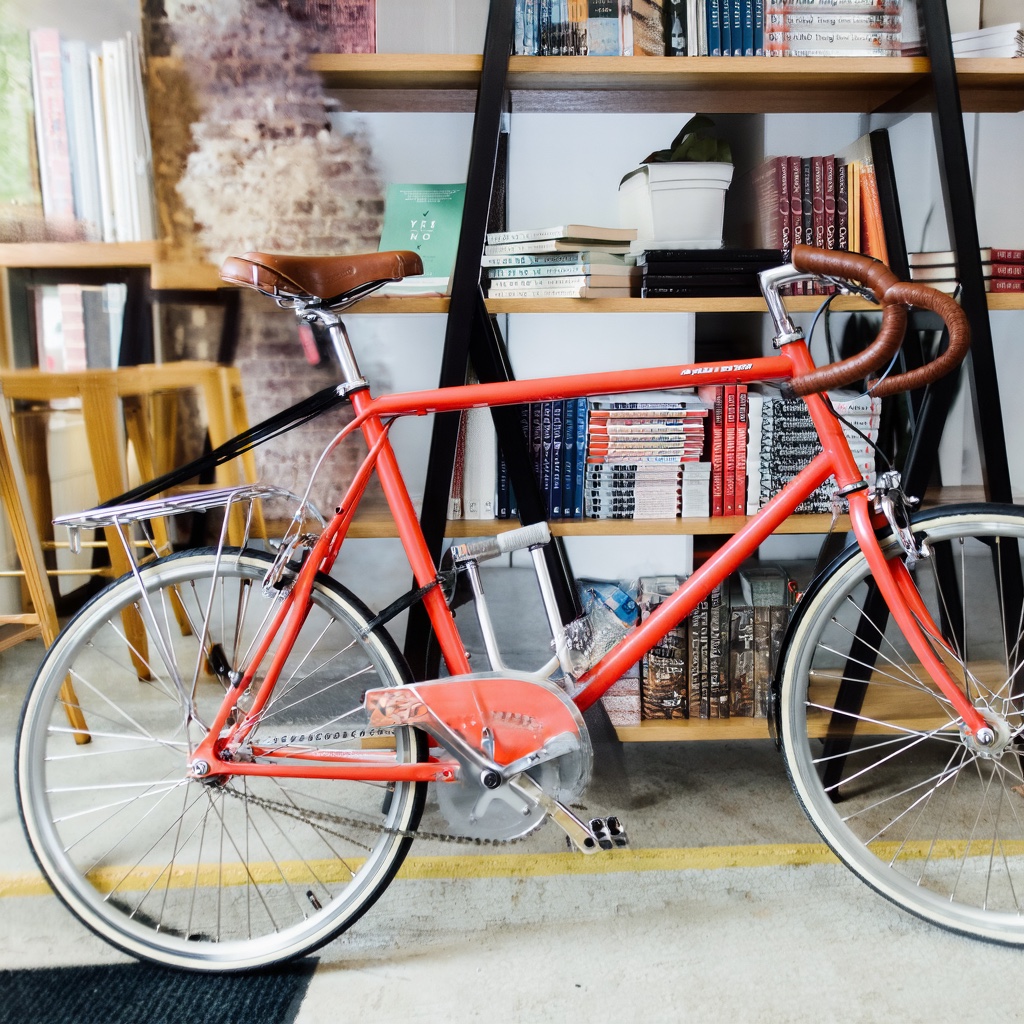}
    \caption*{\textbf{yellow bike $\to$ red bike}}
  \end{subfigure}\hfill
  \caption{Qualitative edits produced by our DRFS. Each pair indicates the source image (left) and edited result (right).}
  \label{fig:qualitative_pairs_app}
\end{figure*}

We provide additional qualitative results. First, we show editing outputs produced by our DRFS in Fig.~\ref{fig:qualitative_pairs_app}. Then, in Fig.~\ref{fig:quali-compa-app}, we present a detailed comparison on images from the PIE benchmark between FlowEdit~\cite{kulikov2024floweditinversionfreetextbasedediting}, iRFDS~\cite{yang2025texttoimage}, FireFlow~\cite{deng2025fireflow}, RF-Solver~\cite{wang2025taming}, RF Inversion~\cite{rout2025semantic}, and Direct Inversion + P2P~\cite{ju2023directinversionboostingdiffusionbased}. Results are produced using SD3.

\subsection{Structural edits}
We also report challenging category-level PIE results. DRFS consistently improves both alignment and fidelity.
\begin{table}[h]
\centering
\small
\caption{\textbf{Category-level PIE results on challenging edits (SD3).}
Lower LPIPS indicates better structural preservation, higher CLIP indicates stronger semantic alignment.}
\label{tab:pie_categories}
\resizebox{\columnwidth}{!}{
\setlength{\tabcolsep}{4.0pt}
\begin{tabular}{lcc|cc}
\toprule
& \multicolumn{2}{c|}{Change Object Pose} 
& \multicolumn{2}{c}{Change Image Style} \\
Method 
& LPIPS $\times 10^3 \downarrow$ 
& CLIP $\uparrow$ 
& LPIPS $\times 10^3 \downarrow$ 
& CLIP $\uparrow$ \\
\midrule
FlowEdit & 102.6 & 22.87 & 116.2 & 27.38 \\
\rowcolor{gray!10}
DRFS (ours) & 91.4 & 23.26 & 107.4 & 27.58 \\
\bottomrule
\end{tabular}}
\end{table}

\subsection{Comparison with Instruction-based methods.}
 We compare DRFS to prominent instruction-driven editors on PIE. DRFS achieves best overall performance \emph{without training on large paired datasets}.
\begin{table}[h]
    \label{tab:instr_pie_results}
    \vspace{-5pt}
    \centering
    \small
    \resizebox{\columnwidth}{!}{%
    \setlength{\tabcolsep}{3.6pt}
    \begin{tabular}{lcccc}
        \toprule
        Method 
        & Struct. dist. $\times 10^3 \downarrow$ 
        & LPIPS $\times 10^3 \downarrow$ 
        & SSIM $\times 10^2 \uparrow$ 
        & CLIP$_{\text{edit}} \uparrow$ \\
        \midrule
        InstructPix2Pix \citep{brooks2023instructpix2pixlearningfollowimage}& 58.80 & 159.38 & 76.75 & 21.75 \\
        MagicBrush \citep{zhang2023magicbrush}& 45.53 & \textbf{86.59} & 83.09 & 22.18 \\
        \rowcolor{gray!10}
        DRFS (ours) & \textbf{23.05} & 93.81 & \textbf{84.85} & \textbf{23.83} \\
        \bottomrule
    \end{tabular}
    }
\end{table}

\subsection{More ablation studies}

\paragraph{Batch size.}

We evaluated the effect of different batch sizes \(B\) when estimating the gradient in \eqref{eq:DRFS_grad}. Increasing \(B\) improves both structural integrity and background preservation, while maintaining comparable editing strength. In particular, the \(B=5\) setting further widens the gap against all baselines. However, since the computational cost scales linearly with \(B\), we opt for \(B=1\) throughout the paper, which already outperforms the baselines.

\begin{table*}[h]
    \caption{Impact of batch size on PIE benchmark. The best is shown in bold.}
    \label{tab:pie_results_batch}
    \vspace{-2mm}
    \centering
    \resizebox{\textwidth}{!}{%
    \setlength{\tabcolsep}{4.6pt}
    \begin{tabular}{l|c|c|c|cccc|cc}
        \toprule
        \multirow{2}{*}{\textbf{Method}} & \multirow{2}{*}{\textbf{Model}} & \multicolumn{2}{c|}{\textbf{Structure}} & \multicolumn{4}{c|}{\textbf{Background Preservation}} & \multicolumn{2}{c}{\textbf{CLIP Similarity}} \\
        \cmidrule(lr){3-10}
        & & Editing & Distance $_{\times 10^3} \downarrow$ & PSNR $\uparrow$ & LPIPS $_{\times 10^3} \downarrow$ & MSE $_{\times 10^4} \downarrow$ & SSIM{$_{\times 10^2} \uparrow$} & Whole $\uparrow$ & Edited $\uparrow$ \\
        \midrule
        \textbf{DRFS (B=1)} & SD3 & - & 23.05 & 23.38 & 93.81 & 67.49 & 84.85 & 26.90 & \best{23.83} \\
        \textbf{DRFS (B=5) }        & SD3 & – & \best{21.50} & \best{23.92} & \best{87.68} & \best{60.22} & \best{85.68} & \best{26.92} & 23.70 \\
        \bottomrule
    \end{tabular}
    }
\end{table*}

\paragraph{Optimizer.}

We compare vanilla SGD and Adam optimizers on the PIE benchmark (Table~\ref{tab:pie_results_lr}). Following Hertz et al.~\cite{Hertz_2023_ICCV}, we find that vanilla SGD produces higher‐quality edits. Using only 50 optimization steps is sufficient to obtain strong results (thanks to the straighter path and larger updates afforded by our added shift term) whereas DDS~\cite{Hertz_2023_ICCV} requires 200 steps. For a fair comparison, we also evaluate (i) SGD with a constant learning rate of 0.02 and (ii) Adam with the same rate, in both cases omitting the first few noisy steps. All other hyperparameters are identical.

We found that the following explanation made in \cite{Hertz_2023_ICCV} was also true in rectified flow models. The difference in quality arises from Adam’s adaptive normalization of gradients. For simplicity, consider the Adagrad update:
$
  \Theta_{k}\leftarrow \Theta_{k-1}
    -\alpha\,\frac{g_{k}}{\sqrt{\sum_{i=1}^{k}g_{i}^{2}}},$
where \(g_{k}\) is the gradient at step \(k\) and \(\alpha\) is the learning rate. Normalizing by the accumulated squared gradients can magnify outliers and downweight consistently informative gradients. This results in a an extremely low adherence to the target prompt in the ADAM case.

\begin{table*}[ht]
    \caption{Comparison of different optimizers on the PIE benchmark. The best and second best results are bolded and underlined, respectively.}
    \label{tab:pie_results_lr}
    \vspace{-2mm}
    \centering
    \resizebox{\textwidth}{!}{%
    \setlength{\tabcolsep}{4.6pt}
    \begin{tabular}{l|c|c|c|cccc|cc}
        \toprule
        \multirow{2}{*}{\textbf{Method}} & \multirow{2}{*}{\textbf{Model}} & \multicolumn{2}{c|}{\textbf{Structure}} & \multicolumn{4}{c|}{\textbf{Background Preservation}} & \multicolumn{2}{c}{\textbf{CLIP Similarity}} \\
        \cmidrule(lr){3-10}
        & & Editing & Distance $_{\times 10^3} \downarrow$ & PSNR $\uparrow$ & LPIPS $_{\times 10^3} \downarrow$ & MSE $_{\times 10^4} \downarrow$ & SSIM{$_{\times 10^2} \uparrow$} & Whole $\uparrow$ & Edited $\uparrow$ \\
        \midrule
        \textbf{DRFS (SGD+dynamic lr)} & SD3 & - & \second{23.05} & \second{23.38} & \second{93.81} & \second{67.49} & \second{84.85} & \best{26.90} & \best{23.83} \\
        \textbf{DRFS (SGD+constant lr)} & SD3 & – & 24.55 & 22.45 & 96.21 & 79.80 & 84.72 & \second{26.54} & \second{23.21} \\
        \best{DRFS (ADAM)}     & SD3 & – & \best{12.09} & \best{25.87} & \best{56.98} & \best{33.59} & \best{88.95} & 24.10 & 21.07 \\
        \bottomrule
    \end{tabular}
    }
\end{table*}

\section{Connection between DRFS and DDIB}
\label{appendix:secDDIB}

\paragraph{DDIB.}
Dual Diffusion Implicit Bridge (DDIB) ~\cite{su2023dual} leverage two ODEs, and was designed for image translation. Given a source images $x_0^\text{src}$, the source ODE runs in the forward direction to convert the source to the latent noise, and the reverse the ODE with target prompts then constructs target images $x_0^\text{tgt}$. The source trajectory $(x_t^\text{src})_{t\in [0,1]}$ and target trajectory $(x_t^\text{tgt})_{t\in [0,1]}$

\begin{align*}
    x_t^\text{src}&=\text{ODESolve}(x_0^\text{src}, v(\cdot, \varphi^\text{src}, t), 0,t)\\
    x_t^\text{tgt}&=\text{ODESolve}(x_1^\text{tgt}, v(\cdot, \varphi^\text{src}, t), 1,t)\\
\end{align*}
With these notations we can now write the edited image $x_0^\text{tgt}$:
\begin{align*}
x_0^\text{tgt}&=x_1^\text{src}+ \int_1^0v_{\theta}(x_t^\text{tgt}, \varphi^\text{tgt}, t)dt \\
    &=x_0^\text{src}+\int_0^1v_{\theta}(x_t^\text{src}, \varphi^\text{src}, t)dt + \int_1^0v_{\theta}(x_t^\text{tgt}, \varphi^\text{tgt}, t)dt\\
    &=x_0^\text{src} + \int_1^0 \big(v_{\theta}(x_t^\text{tgt}, \varphi^\text{tgt}, t) -  (v_{\theta}(x_t^\text{src}, \varphi^\text{src}, t) \big)dt
\end{align*}

DDIB are two concatenated Schrodinger Bridges: they traverse through two one forward and one reversed (PF-ODE is special linear or degenerate Schrodinger Bridge). DDIB has the interesting Exact Cycle Consistency property.

\paragraph{DDS sampling.}
A recent work ~\cite{Huang_2025_CVPR} proposed using diffusion weighting to define a DDS style diffusion sampling,  we define $(x_{t, DDS}^\text{tgt})_{t\in [1,0]}$ by solving an ODE starting from $x_{1, DDS}^\text{tgt}=x_0^\text{src}$:

\small
\begin{align*}
    x_{0, \mathrm{DDS}}^\text{tgt} 
    &= x_0^\text{src} + \int_0^1\! w(t)\bigl(\varepsilon_\theta(\tilde{x}_t^\text{tgt}, \varphi^\text{tgt}, t) - \varepsilon_\theta(\tilde{x}_t^\text{src}, \varphi^\text{src}, t)\bigr) dt \\
    &= x_0^\text{src} + \int_1^0\! \tilde{w}(t)\bigl(v_\theta(\tilde{x}_t^\text{tgt}, \varphi^\text{tgt}, t) - v_\theta(\tilde{x}_t^\text{src}, \varphi^\text{src}, t)\bigr) dt
\end{align*}

where $$\tilde x_t^\text{tgt}=a_t x_{t, DDS}^\text{tgt}+b_t \varepsilon \quad \text{and} \quad \tilde x_t^\text{src}=a_t x_0^\text{src}+b_t \varepsilon.$$

Hence we can clearly see that DDS is an approximation of the DDIB, with approximated $\tilde x_t^\text{tgt}$ and $\tilde x_t^\text{src}$.
Taking $\tilde x_t^\text{src}=a_t x_0^\text{src}+b_t \varepsilon$ is a relatively valid hypothesis, whereas $\tilde x_t^\text{tgt}=a_t x_{t, DDS}^\text{tgt}+b_t \varepsilon \neq x_t^\text{tgt}=a_t x_{t}^\text{tgt}+b_t \varepsilon$ is not, because for example $x_{1, DDS}^\text{tgt}=x_0^\text{src}\neq x_{0}^\text{tgt}$. $\tilde x_t^\text{tgt}$ will always be too close from the source image than the real $x_t^\text{tgt}$ from DDIB.

\paragraph{DRFS.}
To tackle this issue, we improve this approximation by adding a term $ c_t(x_{t, DDS}^\text{tgt}-x_t^\text{src})$ to push more $\tilde x_t^\text{tgt}$ toward the target distribution: $$\tilde x_t^\text{tgt}=a_t x_{t, DDS}^\text{tgt}+b_t \varepsilon + c_t(x_{t, DDS}^\text{tgt}-x_t^\text{src})$$

DRFS can be interpreted as an enhanced approximation of DDIB, with a careful designed $c_t$ in order to get strong editing perfomance while keeping  good fidelity.

\section{Limitations and Future Works}

Our current formulation is limited to rectified flow models. 

Extending DRFS into more models would be a promising direction. Additionally, DRFS could benefit from integration with existing inversion techniques, such as attention injection, to further enhance edit controllability. We also include and discuss DRFS failure cases.

\section{Broader Impact}
\label{appendix:secD}

Our work introduces a new method for editing real images using state-of-the-art text-to-image rectified flow models. A potential societal risk of this technology includes the generation and spread of misinformation, misleading imagery, or manipulated content. To mitigate such risks, we will release our code with an appropriate licence that discourages harmful uses (offensive, or dehumanizing content, or content negatively targeting individuals, communities, cultures, or religions).
Furthermore, we highlight that significant research and technological progress has recently been made towards detecting and limiting these harmful applications.

\begin{figure*}[ht]
  \centering

  \setlength{\tabcolsep}{0.4pt}
  \renewcommand{\arraystretch}{0.6}

  \newlength{\thumbsize}
  \setlength{\thumbsize}{\dimexpr(\linewidth - 18\tabcolsep)/9\relax}

  \resizebox{\textwidth}{!}{%
    \begin{tabular}{@{}*{9}{c}@{}}
      \scriptsize Source &
      \scriptsize DRFS &
      \scriptsize FlowEdit (SD3) &
      \scriptsize FlowEdit (Flux) &
      \scriptsize iRFDS &
      \scriptsize FireFlow &
      \scriptsize RF-Solver &
      \scriptsize RF-Inv &
      \scriptsize Direct+P2P \\[0.3em]

      \Thumb{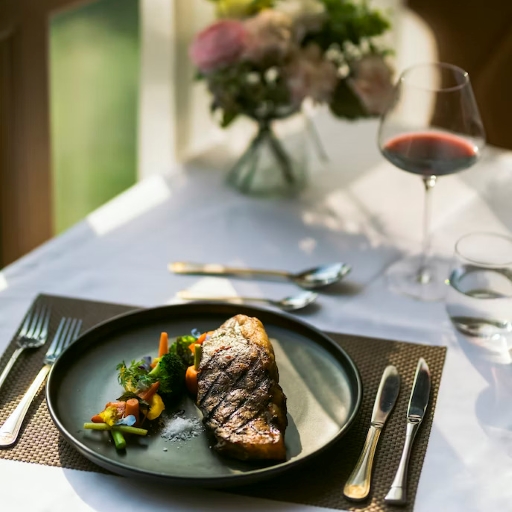} &
      \Thumb{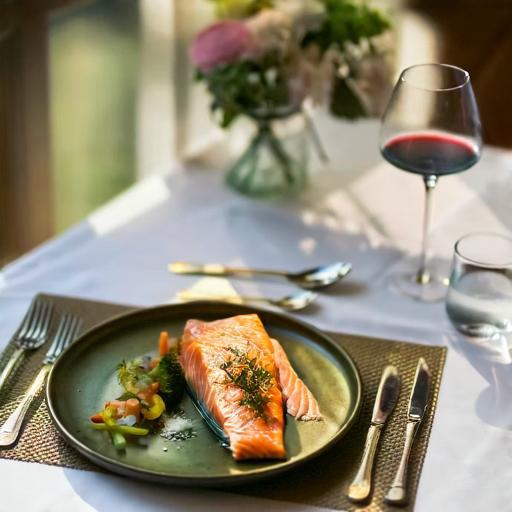} &
      \Thumb{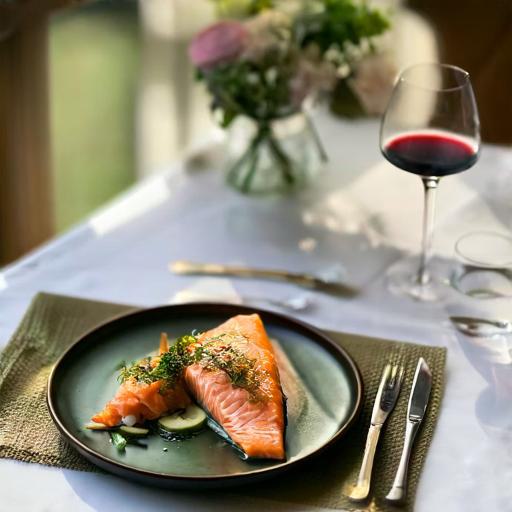} &
      \Thumb{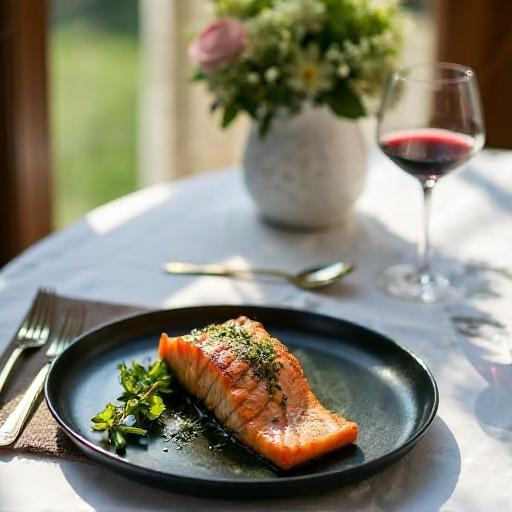} &
      \Thumb{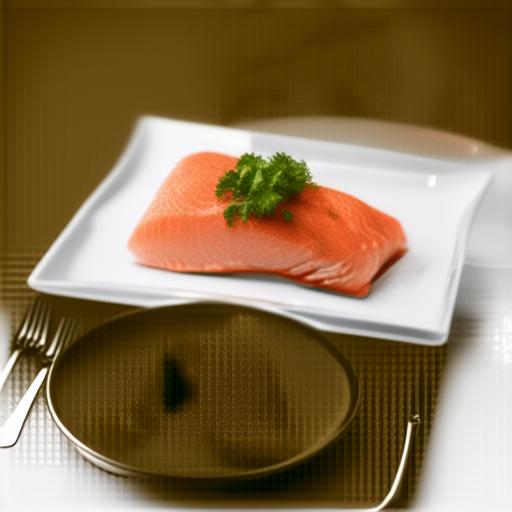} &
      \Thumb{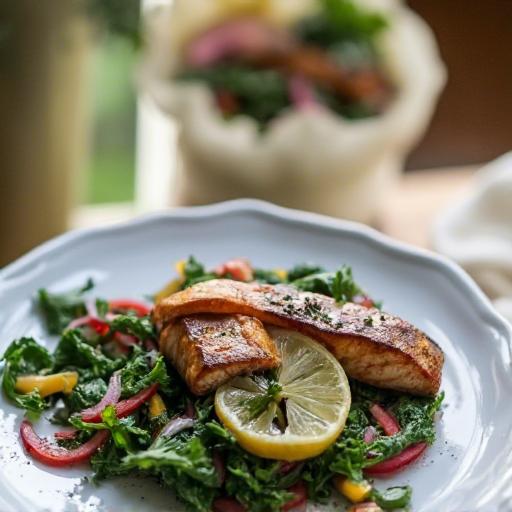} &
      \Thumb{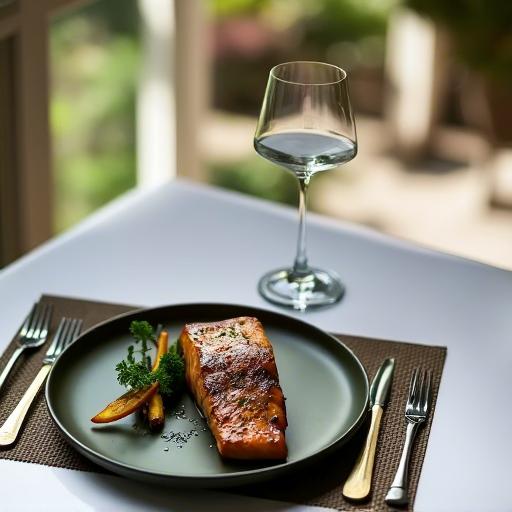} &
      \Thumb{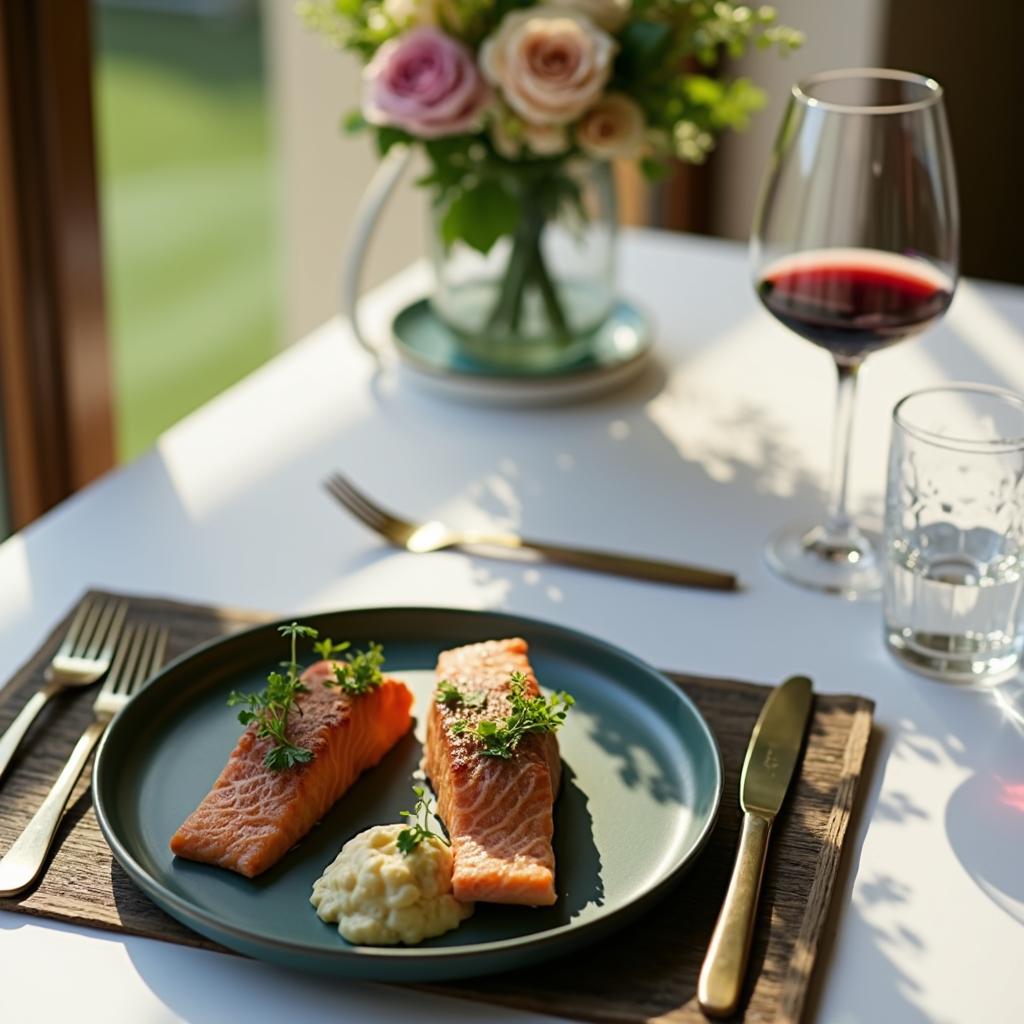} &
      \Thumb{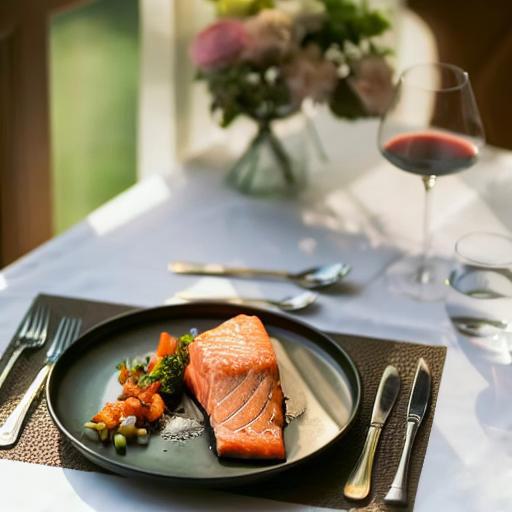}
      \\[-0.18em]
      \multicolumn{9}{c}{\scriptsize steak $\rightarrow$ salmon} \\[0.20em]

      \Thumb{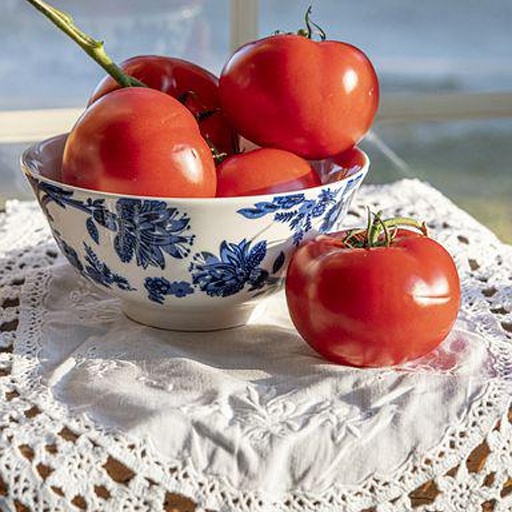} &
      \Thumb{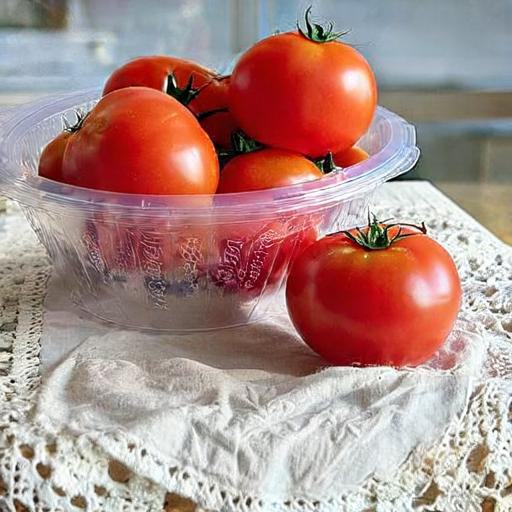} &
      \Thumb{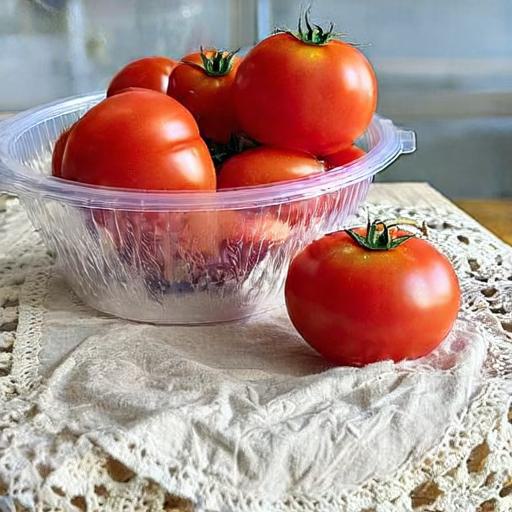} &
      \Thumb{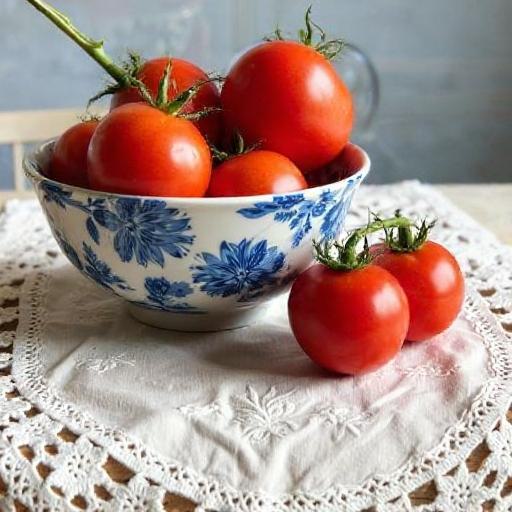} &
      \Thumb{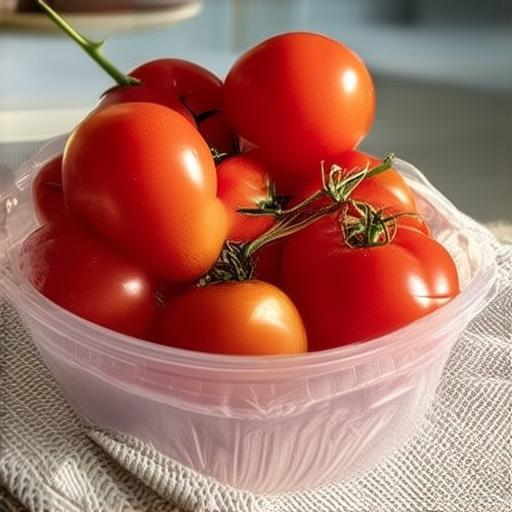} &
      \Thumb{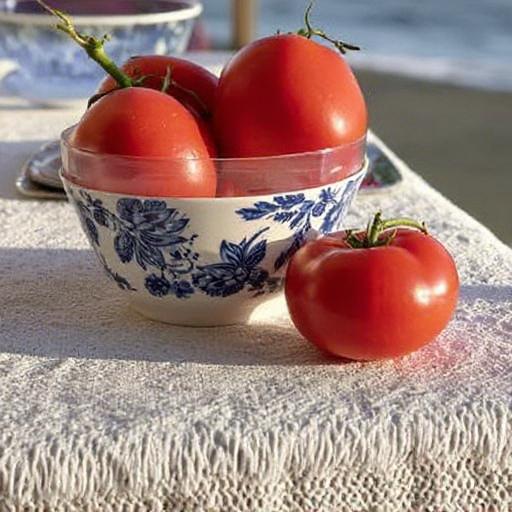} &
      \Thumb{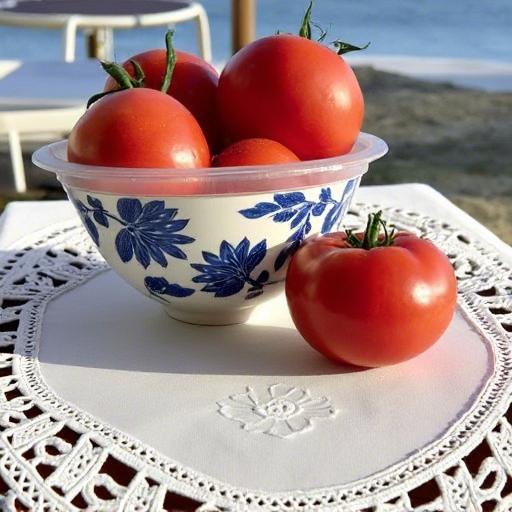} &
      \Thumb{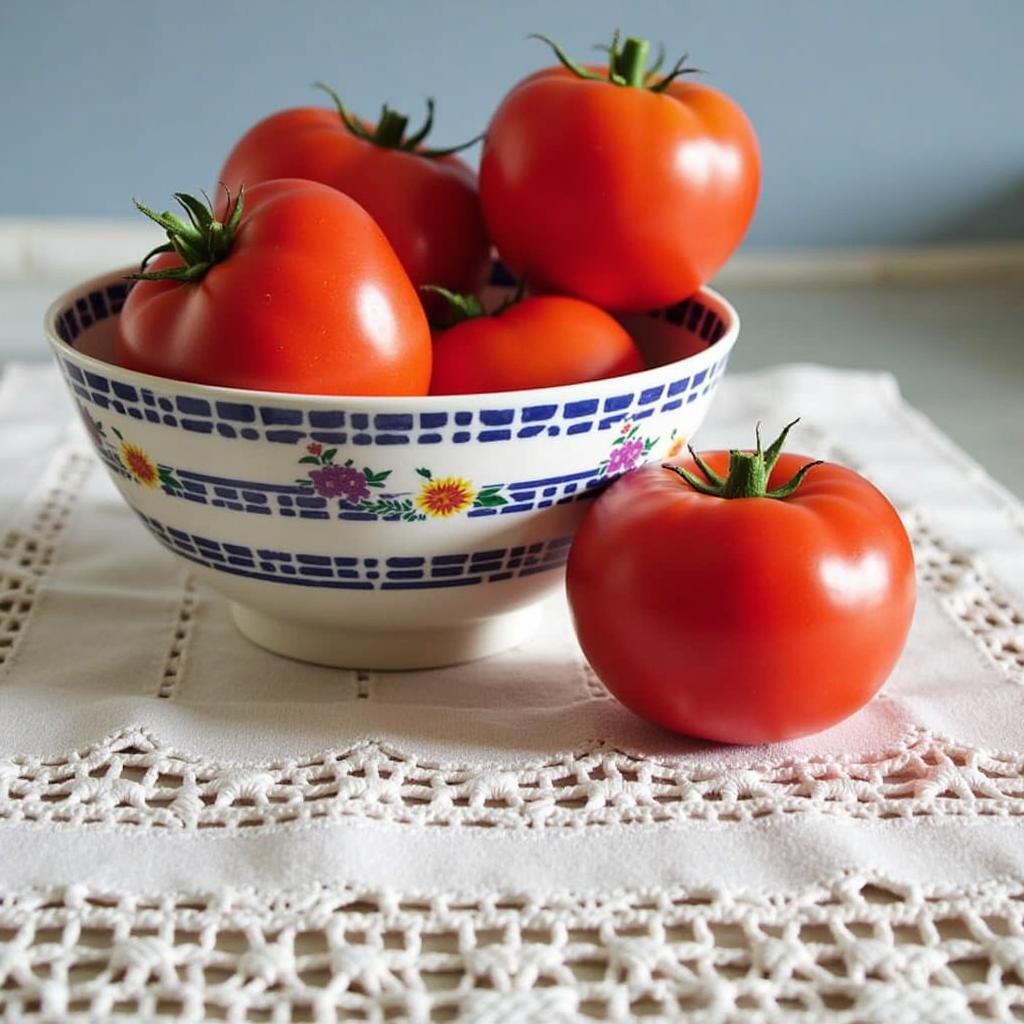} &
      \Thumb{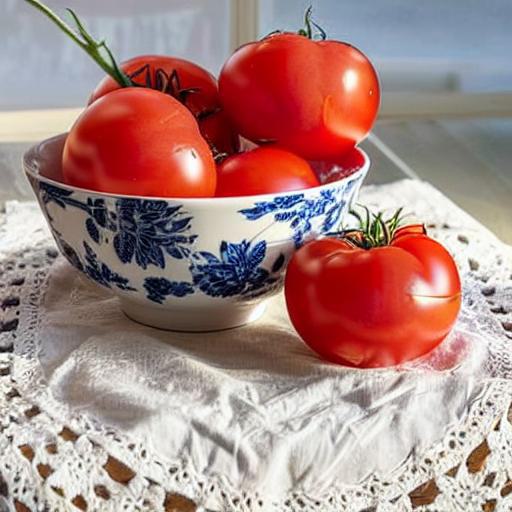}
      \\[-0.18em]
      \multicolumn{9}{c}{\scriptsize + plastic} \\[0.20em]

      \Thumb{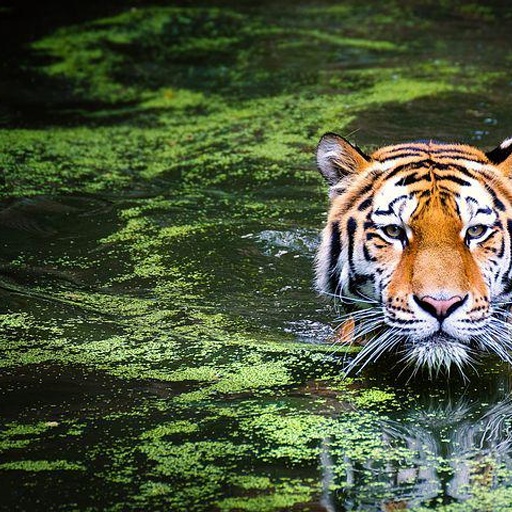} &
      \Thumb{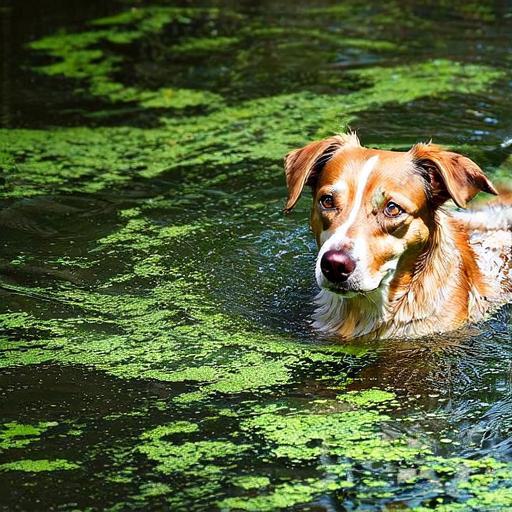} &
      \Thumb{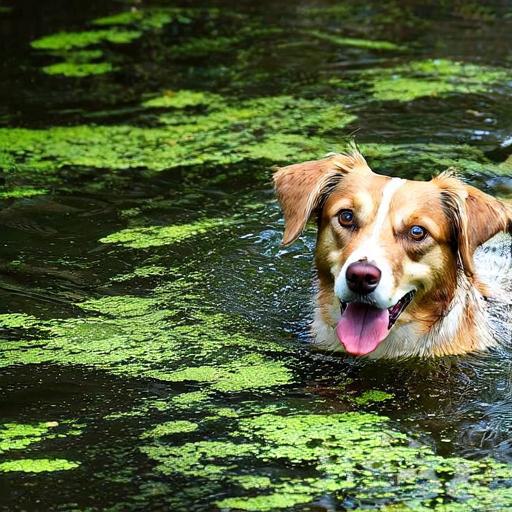} &
      \Thumb{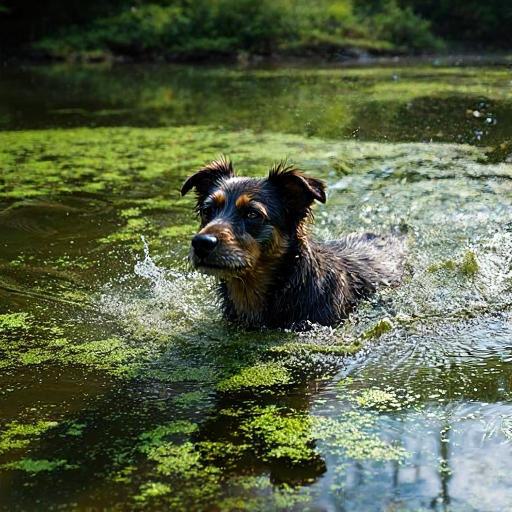} &
      \Thumb{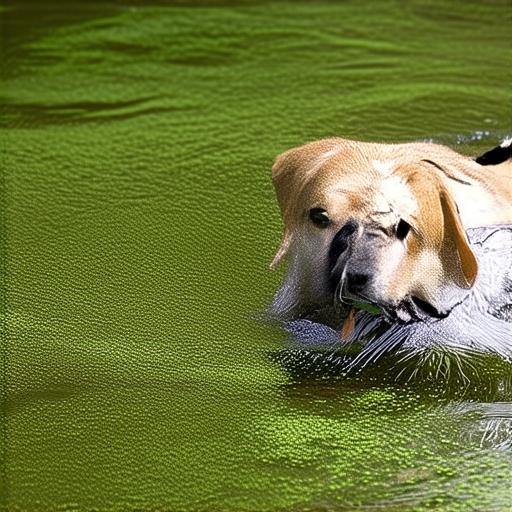} &
      \Thumb{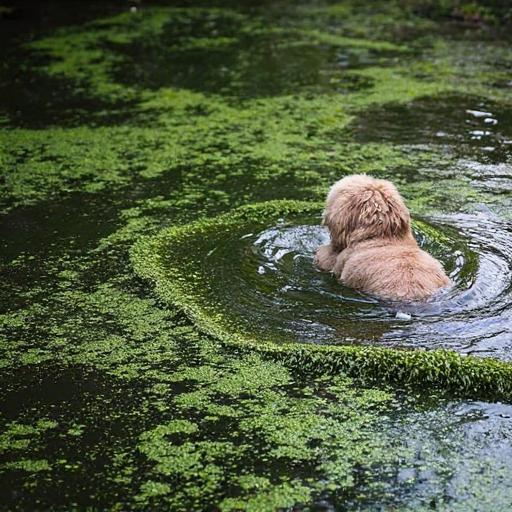} &
      \Thumb{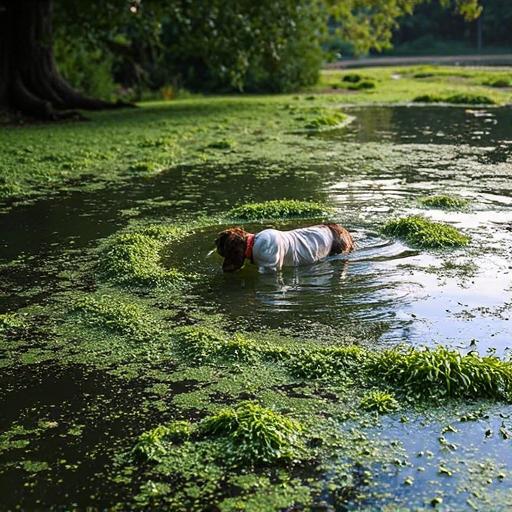} &
      \Thumb{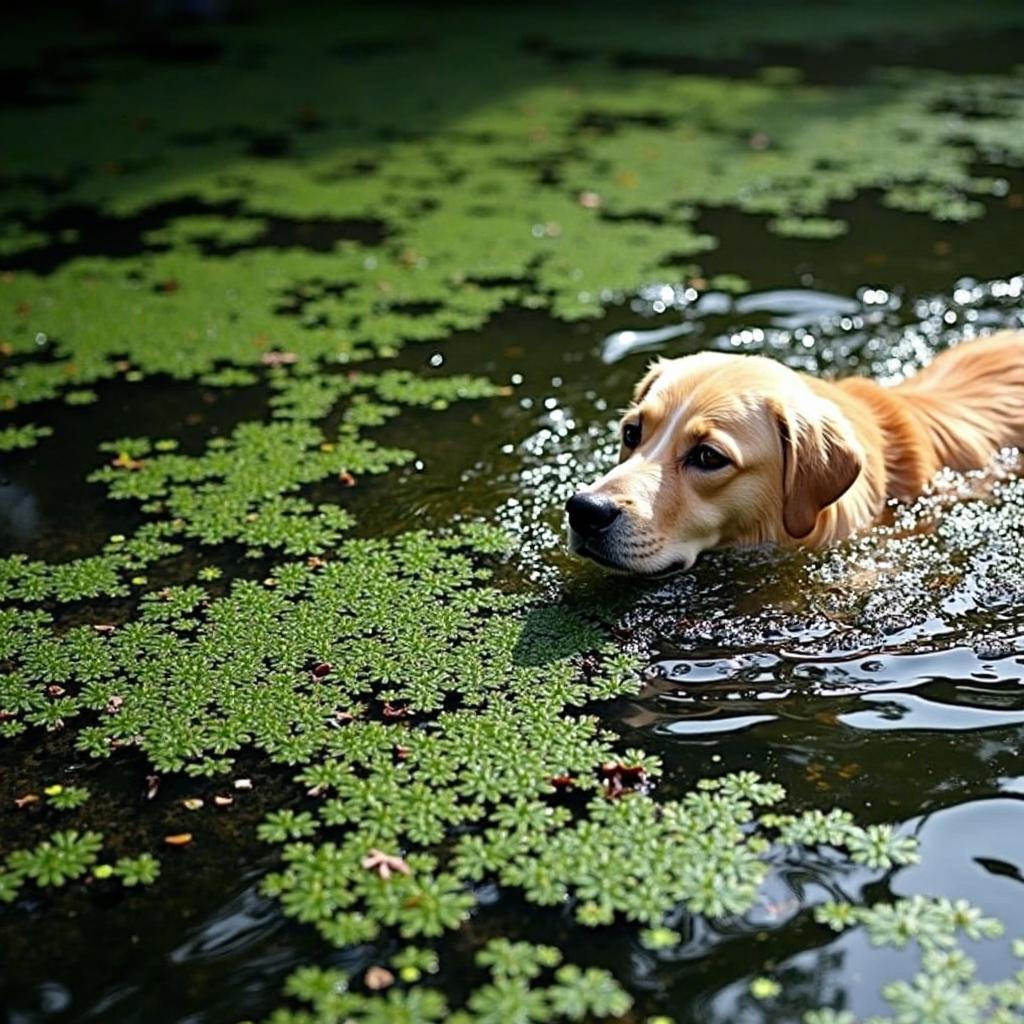} &
      \Thumb{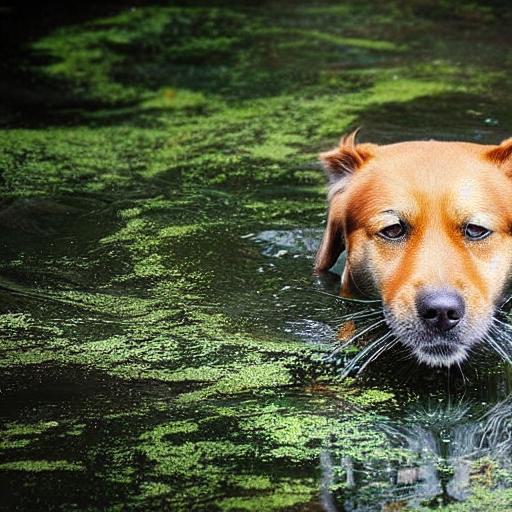}
      \\[-0.18em]
      \multicolumn{9}{c}{\scriptsize tiger $\rightarrow$ dog} \\[0.20em]

        \Thumb{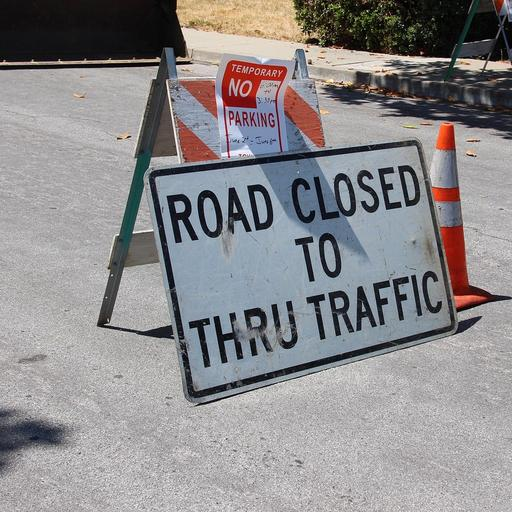} &
        \Thumb{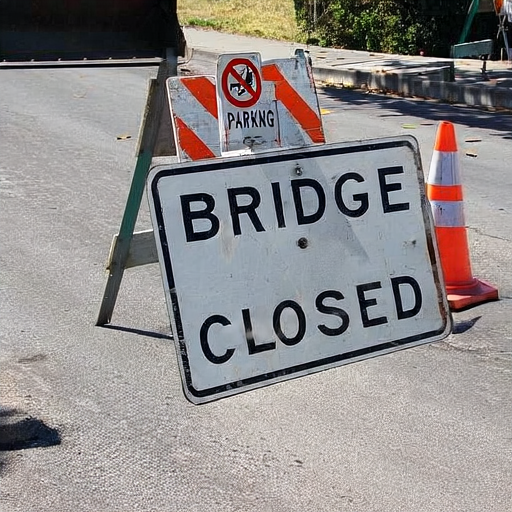} &
        \Thumb{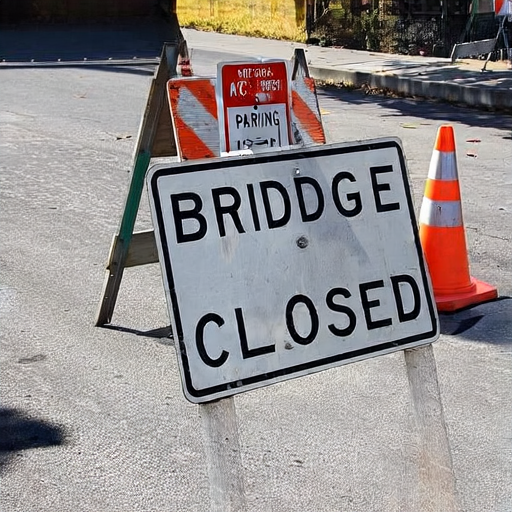} &
        \Thumb{figures/quali_comparai_512/FE/bridge_custom_2.2_eta_1_prog_descendingSGDT_steps_50_n_max_33_n_min_0_n_avg_1_cfg_enc_3.5_cfg_dec13.5_seed41_target.png} &
        \Thumb{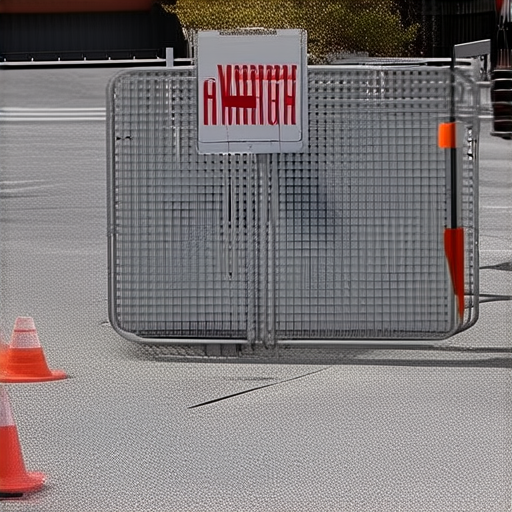} &
        \Thumb{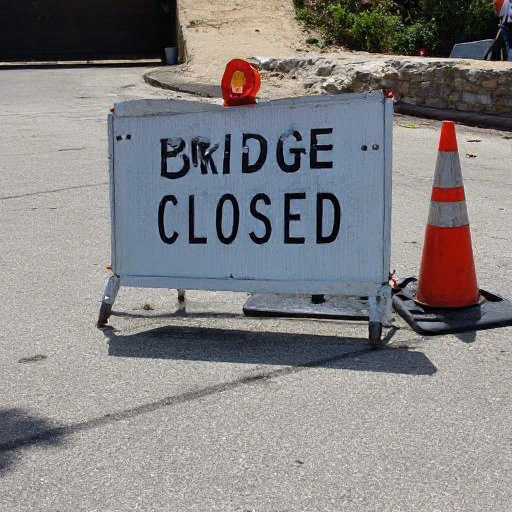} &
        \Thumb{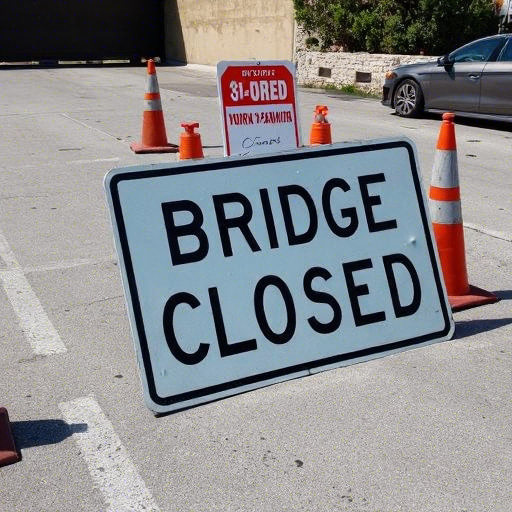} &
        \Thumb{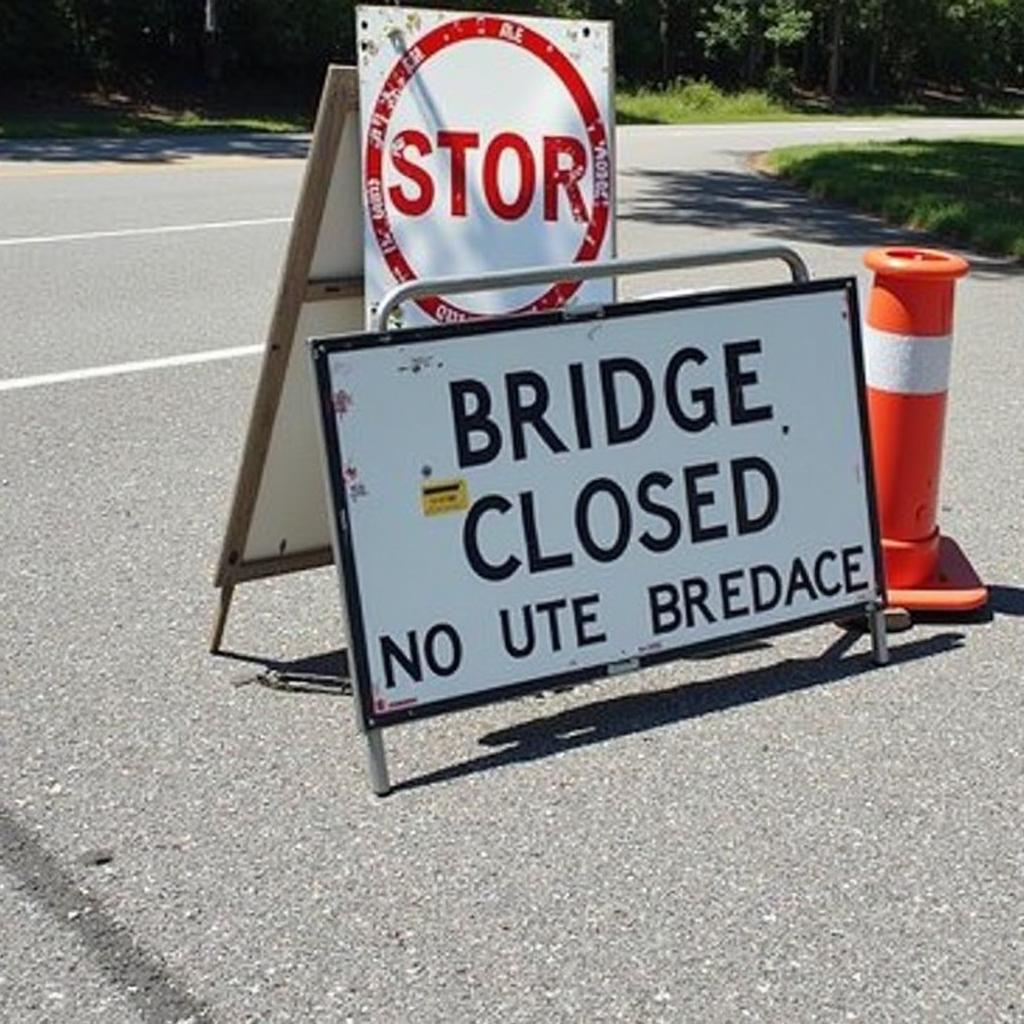} &
        \Thumb{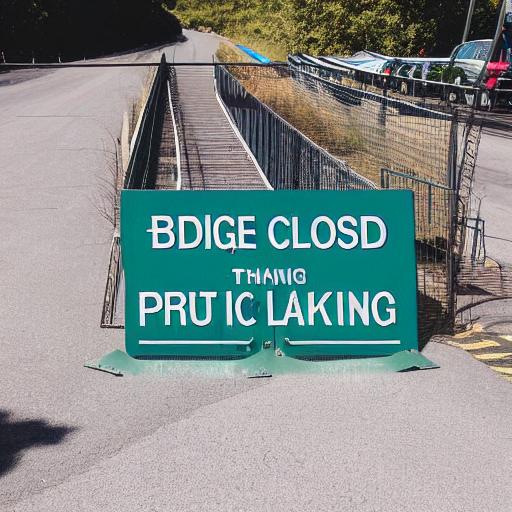} 
        \\[-0.18em]
      \multicolumn{9}{c}{\scriptsize“road closed” sign $\rightarrow$ “bridge closed” sign} \\[0.20em]

      \Thumb{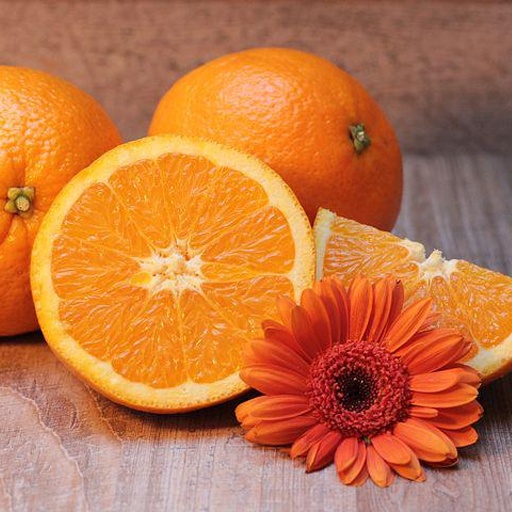} &
      \Thumb{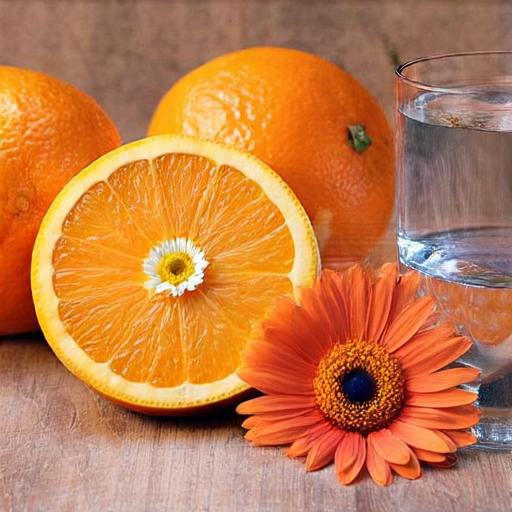} &
      \Thumb{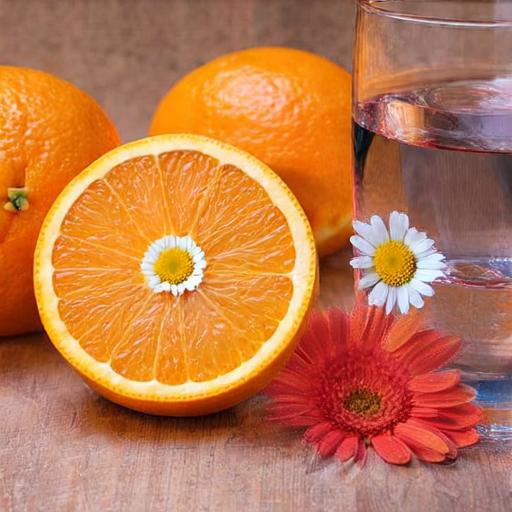} &
      \Thumb{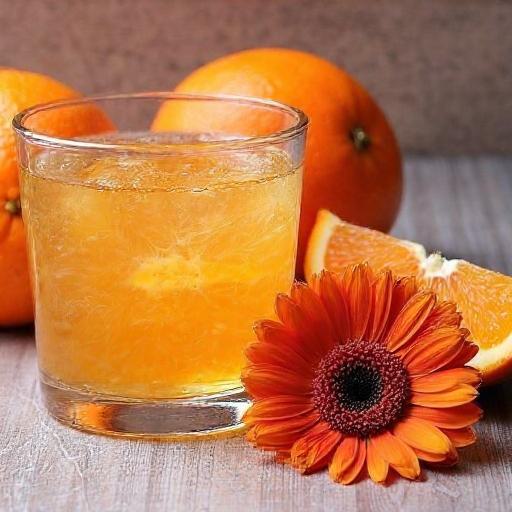} &
      \Thumb{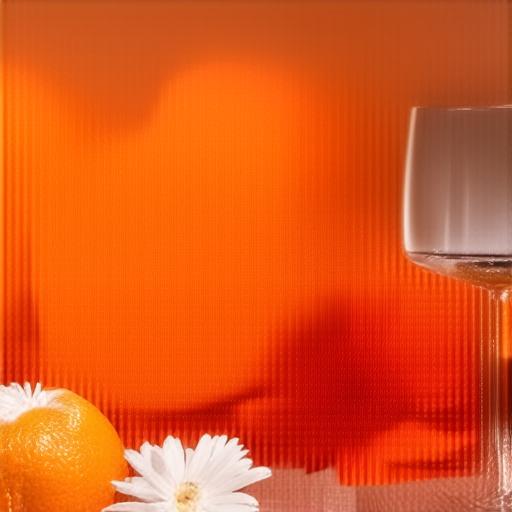} &
      \Thumb{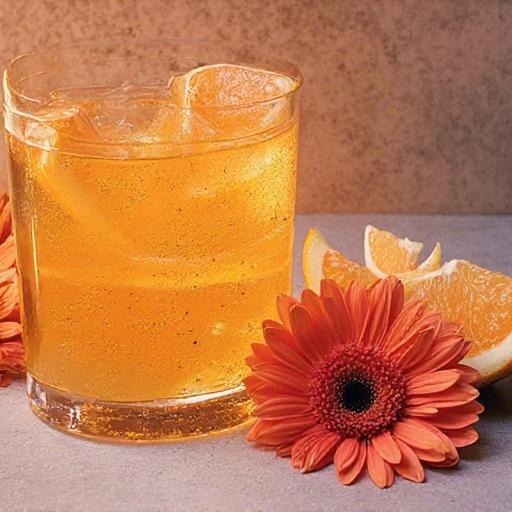} &
      \Thumb{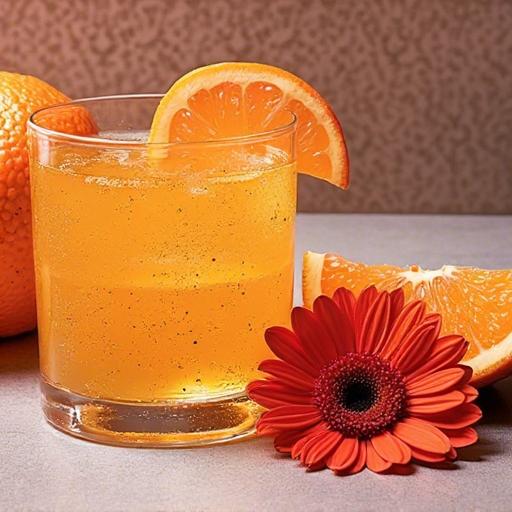} &
      \Thumb{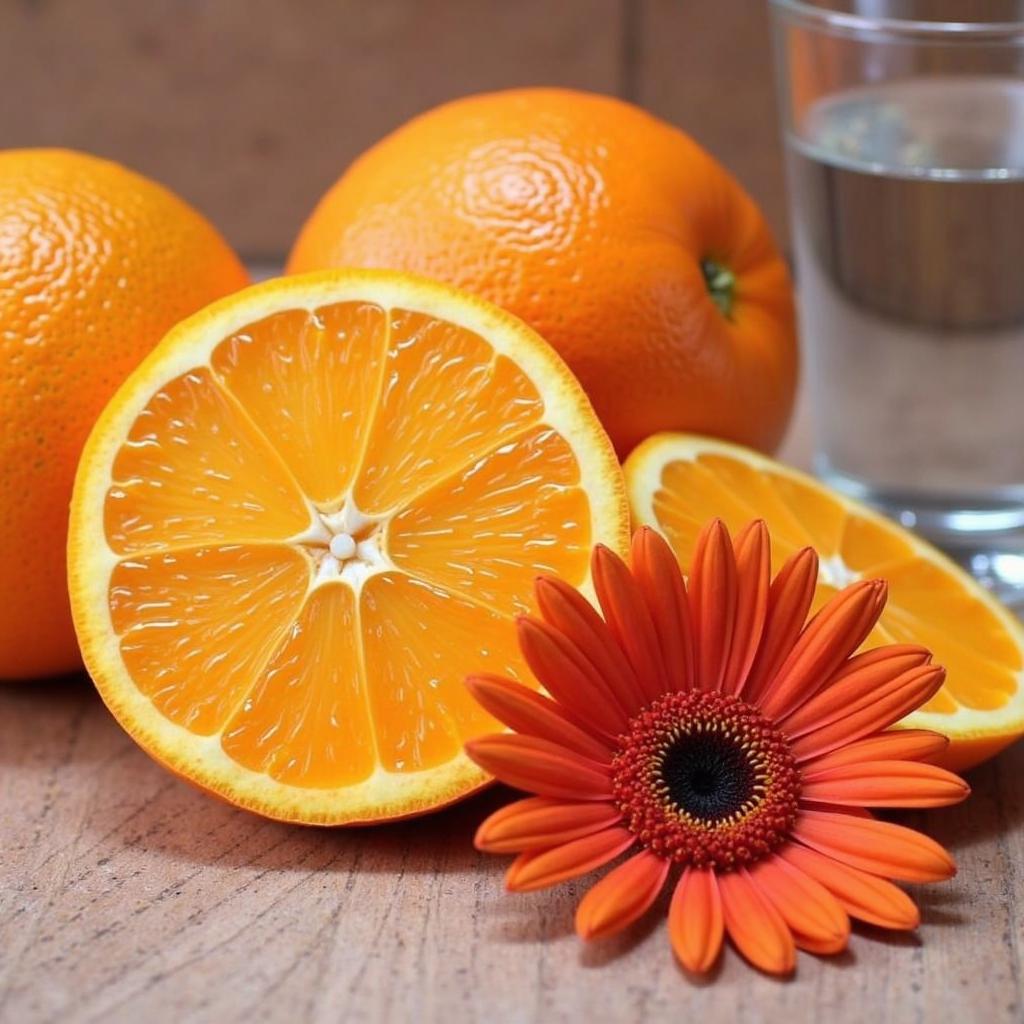} &
      \Thumb{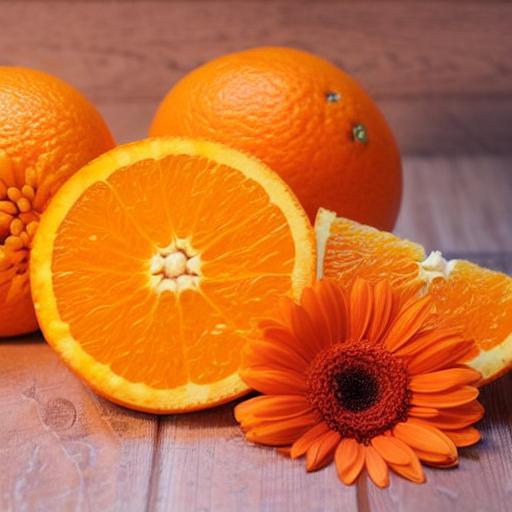}
      \\[-0.18em]
      \multicolumn{9}{c}{\scriptsize + glass of water} \\[0.20em]

      \Thumb{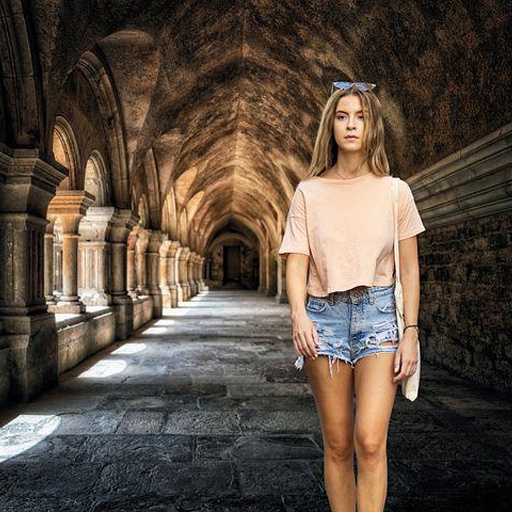} &
      \Thumb{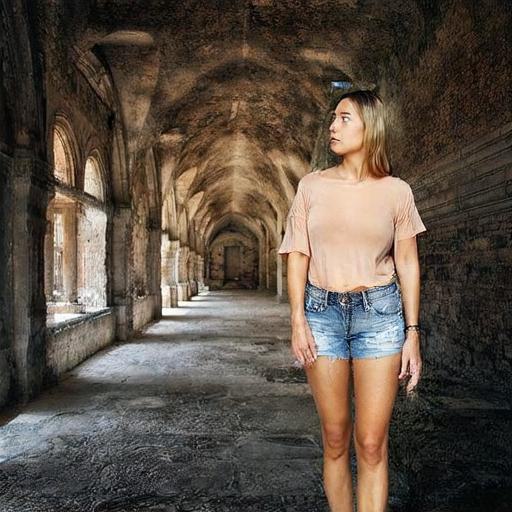} &
      \Thumb{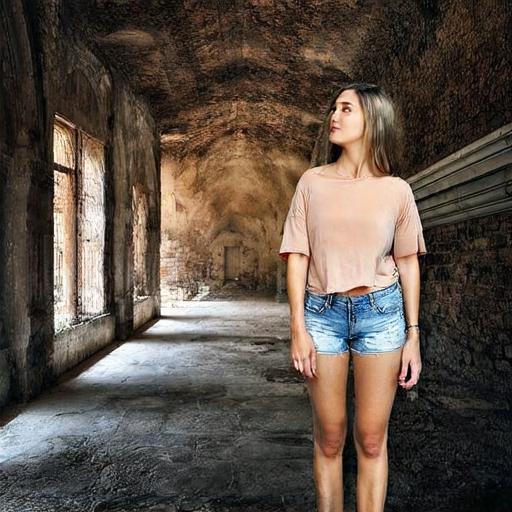} &
      \Thumb{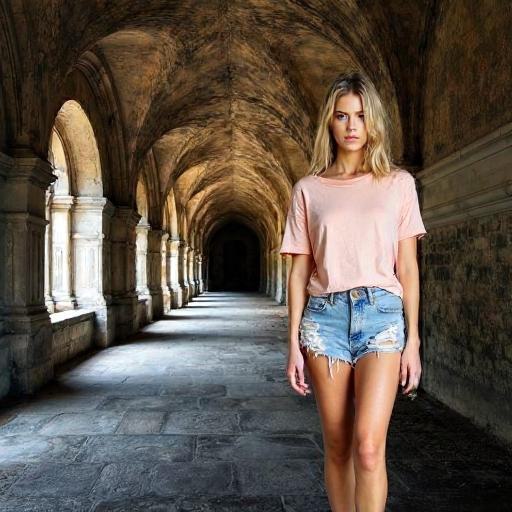} &
      \Thumb{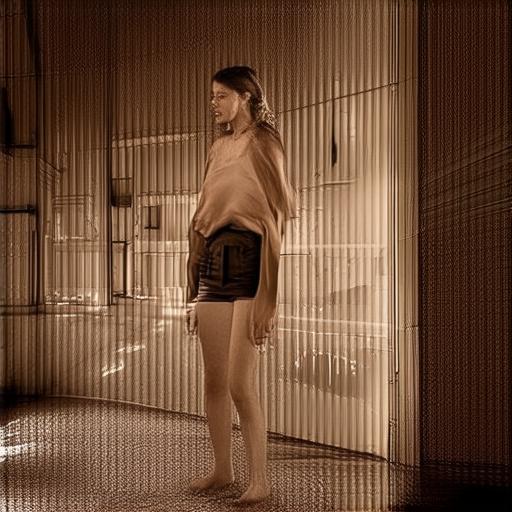} &
      \Thumb{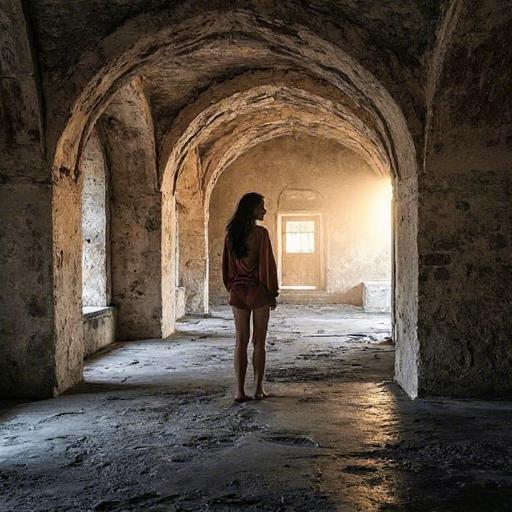} &
      \Thumb{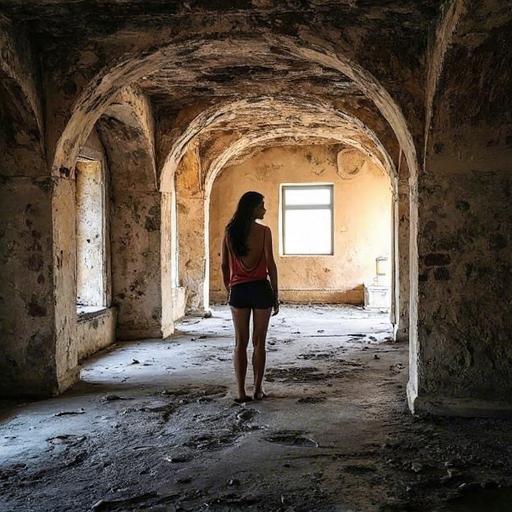} &
      \Thumb{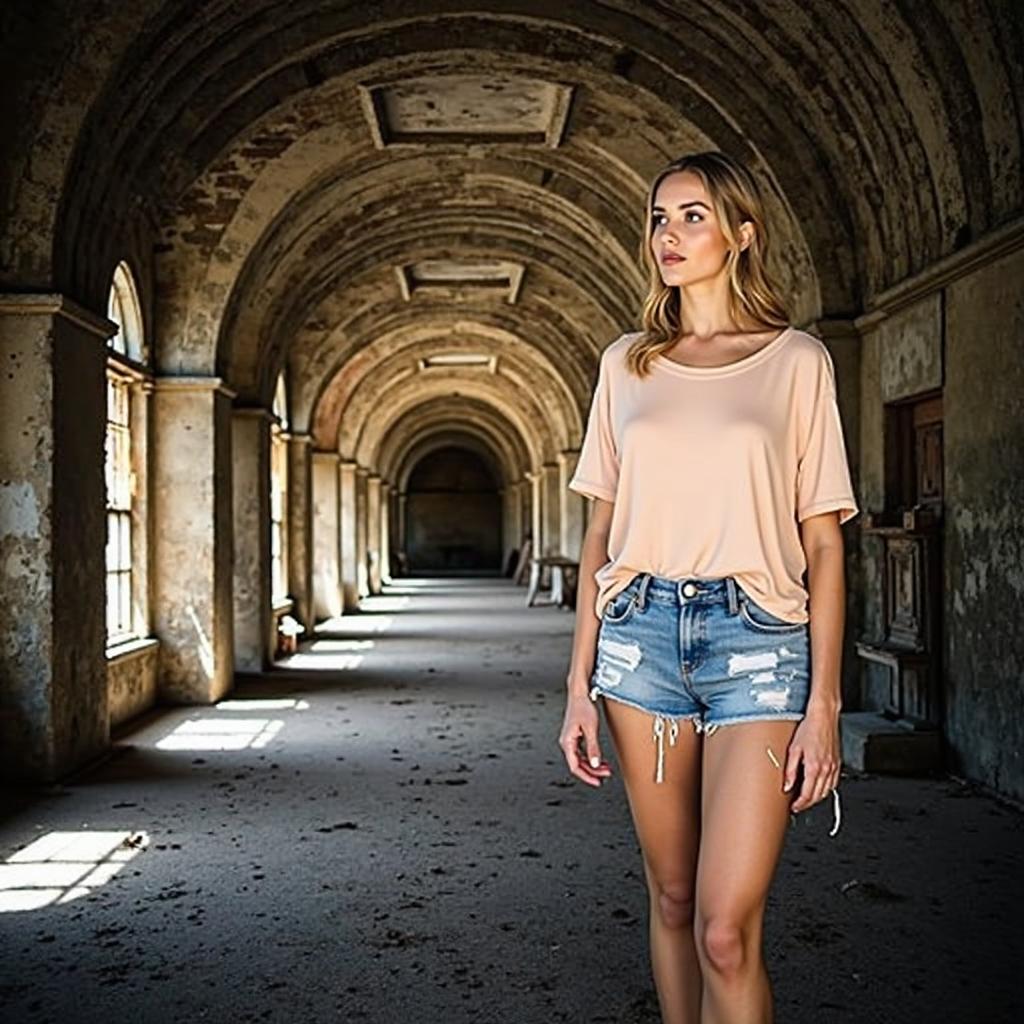} &
      \Thumb{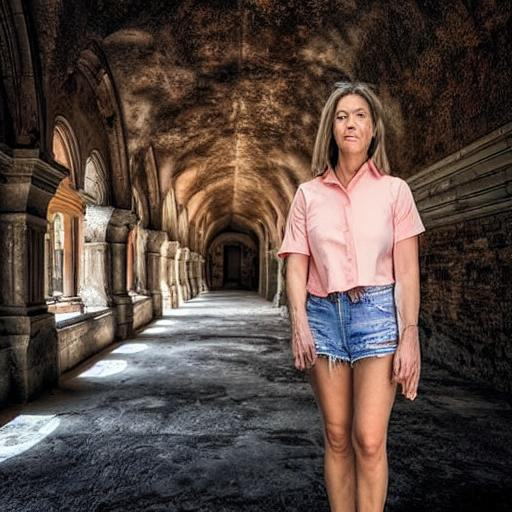}
      \\[-0.18em]
      \multicolumn{9}{c}{\scriptsize + looking at right side}
    \end{tabular}%
  }

  \caption{Qualitative comparisons on images from the PIE benchmark.}
  \label{fig:quali-compa-app}
\end{figure*}

\section{ Failure case study}

\begin{figure*}[ht]
  \centering

  \setlength{\tabcolsep}{0.4pt}
  \renewcommand{\arraystretch}{0.60}

  \newcommand{\figwidth}{0.95\linewidth}

  \setlength{\thumbsize}{\dimexpr(\figwidth - 18\tabcolsep)/9\relax}

  \resizebox{\figwidth}{!}{%
    \begin{tabular}{@{}*{9}{c}@{}}
      \scriptsize Source &
      \scriptsize DRFS &
      \scriptsize FlowEdit (SD3) &
      \scriptsize FlowEdit (Flux) &
      \scriptsize iRFDS &
      \scriptsize FireFlow &
      \scriptsize RF-Solver &
      \scriptsize RF-Inv &
      \scriptsize Direct+P2P \\[0.3em]

      \Thumb{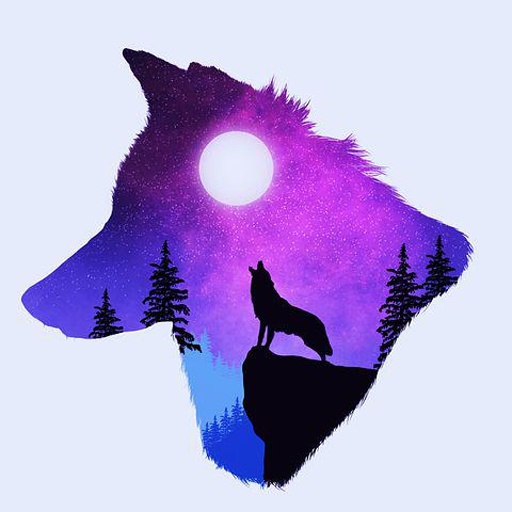} &
      \Thumb{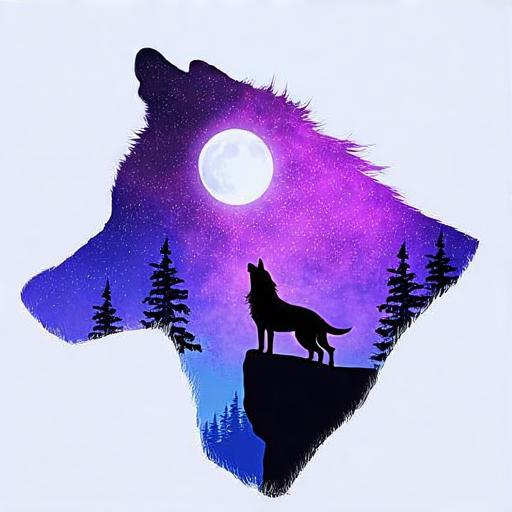} &
      \Thumb{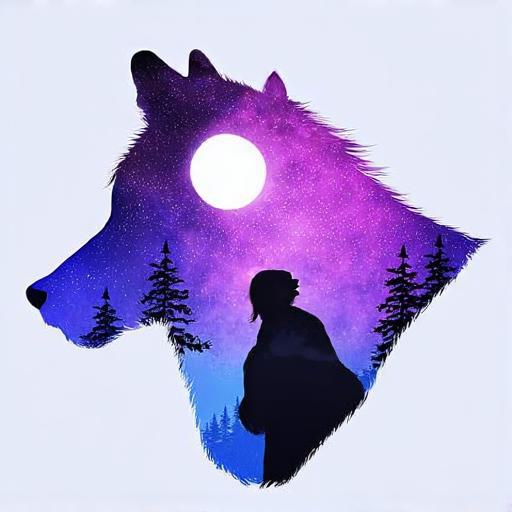} &
      \Thumb{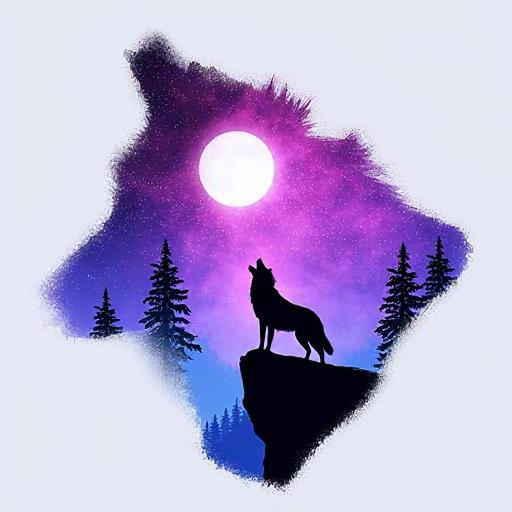} &
      \Thumb{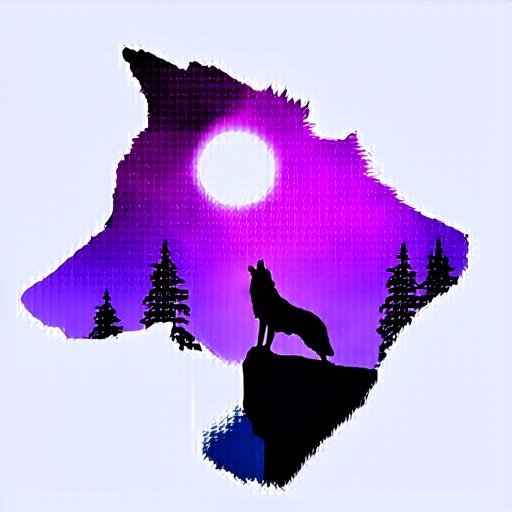} &
      \Thumb{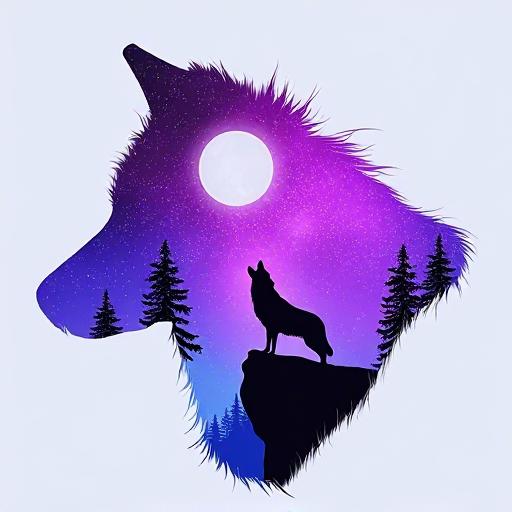} &
      \Thumb{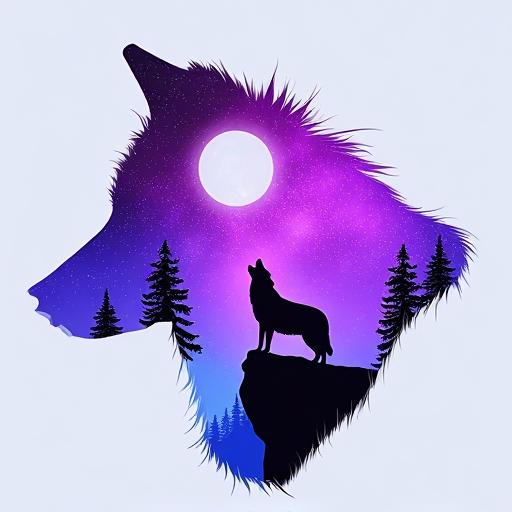} &
      \Thumb{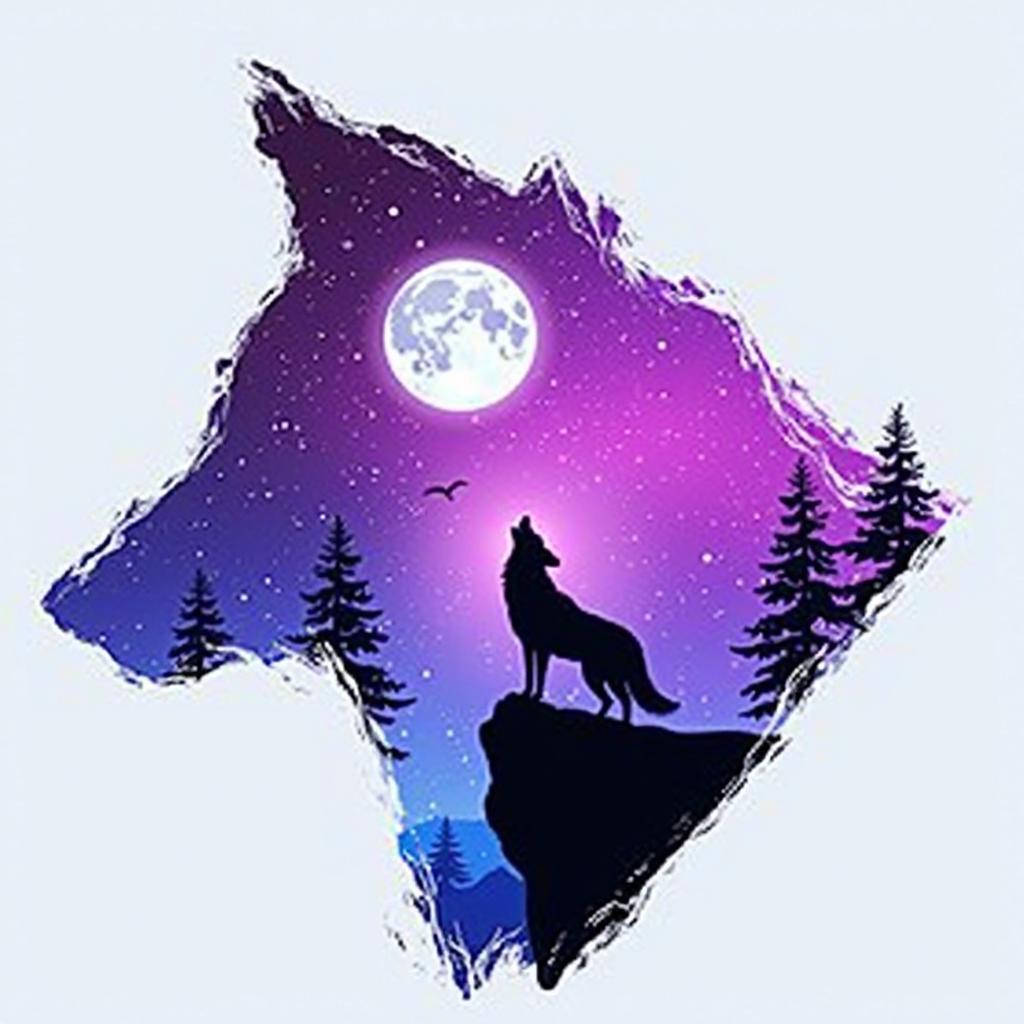} &
      \Thumb{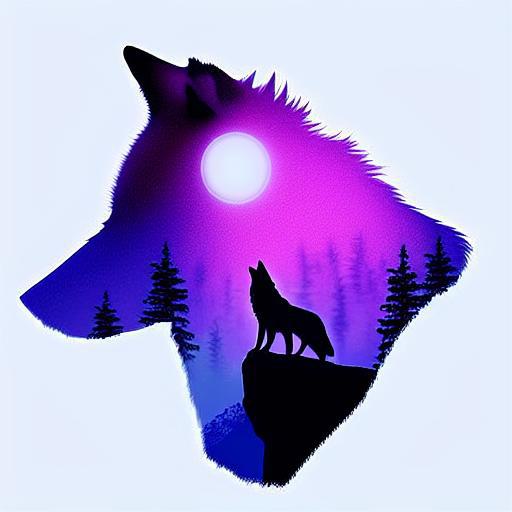}\\[-0.18em]
      \multicolumn{9}{c}{\scriptsize outline of a wolf $\rightarrow$ outline of a man}\\[0.20em]

      \Thumb{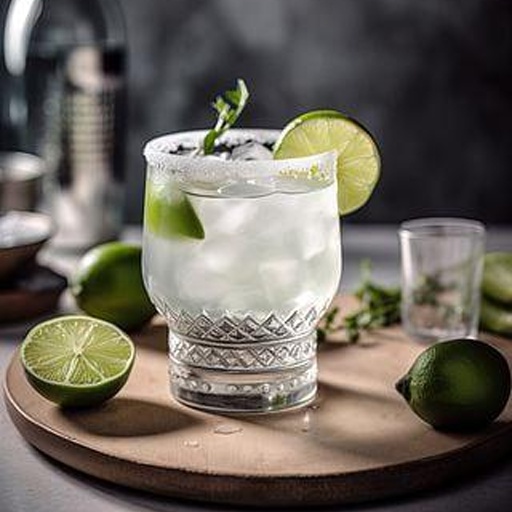} &
      \Thumb{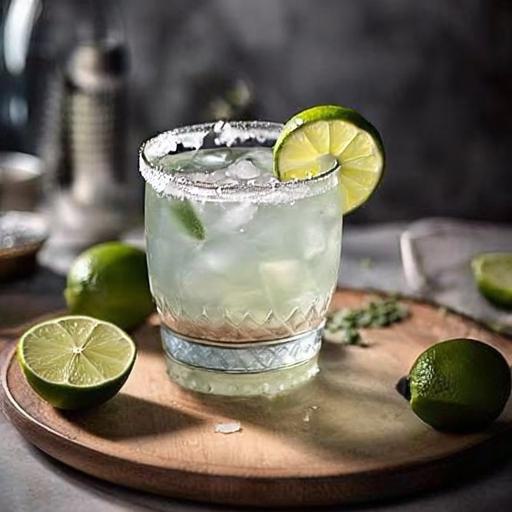} &
      \Thumb{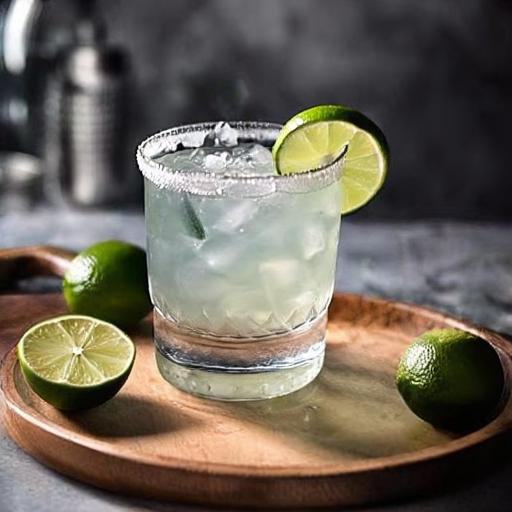} &
      \Thumb{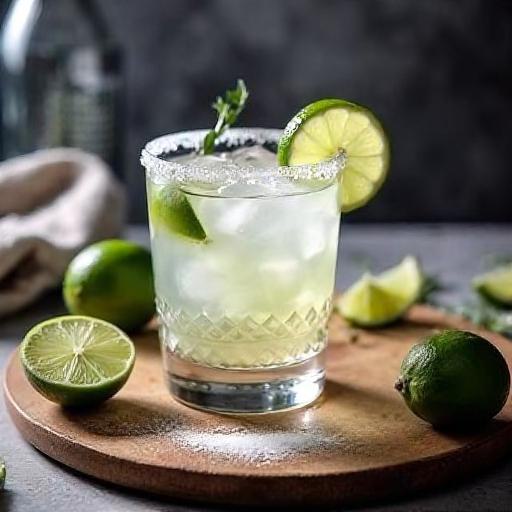} &
      \Thumb{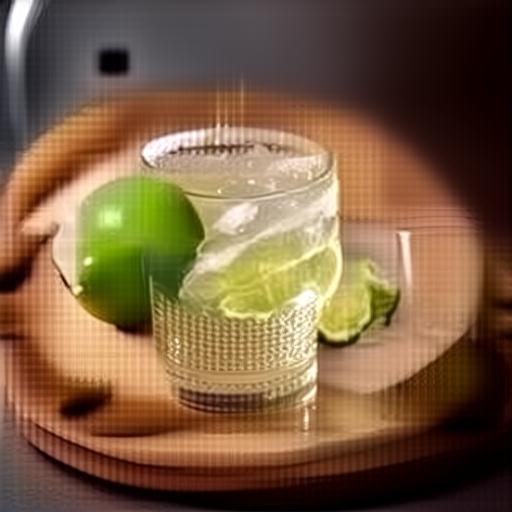} &
      \Thumb{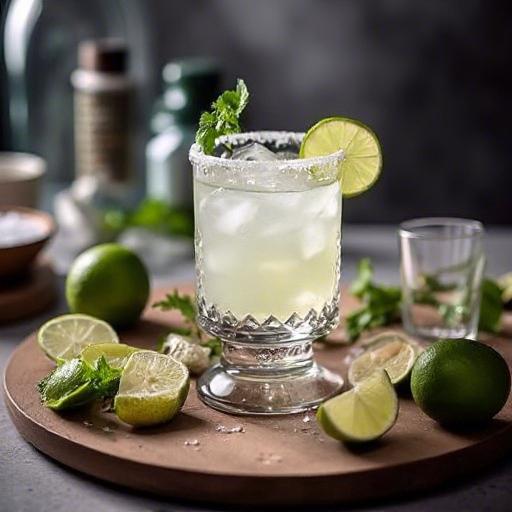} &
      \Thumb{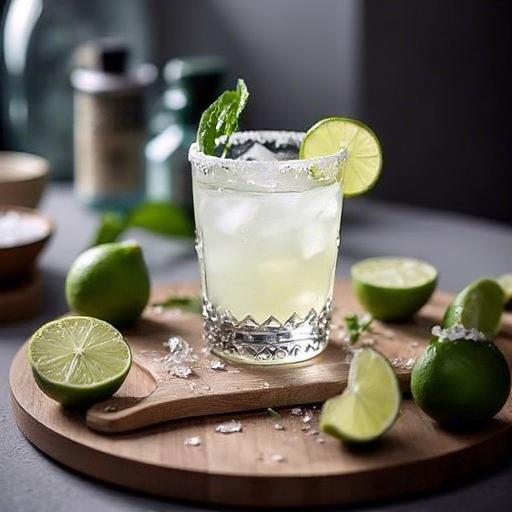} &
      \Thumb{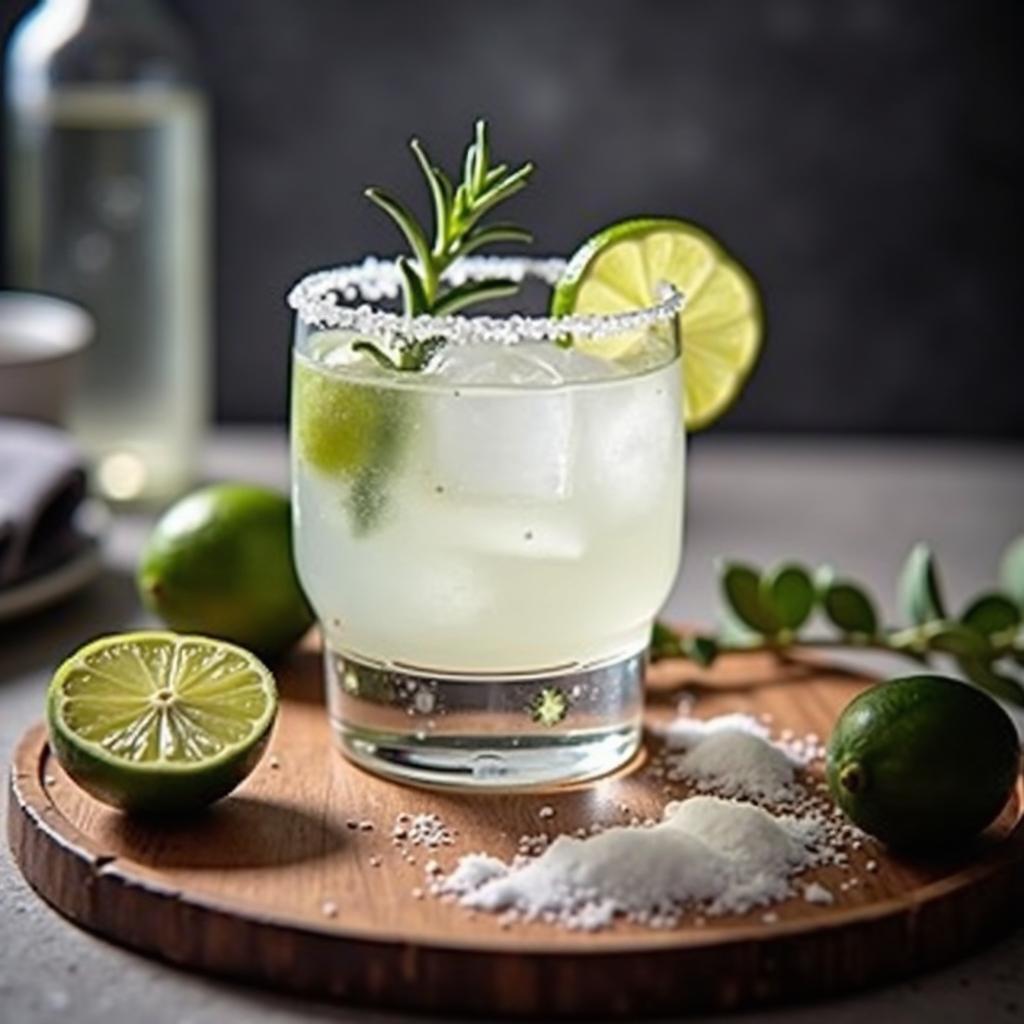} &
      \Thumb{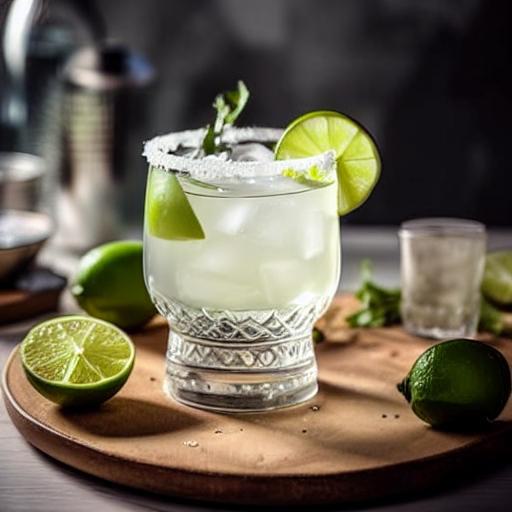}\\[-0.18em]
      \multicolumn{9}{c}{\scriptsize + with aerial view}\\[0.20em]
    \end{tabular}%
  }

  \caption{Examples of failure cases from our method.}
  \label{fig:failures}
\end{figure*}

Fig. \ref{fig:failures} shows several failure cases of DRFS. We observe that these failure cases reflect previously acknowledged common challenges in T2I editing, where a pretrained model struggles to predict an accurate velocity for out-of-distribution images.

Furthermore, in scenarios that require substantial changes, DRFS exhibits limited editing strength due to its inherent design focus on preserving details of the source image. For example, in the first row of Fig.~\ref{fig:failures}, the desired transformation--from \emph{outline of a wolf} to \emph{outline of a man}--is both ambiguous and semantically complex. In the second row, \emph{$+$ with aerial view} requires extensive structural alterations, effectively amounting to the generation of a completely new image. While DRFS enhances both background preservation and alignment with target semantics, these examples underscore fundamental limitations that are prevalent across T2I approaches.

\end{document}